\documentclass[10pt]{article}
\usepackage[preprint]{tmlr} 
\usepackage[utf8]{inputenc}
\usepackage[T1]{fontenc}
\usepackage{amsmath,amssymb,amsfonts}
\usepackage{booktabs,longtable,nicefrac,microtype,graphicx}
\usepackage{hyperref}
\usepackage{url}
\usepackage{xcolor}
\usepackage{tikz}
\usetikzlibrary{arrows.meta}
\usepackage{enumitem}
\setlist[itemize]{topsep=2pt, itemsep=1pt, parsep=0pt}
\usepackage[skip=3pt]{caption}

\newcommand{\ActivationShare}{5.7}

\newcommand{\BootIndHigh}{0.981}
\newcommand{\BootIndLow}{0.946}
\newcommand{\BootRetHigh}{0.720}
\newcommand{\BootRetLow}{0.430}
\newcommand{\CkaInversionPct}{51}

\newcommand{\CkaTotal}{1,080}

\newcommand{\ControlMedian}{0.041}

\newcommand{\HalfwidthRetrieval}{0.061}

\newcommand{\MildBandCount}{22}
\newcommand{\MildBandInduction}{0.959}
\newcommand{\MildBandRetrieval}{0.554}
\newcommand{\MisorderCka}{37.6}
\newcommand{\MisorderMse}{29.0}
\newcommand{\MisorderSds}{25.1}
\newcommand{\NumModels}{12}
\newcommand{\NumPairs}{1,080}

\newcommand{\OverlapMedian}{0.266}

\newcommand{\Pairs}{780}

\newcommand{\PositionsRetrieval}{256}
\newcommand{\PremFeatureEight}{1.00}
\newcommand{\PremFeatureFour}{1.00}
\newcommand{\PremFeatureSix}{1.00}
\newcommand{\PremFeatureThree}{1.01}
\newcommand{\PremFeatureTwo}{16.87}
\newcommand{\PremTensorEight}{31.27}
\newcommand{\PremTensorFour}{655.59}
\newcommand{\PremTensorSix}{230.54}
\newcommand{\PremTensorThree}{856.30}
\newcommand{\PremTensorTwo}{728.48}
\newcommand{\PremTokenEight}{1.00}
\newcommand{\PremTokenFour}{2.25}
\newcommand{\PremTokenSix}{1.00}
\newcommand{\PremTokenThree}{37.36}
\newcommand{\PremTokenTwo}{299.92}

\newcommand{\ReversalInd}{2.1}
\newcommand{\ReversalRet}{4.0}

\newcommand{\ShuntNoiseFour}{0.0219}
\newcommand{\ShuntNoiseTwo}{0.0250}
\newcommand{\ShuntQuantFour}{-0.0231}
\newcommand{\ShuntQuantTwo}{-0.0450}
\newcommand{\SpearmanMax}{0.68}
\newcommand{\SpearmanMedian}{-0.050}
\newcommand{\SpearmanMin}{-0.84}
\newcommand{\TightBandRetrieval}{0.991}

\newcommand{\WorstBlowup}{7,437,245}
\newcommand{\WorstCka}{0.916}

\newcommand{\PythiaAwqIndPct}{16.5}
\newcommand{\PythiaAwqRetPct}{2.8}
\newcommand{\QwenAwqIndPct}{86.3}
\newcommand{\QwenAwqRetPct}{18.8}
\newcommand{\QuarotTwoBitPerToken}{0.291}
\newcommand{\DenseTwoBitPerToken}{0.051}
\newcommand{\QuarotSixBitPerTensor}{0.955}
\newcommand{\DenseSixBitPerTensor}{0.649}
\newcommand{\RotationLatencyLow}{51}
\newcommand{\RotationLatencyHigh}{75}

\newcommand{\TableProbeThreeBitQuant}{0.001}
\newcommand{\TableProbeThreeBitNoise}{0.681}

\newcommand{\TableScaleBody}{%
  Pythia 70M & 1.00 & 1.00 & 1.01 & 6.60 & 30.5 \\
  GPT-2 124M & 1.01 & 1.01 & 1.19 & 18.9 & 93.9 \\
  Pythia 160M & 1.01 & 1.03 & 1.44 & 6.48 & 20.4 \\
  Pythia 410M & 1.01 & 0.99 & 1.18 & 29.7 & 97.5 \\
  Qwen3 1.7B & 0.99 & 1.00 & 1.72 & 21.6 & 771.3 \\
  Pythia 1.4B & 1.02 & 1.07 & 51.6 & 153.9 & 297.7 \\
  Gemma-2 2B & 1.00 & 1.00 & 2.69 & 51.1 & 725.6 \\
  Pythia 2.8B & 1.07 & 1.17 & 104.4 & 169.5 & 83.3 \\
  Pythia 6.9B & 1.04 & 1.33 & 110.7 & 118.9 & 92.1 \\
  Qwen3 8B & 1.02 & 1.01 & 2.20 & 37.9 & 964.7 \\
  Gemma-2 9B & 1.00 & 0.99 & 14.6 & 117.1 & 18,166 \\
  Pythia 12B & 1.04 & 2.42 & 355.7 & 258.2 & 184.4 \\
}

\newcommand{\TableCausalBody}{%
  8 & 0.940 & 0.998 & 1.000 & 1.30 \\
  16 & 0.941 & 0.996 & 1.000 & 1.50 \\
  32 & 0.870 & 0.991 & 1.000 & 1.66 \\
  64 & 0.737 & 0.981 & 1.000 & 1.72 \\
  128 & 0.567 & 0.964 & 0.999 & 1.73 \\
  256 & 0.460 & 0.948 & 0.997 & 2.05 \\
  512 & 0.387 & 0.876 & 0.992 & 2.25 \\
  1024 & 0.261 & 0.753 & 0.983 & 2.39 \\
  2048 & 0.150 & 0.547 & 0.959 & 2.64 \\
}

\newcommand{\TableSensitivityBody}{%
  GPT-2 & -0.040 & -0.61 to +0.18 & 0.320 & 20.5 & 18.0 & 4.6 \\
  Gemma-2 & +0.100 & -0.49 to +0.68 & 0.328 & 2.1 & 3.4 & 5.5 \\
  Pythia & -0.115 & -0.46 to +0.38 & 0.203 & 1.5 & 2.6 & 10.6 \\
  Qwen3 & -0.196 & -0.84 to +0.48 & 0.359 & 3.2 & 3.1 & 6.2 \\
}

\newcommand{\TableInductionBody}{%
  Quantised, one scale per token & 1.000 & 0.998 & 0.681 & 0.001 & 0.000 \\
  Gaussian noise, matched & 1.000 & 0.998 & 0.936 & 0.681 & 0.535 \\
  Signs randomised & 1.000 & 0.998 & 0.943 & 0.695 & 0.493 \\
  \midrule
  Quantised, one scale per tensor & 0.000 & 0.000 & 0.000 & 0.000 & 0.000 \\
  Gaussian noise, matched & 0.684 & 0.500 & 0.450 & 0.453 & 0.443 \\
}

\newcommand{\TablePerModelBody}{%
  Qwen3-1.7B-Base & 0.976 & 0.461 & 0.745 & 0.001 & 0.992 \\
  gemma-2-2b & 0.983 & 0.418 & 0.651 & 0.001 & 0.988 \\
  gpt2 & 0.987 & 0.441 & 0.926 & 0.042 & 0.864 \\
  pythia-1.4b & 0.954 & 0.270 & 0.014 & 0.000 & 0.932 \\
}

\newcommand{\TableProbeBody}{%
  Quantised, 4 bits & 0.681 & 0.662 & 0.272 \\
  Quantised, 3 bits & 0.001 & 0.001 & 0.000 \\
  Gaussian noise, matched, 3 bits & 0.681 & 0.698 & 0.219 \\
  Signs randomised, 3 bits & 0.695 & 0.677 & 0.188 \\
  Rotated basis, 3 bits & 0.980 & 0.981 & 0.694 \\
}

\newcommand{\TableParetoBody}{%
  Uniform & 2.00 & 30,429.6 & 0.000 & 0.000 \\
  Uniform & 3.00 & 1,310.6 & 0.001 & 0.000 \\
  Uniform & 4.00 & 55.1 & 0.681 & 0.272 \\
  Uniform & 6.00 & 25.5 & 0.998 & 0.960 \\
  Uniform & 8.00 & 22.0 & 1.000 & 0.986 \\
  \midrule
  Protected, Fisher & 3.00 & 324.3 & 0.007 & 0.000 \\
  Protected, Fisher & 4.00 & 89.7 & 0.109 & 0.009 \\
  Protected, Fisher & 5.00 & 47.0 & 0.643 & 0.177 \\
  Protected, Fisher & 6.00 & 31.1 & 0.916 & 0.559 \\
  \midrule
  Rotated & 2.00 & 814.0 & 0.051 & 0.009 \\
  Rotated & 3.00 & 33.1 & 0.980 & 0.694 \\
  Rotated & 4.00 & 23.5 & 0.997 & 0.949 \\
  Rotated & 6.00 & 21.9 & 1.000 & 1.000 \\
  Rotated & 8.00 & 21.7 & 1.000 & 1.000 \\
}

\newcommand{\TableAllLayerFullBody}{%
  Uniform & 2.00 & 199,138.2 & 0.000 & 0.000 \\
  Uniform & 3.00 & 453,736.7 & 0.000 & 0.000 \\
  Uniform & 4.00 & 20,442.3 & 0.003 & 0.000 \\
  Uniform & 6.00 & 38.4 & 0.905 & 0.597 \\
  Uniform & 8.00 & 25.1 & 0.999 & 0.965 \\
  \midrule
  Matched noise & 2.00 & 7,209,360.4 & 0.000 & 0.000 \\
  Matched noise & 3.00 & 417,203.6 & 0.000 & 0.000 \\
  Matched noise & 4.00 & 15,815.9 & 0.000 & 0.000 \\
  Matched noise & 6.00 & 50.1 & 0.894 & 0.407 \\
  Matched noise & 8.00 & 27.8 & 0.997 & 0.932 \\
  \midrule
  Protected, Fisher & 3.00 & 21,227.3 & 0.000 & 0.000 \\
  Protected, Fisher & 4.00 & 4,893.3 & 0.000 & 0.000 \\
  Protected, Fisher & 5.00 & 1,287.5 & 0.007 & 0.000 \\
  Protected, Fisher & 6.00 & 341.6 & 0.235 & 0.004 \\
  \midrule
  Rotated & 2.00 & 78,269.3 & 0.001 & 0.000 \\
  Rotated & 3.00 & 201.0 & 0.380 & 0.072 \\
  Rotated & 4.00 & 35.7 & 0.968 & 0.534 \\
  Rotated & 6.00 & 22.8 & 0.999 & 0.972 \\
  Rotated & 8.00 & 22.2 & 1.000 & 1.004 \\
}

\newcommand{\TableRestoreBody}{%
  Rotated basis, one scale per token & 1.000 & 1.000 & 0.997 & 0.980 & 0.051 \\
  Rotated basis, one scale per tensor & 0.997 & 0.680 & 0.000 & 0.000 & 0.000 \\
}

\newcommand{\TableProtectBody}{%
  3.00 & 0.007 & 0.001 & 0.001 at 3 \\
  4.00 & 0.109 & 0.098 & 0.681 at 4 \\
  5.00 & 0.643 & 0.591 & 0.998 at 6 \\
  6.00 & 0.916 & 0.893 & 0.998 at 6 \\
}

\newcommand{\TableLadderBody}{%
  Gaussian, second moment matched & 0.631 & 0.632 & 0.173 & 0.020 & 0.001 \\
  Signs randomised, magnitudes exact & 0.651 & 0.646 & 0.164 & 0.017 & 0.001 \\
  Errors moved to other tokens & 0.473 & 0.730 & 0.201 & 0.034 & 0.004 \\
  Quantised in a rotated basis & 0.064 & 0.188 & 0.047 & 0.011 & 0.033 \\
}

\newcommand{\TableSignDialBody}{%
  0\% & 0.731 & 0.946 & 0.266 & 0.412 \\
  25\% & 0.846 & 0.946 & 0.288 & 0.412 \\
  50\% & 0.889 & 0.946 & 0.298 & 0.412 \\
  75\% & 0.931 & 0.946 & 0.313 & 0.412 \\
  100\% & 0.949 & 0.946 & 0.400 & 0.412 \\
}

\newcommand{\TableShuntBody}{%
  8 & +0.0005 & +0.0007 \\
  6 & +0.0009 & +0.0052 \\
  4 & -0.0231 & +0.0219 \\
  3 & -0.0397 & +0.0242 \\
  2 & -0.0450 & +0.0250 \\
}

\newcommand{\TableAllocationBody}{%
  2 & 75,912 & 12,923 & 5.87 & 6.27 \\
  3 & 87,789 & 9,077 & 9.67 & 1.90 \\
  4 & 56,719 & 2,478 & 22.89 & 0.16 \\
  6 & 10,541 & 504 & 20.91 & 0.37 \\
  8 & 2,608 & 294 & 8.88 & 0.07 \\
}

\newcommand{\TableDictionaryBody}{%
  pythia-70m & 512 & 4096 & 8 & 64 & 0.143 \\
  pythia-160m & 768 & 6144 & 8 & 64 & 0.165 \\
  pythia-410m & 1024 & 8192 & 8 & 64 & 0.174 \\
  pythia-1.4b & 2048 & 16384 & 8 & 64 & 0.222 \\
  pythia-2.8b & 2560 & 20480 & 8 & 64 & 0.243 \\
  pythia-6.9b & 4096 & 32768 & 8 & 64 & 0.264 \\
  pythia-12b & 5120 & 40960 & 8 & 64 & 0.282 \\
}

\newcommand{\TableConcentrationBody}{%
  Quantisation & 5.6 & 11.5 & 15.7 & 25.6 \\
  Matched noise & 3.2 & 7.1 & 10.5 & 17.6 \\
  \midrule
  Activation reference & 5.7 & 11.6 & 16.7 & 26.1 \\
}

\newcommand{\TableDoseModelBody}{%
  pythia-1.4b & 5 & 0.943 & 0.899 to 0.969 & 0.681 & 0.464 to 0.870 \\
  Qwen3-1.7B-Base & 3 & 0.984 & 0.480 to 0.985 & 0.720 & 0.034 to 1.025 \\
  gemma-2-2b & 5 & 0.960 & 0.711 to 0.987 & 0.551 & 0.374 to 1.000 \\
  gpt2 & 9 & 0.969 & 0.842 to 0.993 & 0.372 & 0.230 to 0.655 \\
}

\newcommand{\TableDoseBody}{%
  1 to 1.2 & 95 & 0.999 & 0.991 \\
  1.2 to 1.5 & 22 & 0.959 & 0.554 \\
  1.5 to 2 & 26 & 0.929 & 0.502 \\
  2 to 3 & 17 & 0.689 & 0.246 \\
  3 to 5 & 15 & 0.377 & 0.080 \\
  5 to 10 & 19 & 0.172 & 0.019 \\
  above 10 & 136 & 0.000 & 0.000 \\
}

\newcommand{\TableRealQuantAppendixBody}{%
  \multicolumn{7}{l}{\emph{pythia-1.4b, full precision: perplexity 19.7, induction 0.955, retrieval 0.281}} \\
  Uniform (activation) & activation & 4 & W4A4 & 31{,}098.5 & 0.005 & 0.000 \\
  Rotated (activation) & activation & 4 & W4A4 & 31.8 & 0.842 & 0.191 \\
  GPTQ & weight & 3 & W3A16 & 22.8 & 0.916 & 0.262 \\
  GPTQ & weight & 4 & W4A16 & 20.1 & 0.948 & 0.273 \\
  GPTQ & weight & 4 & W4A4 & 1{,}037.7 & 0.019 & 0.000 \\
  AWQ & weight & 3 & W3A16 & 24.3 & 0.912 & 0.207 \\
  AWQ & weight & 4 & W4A16 & 20.5 & 0.949 & 0.266 \\
  AWQ & weight & 4 & W4A4 & 174.2 & 0.158 & 0.008 \\
  SmoothQuant & weight & 3 & W3A16 & 25.1 & 0.910 & 0.176 \\
  SmoothQuant & weight & 4 & W4A16 & 20.5 & 0.948 & 0.266 \\
  SmoothQuant & weight & 8 & W8A8 & 20.4 & 0.949 & 0.258 \\
  \midrule
  \multicolumn{7}{l}{\emph{Qwen3-8B-Base, full precision: perplexity 19.9, induction 0.973, retrieval 0.480}} \\
  Uniform (activation) & activation & 4 & W4A4 & 20{,}965.9 & 0.002 & 0.000 \\
  Rotated (activation) & activation & 4 & W4A4 & 26.1 & 0.966 & 0.379 \\
  GPTQ & weight & 3 & W3A16 & 21.7 & 0.966 & 0.402 \\
  GPTQ & weight & 4 & W4A16 & 20.3 & 0.972 & 0.477 \\
  GPTQ & weight & 4 & W4A4 & 1{,}117.3 & 0.147 & 0.020 \\
  AWQ & weight & 3 & W3A16 & 25.4 & 0.959 & 0.375 \\
  AWQ & weight & 4 & W4A16 & 21.0 & 0.970 & 0.383 \\
  AWQ & weight & 4 & W4A4 & 47.4 & 0.840 & 0.090 \\
  SmoothQuant & weight & 3 & W3A16 & 36.5 & 0.933 & 0.215 \\
  SmoothQuant & weight & 4 & W4A16 & 22.8 & 0.971 & 0.473 \\
  SmoothQuant & weight & 8 & W8A8 & 22.1 & 0.974 & 0.531 \\
}

\newcommand{\TableQuarotMainBody}{%
  8 & 0.980 & 0.979 & 0.967 & 0.979 \\
  6 & 0.979 & 0.979 & 0.649 & 0.955 \\
  4 & 0.974 & 0.978 & 0.000 & 0.002 \\
  3 & 0.965 & 0.968 & 0.000 & 0.000 \\
  2 & 0.051 & 0.291 & 0.000 & 0.000 \\
}

\newcommand{\TableQuarotLatencyBody}{%
  Pythia-1.4b & 0.83 & 1.46 (+75\%) & 1.45 (+74\%) \\
  GPT-2 & 0.32 & 0.52 (+65\%) & 0.52 (+65\%) \\
  Gemma-2-2b & 1.60 & 2.43 (+52\%) & 2.41 (+51\%) \\
  Qwen3-1.7B & 1.12 & 1.77 (+58\%) & 1.76 (+58\%) \\
}

\title{Perplexity Ranks Activation Quantisers\\ but Understates What They Remove}
\author{\name Anish Sathyanarayanan \\
      \addr Birla Institute of Technology and Science, Goa
      \AND
      \name Min Xu \\
      \addr Carnegie Mellon University}

\begin{document}
\maketitle
\begin{abstract}
Four-bit activation quantisation of language models is usually tuned and compared by
perplexity. We ask what that misses, and find three things. First, perplexity gets the order of
quantisers right but the size of their damage wrong. Across four models it ranks 33 of 34 pairs of
settings correctly by closed-book QA loss, yet a 6--16\% rise in perplexity comes with a 10--23\% drop in
QA accuracy, and the lost facts are mostly rare ones. Second, the ranking only holds while the
quantisation error acts like random noise. Without rotation, rounding small values to zero deletes part
of the signal, and this hurts in-context retrieval far more than noise of the same size would. The crest
factor of the quantiser's input (its peak-to-average ratio) helps tell the two cases apart. Third,
rotating in blocks, the usual workaround when a layer's width has no ready-made Hadamard matrix, can land
a model in between and leave outliers behind; 10 of 42 popular open models have such a width. A
full-width Hadamard exists for every such width we found, costs about the same, and on Qwen3-4B halves
the closed-book TriviaQA loss (0.112 to 0.060, over three rotation seeds). We recommend reporting a
knowledge-heavy task alongside perplexity, and rotating at full width wherever blocks leave an input
outlier-heavy.
\end{abstract}

\section{Introduction}
\label{sec:intro}

\begin{figure}[t]
  \centering
  \includegraphics[width=\linewidth]{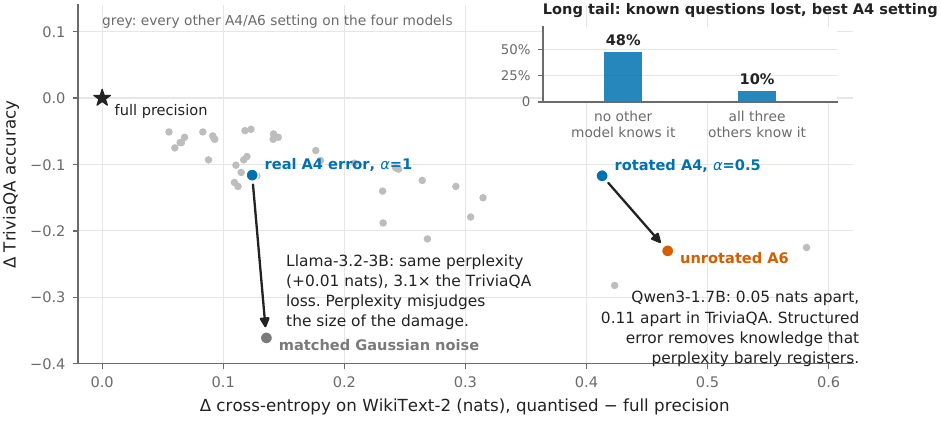}
  \caption{Perplexity understates what activation quantisation removes. Each grey point is one A4 or A6
  setting of one of four models, plotted as its change in WikiText-2 cross-entropy against its change in
  closed-book TriviaQA accuracy (1000 questions). The arrows mark two examples. On Llama-3.2-3B, Gaussian
  noise matched to the real A4 error ($\alpha = 1$) gives the same perplexity but loses 3.1 times as much
  TriviaQA accuracy. On Qwen3-1.7B, an unrotated A6 model is only 0.05 nats from a rotated A4 model
  ($\alpha = 0.5$) but loses 0.11 more TriviaQA accuracy. Perplexity gets the order right but barely
  separates the two. Inset: of the TriviaQA and NQ-Open questions a model gets right at full precision,
  the share its best A4 setting loses, split by whether none or all three of the other full-precision
  models also get the question right (pooled over the four models).}
  \label{fig:teaser}
\end{figure}

Language models are usually quantised before they are deployed, and the quantised model is usually
judged by its perplexity. We already know that compression can hurt some skills more than perplexity
suggests \citep{jaiswal2024compressing,dutta2024accuracy,duan2026quantization}. Here we look closely at
one case: four-bit activation quantisation. Today this is commonly done after an orthogonal
rotation that spreads out activation outliers
\citep{ashkboos2024quarot,liu2024spinquant,liu2024duquant,sun2024flatquant}, and the resulting methods
are usually tuned and compared by perplexity. Our question is simple. What does perplexity get
right about these quantisers, and what does it get wrong?

The short answer is that perplexity gets the \emph{order} right and the \emph{size} wrong. If one
quantiser raises perplexity more than another, it usually also does more damage on natural tasks. But
the damage is larger than the perplexity change suggests, and it lands on the things perplexity weighs
least: rarely seen facts, and retrieval from the context (Figure~\ref{fig:teaser}). The order itself
can only be trusted while the quantisation error behaves like random noise. Whether it does depends on
how spiky the quantiser's input is (outlier-heavy, i.e.\ with a high crest factor). Following that
thread led us to a rotation workaround that
leaves some inputs spiky, and to a fix that costs about the same.

\paragraph{Perplexity gets the order right and the size wrong.}
We quantise every linear-layer input of four models to four bits per token, after QuaRot's Hadamard
rotation, and test them on closed-book QA (TriviaQA, NQ-Open), reading comprehension (CoQA, DROP) and six
long-document QA tasks from LongBench (Section~\ref{sec:natural}). Every quantised setting loses
closed-book QA on every model, including the best-tuned ones. Within a model, the setting with the
larger perplexity increase also has the larger closed-book loss in 33 of 34 pairs. That is better than
a random-token retrieval probe manages (27 of 34). The sizes are another matter. At the tuned settings
perplexity rises by 6--16\%, while closed-book accuracy falls by 10--23\%.

The lost answers are knowledge being removed, not reshuffled. Each model loses three to seventeen times
as many questions as it gains. Different quantisers lose largely the same questions. And a fact that
none of the other models knows is lost six to eight times as often as one they all know
(Section~\ref{sec:longtail}). Noise of the same size removes the same kind of fact, so this long tail is
simply where any perturbation of the activations does its damage. Perplexity on running text is
dominated by common patterns, so it is the measure least sensitive to that damage.

\paragraph{When the order breaks: deleted signal is not noise.}
The order only holds while the error acts like noise. In a controlled study that quantises the residual
stream at a single block of \NumModels{} models (Section~\ref{sec:mask}), we compare the quantiser with
Gaussian noise of the same size, and with the real error after randomly flipping its signs. Both leave
far more of the model's in-context abilities intact than the real error does. The reason is the
quantiser's \emph{deadzone}: small values round to zero, so part of the signal is deleted in a
coordinated way.

In the deployed setting (Section~\ref{sec:regimes}), the \emph{crest factor} of the quantiser's input,
its largest value divided by its root-mean-square, separates two regimes. We treat it as a warning sign
rather than as a predictor of benchmark scores. Unrotated inputs are very spiky (crest factors of 7 to
47). Most of their values fall in the deadzone, and that deletion alone reproduces all of the retrieval
damage. On Qwen3-1.7B, an unrotated model loses 0.11 more TriviaQA accuracy than a rotated one with
slightly lower perplexity. After a full Hadamard rotation the input looks Gaussian, the error behaves
much more like noise, and choosing the clipping ratio by perplexity works well. Inputs that are only
rotated in blocks fall in between.

\paragraph{A gap in partial rotation, and a cheap fix.}
QuaRot can only build Hadamard matrices for some widths, and its reference code rejects the rest. Ten
of 42 popular open models have at least one such width, usually the input to the MLP down-projection
(Table~\ref{tab:widths}). The obvious workaround is to rotate in smaller blocks that QuaRot does
support, for example 19 blocks of 512 for width 9728. Deployed tooling offers the same option:
llm-compressor's SpinQuant and QuIP rotations take a block size, and its examples use 128
(Section~\ref{sec:setup}). Its examples are not our exact setting, though. The one that also quantises
activations uses NVFP4, which shares a scale across every 16 values rather than across a whole token.

Block rotation can leave an input spiky. On Qwen3-4B (width 9728), the block-rotated input has a crest
factor of 7.0, where every fully rotated input sits at 3.5--3.9. Its quantisation error is three to four
times larger, and quantising this one input alone raises perplexity from 7.35 to 10.3. Yet a full-width
Hadamard of the same Kronecker kind QuaRot already uses exists for this width ($H_{76} \otimes H_{128}$),
and for every other width in our survey. The construction is classical, and compressed-tensors, the
library under llm-compressor, already stores $H_{76}$ and $H_{148}$. What we add is evidence that the
choice between block and full width matters, and a way to tell which inputs it matters for.

The full-width rotation costs about as much as the block rotation on 256 tokens or more. On Qwen3-4B it
halves the closed-book TriviaQA loss (0.112 to 0.060 over three rotation seeds, clear on each seed),
shrinks the NQ-Open loss less (0.072 to 0.056, clear only when seeds are pooled), and brings in-context
recall to within 0.02 of full precision (Section~\ref{sec:partial}). A learned SpinQuant-style rotation of
the other inputs does not make it unnecessary (Section~\ref{sec:spin}). On Qwen2.5-7B, whose block-rotated
input is the next spikiest we measured, the gain is smaller and shows up on TriviaQA only (a loss of
0.018 against 0.038 over three seeds, resolved only when the seeds are pooled). On Qwen2.5-0.5B, whose block-rotated input is already close to
Gaussian (crest factor 4.1, against 3.8 at full width), the fix changes little
(Section~\ref{sec:partial-q25}). The crest factor flagged in advance which of these inputs would benefit.

\paragraph{Other checks.}
Random-token retrieval probes miss the full-width rotation's gain on natural recall. A probe with
ordinary words in a natural format picks it up (Section~\ref{sec:bench}). Quantising the weights to four
bits as well (W4A4) leaves every conclusion in place (Section~\ref{sec:w4a4}). In the appendix,
activation and path patching on three models show that the retrieval damage flows through a few dozen
attention heads, that matched noise flows through the same heads, and that no single head or path
carries it (Appendix~\ref{sec:circuit}). We also tested five ways to repair or explain the damage, with
success criteria fixed in advance: steering those heads, keeping them at eight bits, correcting the
error's gain, protecting loss-sensitive directions, and an attention-margin model. None worked.

\paragraph{Contributions.}
\begin{itemize}
\item We show that perplexity ranks activation quantisers correctly on natural tasks but understates how
much they cost, and that the cost is lost long-tail knowledge: the same facts across quantisers, mostly
rare ones, and removed by any activation perturbation of the same size.
\item Using matched-noise and sign-randomised controls, in a controlled study of \NumModels{} models
and in the deployed setting, we show that the ranking can fail when the error is structured deadzone
deletion rather than noise, and that the input's crest factor flags this regime.
\item We find that block rotation, a natural extension of QuaRot offered by current tooling, is needed
for 10 of 42 popular open models and can leave the input spiky, as on Qwen3-4B. A classical full-width Kronecker Hadamard fixes it at
about the same cost, and halves the closed-book QA loss on Qwen3-4B. Where the block rotation already
leaves the input near Gaussian, the fix changes little.
\end{itemize}

\section{Setup}
\label{sec:artifacts}
\label{sec:theory}
\label{sec:setup}

\begin{figure}[t]
  \centering
  \resizebox{0.72\linewidth}{!}{\definecolor{cnorot}{HTML}{D55E00}  
\definecolor{crot}{HTML}{0072B2}    
\definecolor{cblock}{HTML}{B4589A}  
\definecolor{cgold1}{HTML}{FFE38A}
\definecolor{cgold2}{HTML}{F5A623}
\definecolor{csub}{HTML}{8E44AD}    
\definecolor{cgrey}{HTML}{6B6B6B}
\begin{tikzpicture}[
  font=\fontfamily{phv}\selectfont\scriptsize,
  >={Latex[length=2.6mm,width=2.2mm]},
  arr/.style={->, line width=1.3pt, black},
  fan/.style={->, line width=1.1pt, black, >={Latex[length=2.2mm,width=1.9mm]}},
  sep/.style={line width=1.4pt, dotted, cgrey!70},
  stepc/.style={circle, fill=black, text=white, inner sep=0.6pt, minimum size=3.6mm,
               font=\fontfamily{phv}\selectfont\tiny\bfseries},
  head/.style={font=\fontfamily{phv}\selectfont\footnotesize\bfseries, align=center},
  sub/.style={font=\fontfamily{phv}\selectfont\scriptsize\bfseries, anchor=west},
  txt/.style={anchor=west, align=flush left},
  lbl/.style={font=\fontfamily{phv}\selectfont\scriptsize\bfseries, align=center},
  qb/.style={draw=black, line width=0.8pt, fill=yellow!40, minimum size=4.2mm, inner sep=0pt,
             font=\fontfamily{phv}\selectfont\scriptsize\bfseries},
  hb/.style={draw=crot, line width=0.8pt, fill=crot!15, text=crot, minimum size=4.2mm, inner sep=0pt,
             font=\fontfamily{phv}\selectfont\scriptsize\bfseries},
  lb/.style={draw=black, line width=0.6pt, fill=white, minimum width=9mm, minimum height=4.2mm,
             inner sep=1pt, font=\fontfamily{phv}\selectfont\scriptsize},
]

\draw[sep] (4.55,-2.35) -- (4.55,2.15);

\node[stepc] at (2.2,3.35) {A};
\node[head] at (2.2,2.75) {Quantise the inputs\\of a frozen LLM};

\begin{scope}[shift={(0.35,0.55)}]
  \draw[line width=0.8pt, fill=gray!15, rounded corners=1pt] (-0.38,-0.48) -- (-0.38,0.48) -- (0.22,0.48) -- (0.38,0.32) -- (0.38,-0.48) -- cycle;
  \foreach \y in {0.28,0.12,-0.04,-0.20,-0.36} \draw[line width=0.7pt, gray!70!black] (-0.26,\y) -- (0.24,\y);
\end{scope}
\node[lbl] at (0.35,-0.15) {Text};
\draw[arr] (0.85,0.55) -- (1.35,0.55);

\node[draw=black, line width=1.3pt, left color=cgold1, right color=cgold2, minimum width=1.55cm,
      minimum height=1.15cm, font=\fontfamily{phv}\selectfont\small\bfseries] (llm) at (2.2,0.55) {LLM};
\begin{scope}[shift={(2.85,1.27)}, line width=0.9pt, crot!70]
  \foreach \a in {0,60,120} {\draw (\a:0.17) -- (\a+180:0.17);}
  \foreach \a in {0,60,...,300} {\draw (\a:0.11) -- ++(\a+40:0.06); \draw (\a:0.11) -- ++(\a-40:0.06);}
\end{scope}
\node[lbl] at (2.2,-0.25) {frozen weights};
\draw[arr] (3.0,0.55) -- (3.5,0.55);
\node[lbl, anchor=west] at (3.5,0.55) {$\hat h$};

\draw[->, line width=1.1pt, black, rounded corners=6pt] (2.2,-0.5) -- (2.2,-0.85) -- (1.2,-0.85) -- (1.2,-1.05);
\node[hb] (H) at (0.55,-1.3) {H};
\node[qb] (Q) at (1.2,-1.3) {Q};
\node[lb] (L) at (2.2,-1.3) {linear};
\draw[->, line width=0.8pt] (0.05,-1.3) node[left, font=\fontfamily{phv}\selectfont\scriptsize\bfseries] {$x$} -- (H);
\draw[->, line width=0.8pt] (H) -- (Q); \draw[->, line width=0.8pt] (Q) -- (L);
\node[txt, font=\fontfamily{phv}\selectfont\scriptsize, text width=1.8cm] at (2.72,-1.3)
  {every linear\\input; \textcolor{crot}{H} optional};
\node[font=\fontfamily{phv}\selectfont\scriptsize, cgrey, align=center] at (2.2,-2.0)
  {Part I: one residual block, 12 models\\Part II: all linear inputs, 5 models};

\begin{scope}[shift={(-8.4,0)}]
\node[stepc] at (15.35,3.35) {B};
\node[head] at (15.35,2.75) {The input's crest factor\\sets the regime};
\draw[fan] (13.15,0.35) -- (13.65,1.5);
\draw[fan] (13.15,0.35) -- (13.65,0.3);
\draw[fan] (13.15,0.35) -- (13.65,-0.95);
\foreach \i/\h in {0/0.06,1/0.04,2/0.08,3/0.03,4/0.42,5/0.05,6/0.04} \fill[cnorot] ({13.8+\i*0.09},1.38) rectangle ++(0.06,\h);
\foreach \i/\h in {0/0.07,1/0.12,2/0.09,3/0.30,4/0.10,5/0.13,6/0.08} \fill[cblock] ({13.8+\i*0.09},0.16) rectangle ++(0.06,\h);
\foreach \i/\h in {0/0.08,1/0.17,2/0.26,3/0.31,4/0.25,5/0.16,6/0.07} \fill[crot] ({13.8+\i*0.09},-1.1) rectangle ++(0.06,\h);
\node[sub, cnorot] at (14.5,1.72) {No rotation \ \mdseries crest 7--47};
\node[txt, text width=3.1cm] at (14.5,1.25) {error \textbf{deletes signal}; perplexity can hide lost skills};
\node[sub, cblock] at (14.5,0.5) {Block rotation \ \mdseries crest $\approx$ 7};
\node[txt, text width=3.1cm] at (14.5,0.05) {outliers survive; \textbf{mixed}. Fix: full-width Hadamard};
\node[sub, crot] at (14.5,-0.75) {Full rotation \ \mdseries crest 3.5--3.9};
\node[txt, text width=3.1cm] at (14.5,-1.2) {error \textbf{acts like noise}; perplexity picks a good $\alpha$};
\end{scope}
\end{tikzpicture}}
  \caption{Setup. (A) We quantise either the residual stream at one block (Part~I, 12 models) or the
  input to every linear layer (Part~II, 5 models), with or without a random Hadamard rotation $H$. We
  compare the real error with controls of the same size: its exact parts, matched Gaussian noise, and the
  error with its signs randomised. (B) How spiky the input is, measured by its crest factor, decides the
  regime. Without rotation, a few outliers set the step size, most values fall in the deadzone, and the
  error deletes signal. After a full rotation the input looks Gaussian on the statistics we measure, and
  the error acts much more like noise. A block-diagonal rotation, used when the library has no Hadamard
  matrix of the right width, leaves outliers behind and falls in between.}
  \label{fig:method}
\end{figure}

\paragraph{The quantiser.}
Let $h \in \mathbb{R}^{d}$ be an activation vector the model computes at one position. In Part~I it is
the residual stream at a chosen block; in Part~II it is the input to a linear layer. A quantiser $Q$
maps $h$ to $\hat h = Q(h)$, and we call $\varepsilon = Q(h) - h$ the quantisation error. Our default is
per-token symmetric round-to-nearest (RTN):
\begin{equation}
Q(h)_j = s \cdot \mathrm{clip}\!\left(\mathrm{round}(h_j/s),\, -q_{\max},\, q_{\max}\right),
\qquad s = \frac{\alpha \max_j |h_j|}{q_{\max}}, \qquad q_{\max} = 2^{b-1}-1.
\label{eq:quantiser}
\end{equation}
Here $b$ is the bitwidth and $\alpha \in (0, 1]$ is the clipping ratio. The step size $s$ is set by the
largest value in the token, so a single outlier makes the step large for every other coordinate. With
$\alpha = 1$ nothing is clipped. With $\alpha < 1$, any $|h_j| \ge (q_{\max} + \tfrac12)s$ is set to the
largest level, $\pm q_{\max} s$. (The grid has $2^b - 1$ levels, rounding is half to even, and $s$ is
floored at $10^{-8}$.)

\paragraph{Two ways to measure damage.}
The usual measure is next-token cross-entropy on running text, reported as perplexity. The other is a
\emph{capability probe}: accuracy $C_k(\hat h) \in [0,1]$ on a task $k$ built from random tokens. We write
the damage to that capability as $\Delta C_k = C_k(h) - C_k(\hat h)$. Nothing forces these two measures
to move together, and this paper is about when they do not.

\paragraph{Matched noise: does structure matter, or only size?}
If quantisation error behaved like ordinary random noise, its cost would depend only on how big it is in
each coordinate. A standard second-order approximation of the loss makes this precise:
\begin{equation}
\Delta\mathcal{L} \;\approx\; \sum_{j} F_j\,\varepsilon_j^{2}, \qquad F_j \;=\;
\mathbb{E}\!\left[\left(\frac{\partial\mathcal{L}}{\partial h_j}\right)^{\!2}\right].
  \label{eq:fisher-loss}
\end{equation}
To get it we drop the first-order term (whose expectation is zero for noise independent of the gradient,
but not in general for quantisation error), drop the off-diagonal terms, and replace the Hessian
diagonal with the empirical Fisher (Appendix~\ref{app:theory-full}). Under this approximation, any
substitute with the same per-coordinate size as $\varepsilon$ should do the same damage. We test that
with two substitutes. The first is Gaussian noise with the same per-coordinate power. The second is
\emph{sign-randomised} error, which keeps every $|\varepsilon_j|$ exactly and flips each sign at
random. We report the ratio
\[
\text{noise/real} = \frac{\Delta C_k(\text{substitute})}{\Delta C_k(\text{real error})}.
\]
A ratio of 1 means the substitute does exactly as much damage as the real error. Below 1, the real
error is worse than its size suggests.

\paragraph{Splitting the error into its parts.}
In Part~II we also ask which part of the error does the damage. Each coordinate falls into one of three
cases, and these split the error exactly, $\varepsilon = \varepsilon_{\text{dz}} +
\varepsilon_{\text{grid}} + \varepsilon_{\text{clip}}$:
\begin{itemize}
\item \emph{Deadzone}: values smaller than half a step ($|h_j| < s/2$) round to zero, so the quantiser
deletes them ($\varepsilon_j = -h_j$).
\item \emph{Grid}: every other unclipped value is off by its rounding error.
\item \emph{Clip}: values above $\alpha \max_j|h_j|$ are cut down to the largest level.
\end{itemize}
We then apply each part on its own. The split is exact for the error but not for its effects, because
damage is not additive. On Qwen3-4B's down-projection input, for example, the grid part alone costs more
retrieval than the whole error does (Table~\ref{tab:decomp}). So when we say a part ``accounts for'' a
share of the damage, we mean that applying it alone does that much damage, not that it owns a unique
share. We also split the error a second way: the part along $h$ for each token, which simply shrinks the
token (\emph{gain only}), and everything else (\emph{gain removed}).

\paragraph{Probes.}
We use three in-context tasks built from random tokens, so the model cannot answer from its language
prior \citep{olsson2022induction}. \emph{Induction}: copy the token that followed the current token the
last time it appeared. \emph{At-distance} copying: the same, over a longer span. \emph{Retrieval}: recall
the value paired with a key, with 16 or 64 key--value pairs in context in Part~II (we call these
retrieval16 and retrieval64). Perplexity is measured with the same hooks, so each comparison reads the
same perturbed model in two ways.

\paragraph{Part I: one residual-stream block.}
We study \NumModels{} models from four families: Pythia 70M to 12B \citep{biderman2023pythia}, GPT-2
small \citep{radford2019gpt2}, Gemma-2 2B/9B \citep{team2024gemma2}, and Qwen3 1.7B/8B
\citep{yang2025qwen3}. Every probe and control is run on one model per family (pythia-1.4b, gpt2,
gemma-2-2b, Qwen3-1.7B-Base). The other eight models extend the granularity analysis with a smaller
sweep. A forward hook replaces the residual stream at one block with $\hat h$, at three relative depths,
on 128-token OpenWebText sequences \citep{gokaslan2019openwebtext}. In an end-to-end variant, every block
from that one onward is perturbed. Probes are scored on 256 positions, which gives a worst-case
retrieval half-width of $\pm\HalfwidthRetrieval{}$. The very large activation at the first position
\citep{sun2024massive,xiao2024streaming} stays in the forward pass but is left out of every statistic.
Section~\ref{sec:scale-old} adds Qwen2.5-32B and instruction-tuned Qwen3-8B.

\paragraph{Part II: every linear input, end to end.}
This is the setting that deployed W4A4 methods target. We quantise the input to every linear layer in
every block, per token, with or without a QuaRot random Hadamard rotation \citep{ashkboos2024quarot}
(seed 0) applied before quantisation and undone after. The models are Qwen3-1.7B-Base, Qwen3-4B-Base,
Llama-3.2-3B, Gemma-2-2B and Qwen3-8B-Base, run in fp16. Gemma's widths ($2304 = 36 \cdot 64$ and $9216 =
36 \cdot 256$) use QuaRot's order-36 Hadamard matrix, and Gemma is scored with a beginning-of-text token
at the start of every window.

\paragraph{Why some widths get rotated in blocks.}
QuaRot builds a Hadamard matrix of width $K \cdot 2^m$, where $K$ must come from a short list of stored
orders. Its reference code rejects any other width.\footnote{\texttt{get\_hadK} in
\texttt{fake\_quant/hadamard\_utils.py} of the QuaRot repository \citep{ashkboos2024quarot}.} In our
models, only one input has such a width: Qwen3-4B's down-projection input, $9728 = 19 \cdot 512$. The
cheapest orthogonal workaround, and the one we implement, is a block-diagonal Hadamard with the largest
valid block. For 9728 that means 19 independent blocks of 512.

Deployed tooling offers the same option with smaller blocks. llm-compressor's SpinQuant and QuIP
rotations take a block size, and its examples set it to 128 because smaller Hadamards are cheaper at run
time. The down-projection rotation (SpinQuant's $R_4$, QuIP's input rotation) is then applied to the
activation online, in 128-wide blocks, before the activation is quantised.\footnote{llm-compressor
\citep{llmcompressor}, commit \texttt{eac6972}: \texttt{transform\_block\_size=128} in
\texttt{examples/transform/\{spinquant,quip,nvfp4\_quip\_gptq,spinquant\_cohere2\_moe\}\_example.py};
$R_4$ is \texttt{location="input"} on \texttt{down\_proj} (\texttt{SpinQuantModifier.\_create\_r4\_scheme}).
compressed-tensors \citep{compressedtensors} registers it as a forward pre-hook that applies the
$128 \times 128$ Hadamard block-diagonally (\texttt{transform/factory/base.py},
\texttt{transform/utils/matrix.py}); activation quantisation runs inside the module's forward, after the
hook.} Most of those examples quantise only the weights (W4A16). The one that also quantises activations
uses NVFP4, which shares a scale across every 16 values rather than across a token. None of them is the
per-token INT4 setting we study, and we do not claim our results carry over to them.

A Hadamard matrix of order 9728 does exist. Since $9728 = 76 \cdot 128$, Paley's second construction gives
a matrix of order $2(q+1) = 76$ for the prime $q = 37 \equiv 1 \pmod 4$, and a Kronecker product of
Hadamard matrices is again Hadamard. So $H_{9728} = H_{76} \otimes H_{128}$. Section~\ref{sec:partial}
compares this full-width rotation with the block and dense ones.

\paragraph{Statistics and perplexity.}
Probes use 2000 items per seed (1000 for the error-split runs of Section~\ref{sec:norot}). Every
comparison is paired over items, with a 95\% bootstrap confidence interval from 10k resamples. These
intervals only cover which items were sampled. We report variation across rotation seeds per seed or as
a range over three seeds, and we never pool variation across models into an interval. Perplexity is
measured on WikiText-2 at context 2048, and every model load is checked against a recorded
full-precision value before any run. When perplexity appears next to probe accuracy
(Tables~\ref{tab:k1}--\ref{tab:decomp}), it uses the same 40 windows as the probe runs. The benchmark
section uses the full test set, so the same setting has a different value there (Qwen3-4B: 7.35 against
7.90 at full precision, and 11.0 against 11.8 at A4 with $\alpha = 1$).

\section{Perplexity Ranks Quantisers but Understates What They Remove}
\label{sec:natural}

We start with the setting people actually deploy and with natural tasks, and turn to mechanism later.
Every linear-layer input is quantised to four bits per token after a QuaRot Hadamard rotation (Part~II,
Section~\ref{sec:setup}). On each model we compare:
\begin{itemize}
\item the clipping ratio that minimises perplexity (``tuned''), and $\alpha = 1$;
\item two deliberately weaker quantisers: a low clipping ratio, and stochastic rounding;
\item an unrotated model at six bits (at four bits without rotation, these models are unusable);
\item for Qwen3-4B, a full-width Hadamard on the one input QuaRot only rotates in blocks
(Section~\ref{sec:partial}).
\end{itemize}
W4 rows add 4-bit GPTQ weights (Section~\ref{sec:w4a4}). Matched-noise rows replace the activation error
with Gaussian noise of the same per-token power. Standard benchmarks (commonsense, MMLU, in-context recall
and RULER) are in Section~\ref{sec:bench}.

\subsection{Natural question answering}
\label{sec:qa}

\begin{figure}[t]
  \centering
  \includegraphics[width=\linewidth]{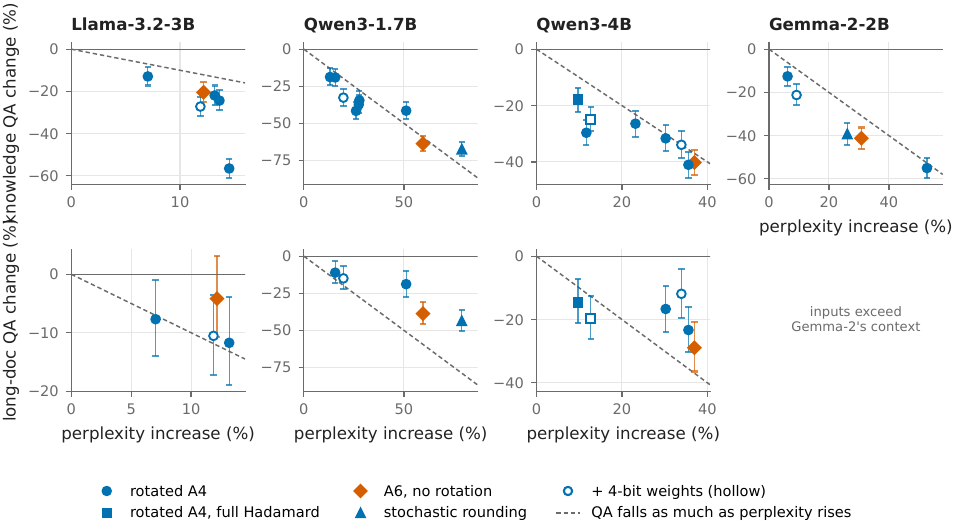}
  \caption{How much QA drops against how much perplexity rises, both relative to full precision
  (Table~\ref{tab:qa}). Top: closed-book QA (mean relative change on TriviaQA and NQ-Open). Bottom:
  LongBench. Bars are 95\% paired bootstrap intervals. On the dashed line, QA falls by the same fraction
  that perplexity rises. Most closed-book points are on or below it, so QA drops by at least as large a
  fraction as perplexity rises, and up to about twice as large. Unlike Figure~\ref{fig:teaser}, both axes
  here are relative, so that the two share a scale.}
  \label{fig:qa-ppl}
\end{figure}

We use tasks that need either knowledge stored in the weights or reading over long documents.
TriviaQA and NQ-Open are closed-book, five-shot and scored by exact match, with 1000 questions each. CoQA
and DROP are scored by F1, with 500 items each. From LongBench we use six QA tasks (NarrativeQA, Qasper,
MultiFieldQA, HotpotQA, 2WikiMQA and MuSiQue), with 100 items each and inputs up to 8k tokens, and
report their average. Table~\ref{tab:qa} gives every setting as a paired difference from full precision,
and Figure~\ref{fig:qa-ppl} plots the relative QA change against the relative perplexity change.

\begin{table}[t]
\centering
\small
\setlength{\tabcolsep}{4pt}
\caption{Natural question answering. The FP row gives the score; other rows give the paired difference from FP, bold where the 95\% bootstrap interval excludes zero. TriviaQA and NQ-Open: exact match, 1000 questions each, five-shot; CoQA and DROP: F1, 500 items; LongBench: mean F1 over six single- and multi-document QA tasks (100 items each, inputs up to 8k tokens). Perplexity is full WikiText-2. W4 rows add GPTQ INT4 weights in the rotated basis (Section~\ref{sec:w4a4}). Gemma-2-2B's LongBench inputs exceed its context and it scores near zero at FP, so we omit that column; Llama-3.2-3B and Gemma-2-2B score below 0.1 on DROP at FP, so their DROP differences carry little information. All rows use rotation seed 0; three-seed means for the Qwen3-4B block and full-width rows are in Section~\ref{sec:partial}.}
\label{tab:qa}
\begin{tabular}{l l r rrrrr}
\toprule
Model & Setting & Perplexity & TriviaQA & NQ & CoQA & DROP & LongBench \\
\midrule
Llama-3.2-3B & FP & 7.81 & 0.562 & 0.215 & 0.788 & 0.063 & 0.288 \\
 & tuned ($\alpha=0.70$) & 8.36 & \textbf{-0.059} & \textbf{-0.033} & \textbf{-0.021} & -0.003 & \textbf{-0.022} \\
 & $\alpha=1$ & 8.84 & \textbf{-0.116} & \textbf{-0.050} & \textbf{-0.042} & +0.008 & \textbf{-0.034} \\
 & matched Gaussian ($\alpha=1$) & 8.94 & \textbf{-0.361} & \textbf{-0.105} & -- & -- & -- \\
 & A6, no rot. & 8.76 & \textbf{-0.112} & \textbf{-0.045} & \textbf{-0.020} & -0.001 & -0.012 \\
 & W4 + tuned ($\alpha=0.70$) & 8.74 & \textbf{-0.133} & \textbf{-0.066} & \textbf{-0.032} & \textbf{-0.015} & \textbf{-0.030} \\
\midrule
Qwen3-1.7B & FP & 9.39 & 0.355 & 0.144 & 0.797 & 0.179 & 0.297 \\
 & tuned ($\alpha=0.70$) & 10.86 & \textbf{-0.059} & \textbf{-0.031} & \textbf{-0.029} & -0.025 & \textbf{-0.033} \\
 & tuned ($\alpha=0.70$), learned rot. & 10.62 & \textbf{-0.047} & \textbf{-0.035} & -- & -- & -- \\
 & $\alpha=1$ & 11.99 & \textbf{-0.107} & \textbf{-0.056} & -- & -- & -- \\
 & matched Gaussian ($\alpha=1$) & 11.84 & \textbf{-0.140} & \textbf{-0.063} & -- & -- & -- \\
 & $\alpha=0.5$ & 14.19 & \textbf{-0.117} & \textbf{-0.072} & \textbf{-0.057} & \textbf{-0.061} & \textbf{-0.056} \\
 & stochastic & 16.80 & \textbf{-0.225} & \textbf{-0.103} & \textbf{-0.177} & \textbf{-0.090} & \textbf{-0.130} \\
 & A6, no rot. & 14.98 & \textbf{-0.230} & \textbf{-0.090} & \textbf{-0.149} & \textbf{-0.081} & \textbf{-0.115} \\
 & W4 + tuned ($\alpha=0.70$) & 11.25 & \textbf{-0.094} & \textbf{-0.056} & \textbf{-0.043} & \textbf{-0.064} & \textbf{-0.044} \\
\midrule
Qwen3-4B & FP & 7.90 & 0.473 & 0.224 & 0.835 & 0.318 & 0.343 \\
 & tuned ($\alpha=0.60$) & 10.29 & \textbf{-0.124} & \textbf{-0.083} & \textbf{-0.050} & \textbf{-0.096} & \textbf{-0.057} \\
 & tuned ($\alpha=0.60$), learned rot. & 9.73 & \textbf{-0.098} & \textbf{-0.072} & -- & -- & -- \\
 & full Had., $\alpha=0.70$ & 8.67 & \textbf{-0.062} & \textbf{-0.051} & -0.008 & \textbf{-0.054} & \textbf{-0.050} \\
 & $\alpha=0.5$ & 10.71 & \textbf{-0.179} & \textbf{-0.099} & \textbf{-0.043} & \textbf{-0.148} & \textbf{-0.080} \\
 & A6, no rot. & 10.82 & \textbf{-0.150} & \textbf{-0.109} & \textbf{-0.055} & \textbf{-0.153} & \textbf{-0.099} \\
 & W4 + tuned ($\alpha=0.60$) & 10.58 & \textbf{-0.133} & \textbf{-0.089} & \textbf{-0.063} & -0.023 & \textbf{-0.041} \\
 & W4 + full Had., $\alpha=0.70$ & 8.90 & \textbf{-0.088} & \textbf{-0.070} & \textbf{-0.020} & +0.009 & \textbf{-0.067} \\
\midrule
Gemma-2-2B & FP & 8.73 & 0.534 & 0.180 & 0.773 & 0.081 & -- \\
 & tuned ($\alpha=0.70$) & 9.27 & \textbf{-0.075} & \textbf{-0.020} & -0.007 & \textbf{-0.017} & -- \\
 & $\alpha=0.4$ & 13.33 & \textbf{-0.282} & \textbf{-0.103} & \textbf{-0.104} & -0.020 & -- \\
 & stochastic & 11.01 & \textbf{-0.188} & \textbf{-0.078} & \textbf{-0.057} & -0.014 & -- \\
 & A6, no rot. & 11.42 & \textbf{-0.212} & \textbf{-0.077} & \textbf{-0.067} & \textbf{-0.023} & -- \\
 & W4 + tuned ($\alpha=0.70$) & 9.53 & \textbf{-0.093} & \textbf{-0.045} & \textbf{-0.024} & -0.015 & -- \\
\bottomrule
\end{tabular}
\end{table}

\paragraph{QA drops by more than perplexity rises.}
Every quantised setting loses closed-book QA on every model, even the best-tuned ones. Take the clipping
ratio that perplexity picks. On Llama-3.2-3B, Qwen3-1.7B and Gemma-2-2B, perplexity rises by 6--16\%,
while TriviaQA accuracy drops by 10--17\% and NQ-Open by 11--22\% (relative). On Qwen3-4B with the
full-width Hadamard, perplexity rises by 10\% and NQ-Open drops by 23\%. LongBench drops by 8--15\% under
the same settings. CoQA, where the answer is in a short passage, drops by at most 6\%. A perplexity cost
that most people would accept thus comes with a loss of stored knowledge that is at least as large in
relative terms, and up to about twice as large (Figure~\ref{fig:qa-ppl}).

\paragraph{But perplexity gets the order right.}
Within one model, does the setting with the larger perplexity increase also lose more QA? We check every
pair among the 18 settings that have both natural-task and probe results. (We leave out the head-mixing
settings of Appendix~\ref{sec:circuit}, which were chosen using a probe.) Perplexity orders 33 of 34 pairs
correctly on closed-book QA (0.97, with a 95\% bootstrap interval over settings of [0.85, 1.00]), 21 of 23
on LongBench and 27 of 29 on CoQA. Retrieval16 accuracy does worse, ordering 0.79, 0.74 and 0.72 of the
pairs correctly, and induction gets 0.88, 0.91 and 0.83. So within a model, perplexity ranks quantisers
by their damage to natural tasks at least as well as our probes do, and better than random-token
retrieval. What it gets wrong is the size of the loss. The retrieval probe can understate the size too:
on Qwen3-4B's tuned block-rotated model, retrieval16 drops by 4\% while closed-book QA drops by 32\%.

\paragraph{The full-width rotation helps here too.}
On Qwen3-4B, the full-width Hadamard halves the TriviaQA loss of the tuned block-rotated model ($-0.062$
against $-0.124$ for rotation seed 0, and $-0.060$ against $-0.112$ over three seeds). It also shrinks the
NQ-Open and LongBench losses, as it did for recall (Section~\ref{sec:bench}). Tuning the clipping ratio
matters as well: on Llama-3.2-3B, leaving it at $\alpha = 1$ doubles the TriviaQA loss ($-0.116$ against
$-0.059$ when tuned).

\paragraph{At equal perplexity, dropping rotation costs QA on one of four models.}
We compare the unrotated A6 model with rotated settings of similar perplexity, as in
Section~\ref{sec:bench}. On Qwen3-1.7B the unrotated model does worse than the rotated $\alpha = 0.5$
model, whose perplexity is only slightly lower (14.19 against 14.98). It loses more on TriviaQA
($-0.113$ [$-0.138$, $-0.087$]), CoQA ($-0.092$ [$-0.127$, $-0.060$]) and LongBench ($-0.059$ [$-0.084$,
$-0.034$]). It is level with stochastic rounding, whose perplexity is much higher (16.80). On
Llama-3.2-3B, Qwen3-4B and Gemma-2-2B, the unrotated model and its rotated neighbours are level within
the intervals, or their order follows perplexity. Qwen3-1.7B is also the model with the clearest recall
gap. So on natural tasks, the structured error of an unrotated input costs more than its perplexity
suggests on one model out of four.

\subsection{Which questions are lost}
\label{sec:longtail}

\begin{figure}[t]
  \centering
  \includegraphics[width=\linewidth]{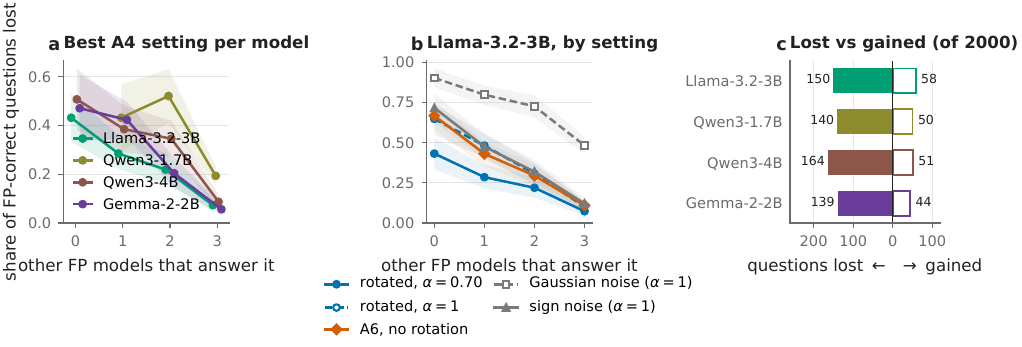}
  \caption{Which closed-book questions quantisation removes (TriviaQA and NQ-Open, 2000 questions).
  (a) Of the questions a model gets right at full precision, the share its best A4 setting loses, by
  how many of the other three full-precision models also get the question right. Bands are 95\%
  bootstrap intervals. Questions that fewer models know are lost far more often. (b) The same for several
  settings of Llama-3.2-3B. (c) Questions lost and gained by the best A4 setting.}
  \label{fig:longtail}
\end{figure}

Every model answers the same 2000 TriviaQA and NQ-Open questions, so we can look at which questions
quantisation takes away (Figure~\ref{fig:longtail}).

\paragraph{Knowledge is removed, not reshuffled.}
Under its best A4 setting, each model loses 139--164 of the questions it got right at full precision,
and gains only 44--58 it got wrong (Figure~\ref{fig:longtail}c). For the harsher settings (no rotation,
stochastic rounding, low clipping ratios), losses outnumber gains by 3 to 17 times. If quantisation only
made answers noisier, gains and losses would be closer in number.

\paragraph{Different quantisers lose the same questions.}
Two settings of the same model lose overlapping sets of questions. The overlap is 1.1--2.6 times what
we would expect if the losses were independent, and it is largest between settings with similar
perplexity (2.4--2.6 times among the full-Hadamard variants of Qwen3-4B, for example). Which questions are
fragile is a property of what the model knows, not of a particular quantiser.

\paragraph{Facts that fewer models know are lost first.}
As a rough measure of how firmly a fact is stored, we count how many of the other three full-precision
models answer the question. Of the questions a model gets right, its best A4 setting loses 43--51\% of
those no other model answers, but only 6--9\% of those all three others answer
(Figure~\ref{fig:longtail}a). Qwen3-1.7B, which answers the fewest, goes from 43\% to 19\%. In every
setting we ran, the questions all three other models know are the ones lost least
(Figure~\ref{fig:longtail}b). Weakly stored, long-tail knowledge is what quantisation removes first. It
is also exactly what perplexity on running text, which is dominated by common patterns, measures least.

\paragraph{This is not just the model's own confidence.}
A fact that other models do not know might simply be one this model was unsure of anyway. If so, ``how
many other models know it'' would just stand in for the model's own confidence. To check, we score how
confident each full-precision model is in the gold answer: its mean per-token log-probability (best of
the first five aliases) after a fixed five-shot prompt in lm-eval's format. This confidence is
informative. It separates the questions the model gets right from those it gets wrong with an AUC of
0.70--0.75. But within each quartile of confidence (per dataset, pooled over the four models), the best
A4 setting still loses 38--56\% of the answered questions no other model knows, against 9--12\% of those
all three know.

We also fit a logistic regression of ``lost'' on both quantities, with model and dataset fixed effects
and standard errors clustered by question (each question appears once per model; 1015 clusters). Each
additional model that knows the answer multiplies the odds of losing it by 0.46 (coefficient $-0.78$,
standard error 0.05). Once that is known, the model's own confidence adds little ($-0.06$, standard error
0.05). Clustering changes the standard errors by less than 0.005. Fitted per model, the coefficient on
the number of other models ranges from $-0.66$ to $-1.00$, each at least five standard errors from zero.
How widely a fact is known captures something the model's own answer probability does not. This fits the
idea that it reflects how robustly the fact is stored, rather than how confidently it is stated.
Qwen3-1.7B has few questions that no other model knows (18), so its per-quartile rates are noisy.

\paragraph{Gaussian noise removes more knowledge than the real error, and sign noise the same.}
A question the model barely knows could be flipped by any perturbation, so we ran the matched-noise
controls on closed-book QA too. On Llama-3.2-3B, Gaussian noise matched to the $\alpha = 1$ error gives
about the same perplexity (8.94 against 8.84). Yet it loses 506 of the 777 questions the full-precision
model gets right, against 227 for the real error. Accuracy is 0.150 [0.131, 0.170] lower than under the
real error, with the same pattern over how widely a fact is known (Figure~\ref{fig:longtail}b). On
Qwen3-1.7B the difference points the same way but is small (247 against 211 lost; 0.020 [0.006,
0.035]). So the long tail is where any perturbation of the activations does its damage. And for stored
knowledge, the rotated real error is \emph{less} harmful than Gaussian noise of the same power, the
opposite of what we find for unrotated inputs in Part~I.

\begin{figure}[t]
  \centering
  \includegraphics[width=0.78\linewidth]{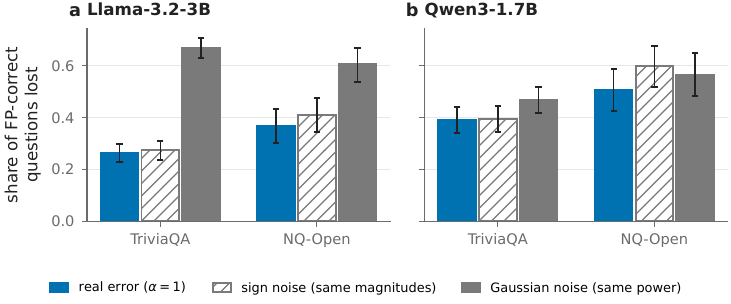}
  \caption{Matched-noise controls on closed-book QA (1000 questions per dataset). Share of the questions
  the full-precision model gets right that are lost under the rotated A4 error at $\alpha = 1$, under sign
  noise (the real error's per-coordinate sizes with random signs), and under Gaussian noise of the same
  per-channel power. Error bars are 95\% bootstrap intervals. Sign noise matches the real error on both
  models and datasets (every difference within $\pm 0.01$, none significant). Gaussian noise loses more,
  by a wide margin on Llama-3.2-3B, at nearly the same perplexity (8.94 against 8.84).}
  \label{fig:qa-noise}
\end{figure}

The sign-randomised control shows where the difference comes from. It keeps each coordinate's error size
and only randomises the signs, and it does as much damage as the real error, not more: 242 questions lost
on Llama-3.2-3B ($-0.005$ [$-0.021$, 0.010] against the real error) and 226 on Qwen3-1.7B ($-0.004$
[$-0.018$, 0.012]). What spares stored knowledge is therefore how the real error's size is spread across
coordinates, which Gaussian noise of the same total power does not copy. The sign pattern, which
mattered for unrotated inputs (Section~\ref{sec:ladder}), does not matter here (Figure~\ref{fig:qa-noise}).
This comparison is also a direct example of perplexity understating knowledge loss. On Llama-3.2-3B the
Gaussian noise and the real error differ by 0.1 in perplexity, yet the noise loses more than twice as many
of the questions the full-precision model answers.

\section{A Controlled Study: Structured Error Is Not Noise}
\label{sec:mask}
\label{sec:noise}

Part~I quantises the residual stream at a single block of \NumModels{} models and compares the
quantiser with perturbations of exactly the same size (Section~\ref{sec:setup}). Its job here is to show
that the \emph{shape} of the quantisation error, not only its size, decides what breaks. We keep the two
results the rest of the paper relies on. Calibration, granularity, sensitivity, feature-level analysis
and the loss-aware repair are in Appendix~\ref{app:part1}.

\subsection{Same size, different damage}
\label{sec:induction}

Table~\ref{tab:probes} gives the first comparison. At three bits, quantisation keeps
\TableProbeThreeBitQuant{} of induction accuracy. Gaussian noise of the same size per channel keeps
\TableProbeThreeBitNoise{}. Randomising just the signs of the quantisation error gives the noise result,
not the quantised one. Retrieval is the most fragile probe throughout, though it also starts lower and
varies more at full precision (0.27 to 0.46, against 0.95 to 0.99 for the copying probes). Two
perturbations of the same size, applied to the same activation, leave different abilities intact.

\begin{table}[t]
  \centering
  \small
  \caption{Part~I. Fraction of accuracy retained on each probe, median over model families at
  three relative depths. Intact accuracy is 0.954--0.987 for the copying probes and 0.27--0.46 for
  retrieval, scored on 256 positions.}
  \label{tab:probes}
  \begin{tabular}{lrrr}
    \toprule
    Perturbation & Induction & At-distance & Retrieval \\
    \midrule
    \TableProbeBody
    \bottomrule
  \end{tabular}
\end{table}

Perplexity orders these perturbations the same way the probes do, but it does not show how large the
damage is. Where perplexity has risen by only 1.2 to 1.5 times, induction still keeps
\MildBandInduction{} of its accuracy, while retrieval has fallen to \MildBandRetrieval{}
(Appendix~\ref{sec:calibration}). This contradicts the size-only account of
Equation~\ref{eq:fisher-loss}.

\subsection{The sign pattern carries the damage}
\label{sec:ladder}

To find what about the error matters, we replace the Gaussian noise with substitutes that copy more and
more of the real error (Table~\ref{tab:ladder}, Appendix~\ref{app:ladder-table}), and report how much of
quantisation's extra perplexity each one reproduces. The sign-randomised substitute keeps the size of
every error value, $\lvert\varepsilon_{ij}\rvert$, for every coordinate of every token. Even so, it
reproduces as little of the damage as Gaussian noise does (0.164 against 0.173 at four bits, and 0.001
against 0.001 at two bits). This is what classical quantisation theory predicts
\citep{bennett1948spectra}, since the substitute's magnitudes are still a fixed function of $h$. Moving
each error to a different token is similarly harmless. What is left is the joint pattern of signs across
coordinates. If we resample a growing fraction of the signs, from 0 to 100 percent
(Figure~\ref{fig:signdial}), both probes recover smoothly and steadily until they match the Gaussian
control.

Sparse-dictionary features show which way the change goes. Quantisation pushes features down, because
values below half a step round to zero (the \emph{deadzone}), while matched noise pushes them up
(Appendix~\ref{app:part1}). How organised the sign pattern is depends on how widely the quantiser shares
its scale. At the same bitwidth, the ratio of quantised to matched-noise perplexity differs by orders of
magnitude between per-token and per-tensor scaling (Appendix~\ref{sec:premium}). Quantising in a randomly
rotated frame keeps induction at 0.980 at three average bits. That result motivates the deployed, rotated
setting of Part~II.

\section{Two Regimes for Linear-Input Quantisation}
\label{sec:regimes}

Part~I perturbs one residual-stream activation at a time. Deployed methods quantise every linear layer's
input end to end, and many of them rotate first. We now repeat the main comparisons in that setting
(Part~II, Section~\ref{sec:setup}).

\subsection{Linear inputs are outlier-heavy, and rotation makes them Gaussian}
\label{sec:crest}

With per-token quantisation the step size is $\alpha\max_j|h_j|/(2^{b-1}-1)$. The error relative to the
signal therefore grows with the square of the \emph{crest factor}, $\max_j|h_j| / \mathrm{rms}(h)$: the
spikier the input, the coarser the grid for everything else. Table~\ref{tab:crest} reports the crest
factor for every type of linear input. Unrotated inputs have mean crest factors of 7.5--47.2 and excess
kurtosis up to 1138. The down-projection input is the most extreme in every model, and Gemma-2-2B's is
the most extreme of all. After a full Hadamard rotation, every input has a crest factor of 3.5--3.9 and
excess kurtosis within 0.16 of zero. That matches a Gaussian vector of the same width (3.6--4.0), and
the relative 4-bit error power is 2.1--2.6\%. These are statistics of single coordinates within a token.
They do not show that the rotated input is jointly Gaussian, or that its quantisation error is
independent of it. Sections~\ref{sec:rot} and \ref{sec:discussion} show where the error still behaves
differently from noise.

The crest factor is a stand-in for what matters directly: how much of the input the deadzone deletes.
We measure that too, on the same retrieval16 inputs as Table~\ref{tab:crest}. Without rotation, 93--98\%
of the down-projection input's values round to zero at 4 bits, and they carry 19--31\% of its energy.
After a full rotation, 19--22\% of the values round to zero, carrying only 0.5\% of the energy. The
block-rotated input sits in between, with 49\% of its values and 4.0\% of its energy deleted (2.57\% on
WikiText-2 tokens; Section~\ref{sec:partial-q25}). All three measures order the regimes the same way.
None of them, at 4 or 6 bits, orders the unrotated models by the benchmark gap of
Section~\ref{sec:bench}. Qwen3-1.7B shows that gap most clearly but deletes the least energy (19\%),
while Llama-3.2-3B shows no gap and deletes 31\%. The single exception is Qwen3-4B's block-rotated
down-projection input (Section~\ref{sec:partial}).

\begin{figure}[t]
  \centering
  \includegraphics[width=\linewidth]{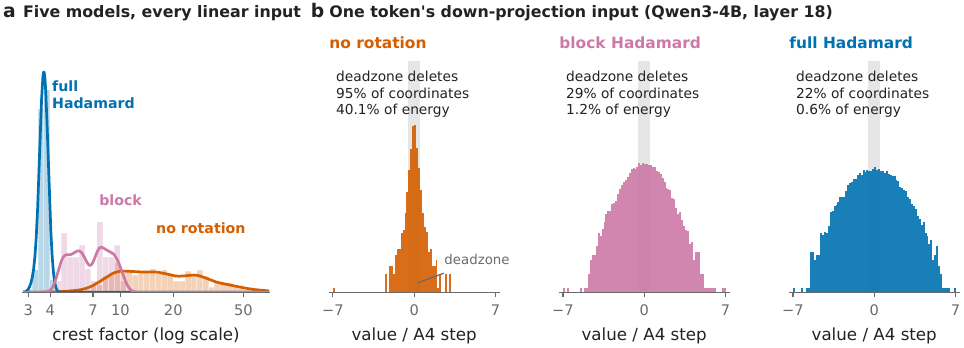}
  \caption{Rotation makes linear inputs Gaussian, except where it is only block-diagonal. (a) Crest
  factor of every linear input (one point per model, layer and input type; q/k/v share an input) in all
  five models, on retrieval16 inputs. Without rotation it ranges from about 5 to over 50. After a full
  Hadamard rotation it sits at the Gaussian value. (b) One token's down-projection input on Qwen3-4B
  (layer 18, the median-crest token of those stored; WikiText-2), in units of the A4 step ($\alpha = 1$):
  without rotation, after the block Hadamard rotation, and after the full-width $H_{76} \otimes H_{128}$
  rotation (log count axis). The grey band is the deadzone, $|x| < \Delta/2$. The text gives the share of
  values and of energy that rounding deletes there.}
  \label{fig:crest}
\end{figure}

Figure~\ref{fig:crest} shows the whole picture. Without rotation, the crest factor spreads over an order
of magnitude. After a full rotation it collapses onto the Gaussian value. The block-rotated input sits
in between. For a single example token, the unrotated input puts 95\% of its values and 40\% of its
energy in the A4 deadzone, the block-rotated input 1.2\% of its energy, and the fully rotated input 0.6\%.

\begin{table}[t]
  \centering
  \small
  \setlength{\tabcolsep}{4pt}
  \caption{Part~II. Per-token crest factor (mean over tokens and layers, sink excluded) and excess
  kurtosis of each linear input type, on retrieval16 inputs, without rotation and after the QuaRot
  Hadamard rotation. q covers the q/k/v projections, gate/up the MLP input. A Gaussian vector of the
  same widths has crest factor 3.6--4.0. $^\dagger$Block-diagonal rotation (QuaRot's construction has
  no order-9728 matrix); a dense rotation gives crest 4.01, kurtosis 0.00, and the full-width
  Hadamard $H_{76} \otimes H_{128}$ crest 3.85, kurtosis $-0.10$. Excess kurtosis of the down-projection input is 222--1138 raw and $-0.25$ to $-0.06$ rotated (Qwen3-4B$^\dagger$: 5.86). Full table with p99 and error power in
  Appendix~\ref{app:part2}.}
  \label{tab:crest}
  \begin{tabular}{l rr rr rr rr}
    \toprule
    & \multicolumn{2}{c}{q/k/v} & \multicolumn{2}{c}{o} & \multicolumn{2}{c}{gate/up} & \multicolumn{2}{c}{down} \\
    \cmidrule(lr){2-3}\cmidrule(lr){4-5}\cmidrule(lr){6-7}\cmidrule(lr){8-9}
    Model & raw & rot. & raw & rot. & raw & rot. & raw & rot. \\
    \midrule
    Llama-3.2-3B    & 12.0 & 3.71 & 10.8 & 3.70 & 7.5 & 3.73 & 27.2 & 3.88 \\
    Qwen3-1.7B-Base & 19.7 & 3.46 & 9.6 & 3.58 & 15.7 & 3.55 & 28.3 & 3.73 \\
    Qwen3-4B-Base   & 19.7 & 3.50 & 10.3 & 3.77 & 16.0 & 3.61 & 33.1 & 6.98$^\dagger$ \\
    Qwen3-8B-Base   & 25.8 & 3.61 & 10.3 & 3.78 & 21.6 & 3.65 & 38.6 & 3.87 \\
    Gemma-2-2B      & 17.3 & 3.52 & 9.9 & 3.59 & 13.1 & 3.62 & 47.2 & 3.65 \\
    \bottomrule
  \end{tabular}
\end{table}

\subsection{Without rotation: the deadzone deletes signal}
\label{sec:norot}

\begin{figure}[t]
  \centering
  \includegraphics[width=\linewidth]{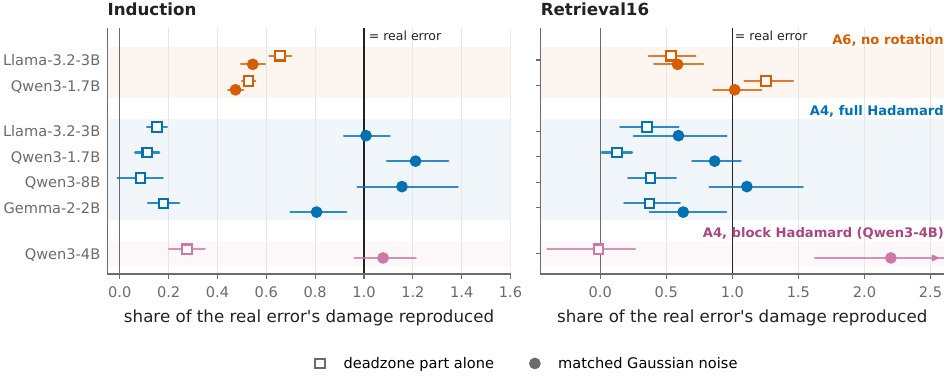}
  \caption{What reproduces the real error's damage. Each row is one model and setting. The marker is
  the share of the real error's damage that a substitute reproduces (1 = all of it), with a 95\% paired
  bootstrap interval. Without rotation (orange), the deadzone part alone does half or more of the
  retrieval damage, and matched noise falls well short on induction. After a full rotation (blue), the
  deadzone part does little, and matched noise does about as much induction damage as the real error and
  most of its retrieval damage. Qwen3-4B's block-rotated input (pink) breaks the pattern on retrieval
  (Section~\ref{sec:partial}). A4 without rotation is left out because every arm is at the floor.}
  \label{fig:controls}
\end{figure}

Table~\ref{tab:k1} applies each part of the error on its own, and Figure~\ref{fig:controls} plots the two
comparisons that matter most.

Without rotation, the deadzone part on its own reproduces all of the real error's retrieval damage at 4
bits (1.02 on Llama, 1.00 on Qwen3-1.7B), and half or more at 6 bits (0.54 and 1.25), at a perplexity
close to the full error's. The error is mostly signal being deleted, and its size does not predict its
effect. At 6 bits, matched Gaussian noise reproduces only 0.54 and 0.47 of the induction damage, and the
sign-randomised substitute 0.65 and 0.60. This repeats the single-block result of
Section~\ref{sec:noise} in the deployed setting. At 4 bits without rotation the model has collapsed
(perplexity 142 and 35{,}162), and every arm sits at the floor, the same floor effect as in the
end-to-end variant of Section~\ref{sec:alllayer}. Retrieval alone does not separate the regimes: on
Llama, matched noise reproduces 0.50--0.59 of the retrieval damage both with and without rotation.

\begin{table}[t]
  \centering
  \small
  \setlength{\tabcolsep}{3.2pt}
  \caption{Part~II. Share of the real error's damage reproduced by each component or control
  ($\Delta C$ of the arm over $\Delta C$ of the real error, 95\% paired bootstrap CI; seed 0,
  1000 items). 1 = as much damage as the real error. A6 + Hadamard is omitted: its damage is
  within noise of zero, so the ratios are undefined. Qwen3-4B's rotated row is dominated by its
  block-rotated down-projection (Section~\ref{sec:partial}). Perplexity is on the 40 WikiText-2 windows of the probe runs (Section~\ref{sec:setup}).}
  \label{tab:k1}
  \begin{tabular}{ll rrrr rr}
    \toprule
    & & \multicolumn{4}{c}{Induction} & \multicolumn{2}{c}{Retrieval16} \\
    \cmidrule(lr){3-6}\cmidrule(lr){7-8}
    Setting & Model & dz only & grid only & Gaussian & sign & dz only & Gaussian \\
    \midrule
    A6, no rot. & Llama-3.2-3B & 0.66 & 0.27 & 0.54 [0.50,0.59] & 0.65 [0.60,0.71] & 0.54 [0.37,0.72] & 0.59 \\
                & Qwen3-1.7B   & 0.53 & 0.27 & 0.47 [0.44,0.51] & 0.60 [0.57,0.64] & 1.25 [1.09,1.46] & 1.02 \\
    A4, no rot. & Llama-3.2-3B & 0.99 & 0.29 & 1.01 & 1.00 & 1.02 [1.00,1.04] & 1.04 \\
                & Qwen3-1.7B   & 1.00 & 0.41 & 1.00 & 1.00 & 1.00 [1.00,1.01] & 1.00 \\
    \midrule
    A4 + Hadamard & Llama-3.2-3B & 0.15 & 0.72 & 1.01 [0.92,1.11] & 1.05 [0.96,1.14] & 0.36 [0.15,0.59] & 0.59 \\
                  & Qwen3-1.7B   & 0.11 & 0.71 & 1.21 [1.09,1.35] & 1.11 [1.00,1.22] & 0.13 [0.01,0.24] & 0.87 \\
                  & Qwen3-4B     & 0.28 & 0.63 & 1.08 [0.96,1.21] & 0.91 [0.80,1.04] & $-0.01$ [$-0.40$,0.26] & 2.20 \\
                  & Qwen3-8B     & 0.08 & 0.71 & 1.16 [0.97,1.38] & 1.15 [0.97,1.37] & 0.38 [0.21,0.57] & 1.11 \\
                  & Gemma-2-2B   & 0.18 & 0.60 & 0.81 [0.70,0.93] & 0.87 [0.75,0.99] & 0.37 [0.18,0.60] & 0.63 \\
    \bottomrule
  \end{tabular}
\end{table}

\subsection{With full rotation: the error is closer to noise, and its size matters most}
\label{sec:rot}

After rotation the picture flips (Table~\ref{tab:k1}, lower block). The grid part now carries most of
the damage (0.60--0.72 of induction), the deadzone at most 0.38 of retrieval, and matched noise
reproduces most or all of the induction damage (0.81--1.21). Gemma-2-2B is the one model where noise
falls short of the real error on induction (0.81 [0.70, 0.93]). On retrieval, matched noise reproduces
only 0.59 of the damage on Llama-3.2-3B and 0.63 on Gemma-2-2B, so some structure that matters survives
rotation. The error is more noise-like on these measures, but it is not the same as independent noise.
Three more results show that in this regime, the size of the error matters more than its bias.

\begin{figure}[t]
  \centering
  \includegraphics[width=\linewidth]{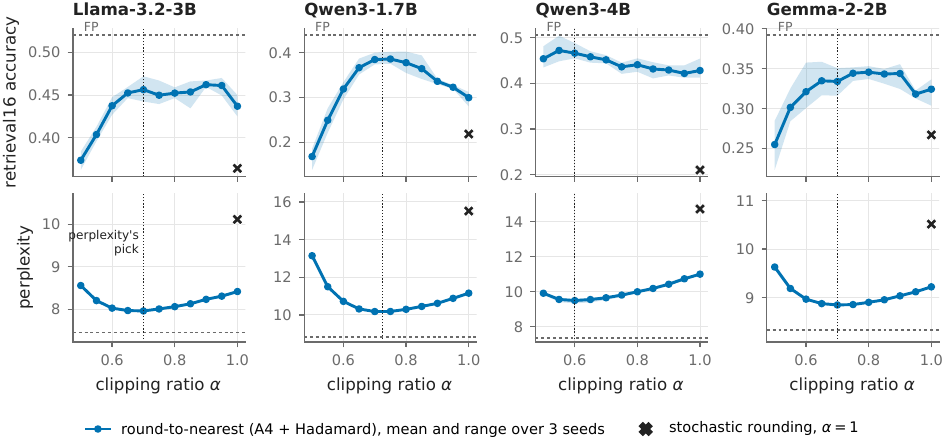}
  \caption{Clipping-ratio sweeps after a full rotation (A4 + Hadamard). Top: retrieval16 accuracy.
  Bottom: perplexity. Bands show the range over three seeds, and the dashed line is full precision. The
  dotted vertical line is the ratio perplexity picks. On held-out items its retrieval16 is within the
  confidence interval of the best ratio's on three models, and 0.015 below it on Gemma-2-2B
  (Table~\ref{tab:clip}). Stochastic rounding (cross) is worse than round-to-nearest at the same ratio on
  both measures, for every model.}
  \label{fig:clipping}
\end{figure}

\paragraph{Stochastic rounding is worse.}
Stochastic rounding removes the bias of round-to-nearest but roughly doubles the error power. If bias
were the problem, it would help. If power is the problem, it should hurt. It hurts, on every model and
seed (Table~\ref{tab:sr}). Retrieval16 drops by a further 0.06--0.22 and induction by 0.02--0.06.
Perplexity rises along with them, so perplexity orders the two correctly.

\begin{table}[t]
  \centering
  \small
  \caption{Part~II. A4 + Hadamard, $\alpha = 1$: round-to-nearest against stochastic rounding, mean
  over three seeds (2000 items each). Stochastic rounding is worse on every model, seed and probe. Perplexity is on the 40 WikiText-2 windows of the probe runs (Section~\ref{sec:setup}).}
  \label{tab:sr}
  \begin{tabular}{l l rrrr}
    \toprule
    Model & Rounding & Induction & Retrieval16 & Retrieval64 & Perplexity \\
    \midrule
    Llama-3.2-3B & FP & 0.971 & 0.520 & 0.185 & 7.46 \\
                 & RTN & 0.954 & 0.437 & 0.128 & 8.42 \\
                 & stochastic & 0.910 & 0.364 & 0.096 & 10.11 \\
    \midrule
    Qwen3-1.7B-Base & FP & 0.978 & 0.439 & 0.111 & 8.82 \\
                 & RTN & 0.964 & 0.299 & 0.077 & 11.16 \\
                 & stochastic & 0.909 & 0.219 & 0.052 & 15.51 \\
    \midrule
    Qwen3-4B-Base & FP & 0.979 & 0.506 & 0.199 & 7.35 \\
                 & RTN & 0.968 & 0.428 & 0.139 & 11.01 \\
                 & stochastic & 0.930 & 0.210 & 0.062 & 14.72 \\
    \midrule
    Gemma-2-2B   & FP & 0.983 & 0.392 & 0.121 & 8.33 \\
                 & RTN & 0.973 & 0.324 & 0.074 & 9.22 \\
                 & stochastic & 0.955 & 0.267 & 0.063 & 10.51 \\
    \bottomrule
  \end{tabular}
\end{table}

\paragraph{Removing the signal-parallel part does not help.}
If the deadzone mostly shrinks each token, an obvious repair is to undo that shrinking. After rotation,
though, the error with that part removed does as much damage as the full error, or more: 0.83 [0.52,
1.21] of the retrieval damage on Llama, 1.00 on Qwen3-1.7B, 1.20 on Qwen3-4B, 1.02 on Qwen3-8B, and 1.37
[1.04, 1.86] on Gemma-2-2B. This was the idea that started Part~II, and the data do not support it.

\paragraph{Perplexity picks the right clipping ratio.}
Clipping trades clipping error against rounding error (Figure~\ref{fig:clipping}). We sweep $\alpha$
from 0.5 to 1.0, pick it on half the items by five different rules, and score the choice on the other
half (Table~\ref{tab:clip}). The ratio that minimises perplexity matches the retrieval-optimal ratio
within the confidence intervals on retrieval16 for Llama, Qwen3-1.7B and Qwen3-4B, and within 0.01 on
retrieval64 for all four models. Gemma-2-2B is the exception on retrieval16. Perplexity picks 0.70 on
every seed, while the best ratio for retrieval moves between 0.65 and 0.80, and the perplexity choice
gives up 0.015 [0.005, 0.025]. That is still less than the 0.025 that $\alpha = 1$ gives up. Across the
four models, fixing $\alpha = 1$ gives up 0.018--0.082 of retrieval16, and $\alpha = 0.9$ up to 0.050 of
retrieval64. In this regime, perplexity is a good enough way to tune the quantiser.

\begin{table}[t]
  \centering
  \small
  \caption{Part~II. Clipping-ratio selection, A4 + Hadamard RTN, three seeds pooled. Gap = accuracy
  of the retrieval-optimal ratio minus that of the rule's ratio, on held-out items (95\% CI);
  positive means the rule loses accuracy. $\alpha$ picked by perplexity: 0.70 (Llama), 0.70--0.75
  (Qwen3-1.7B), 0.60 (Qwen3-4B), 0.70 (Gemma). Perplexity is on the 40 WikiText-2 windows of the probe runs (Section~\ref{sec:setup}).}
  \label{tab:clip}
  \resizebox{\linewidth}{!}{%
  \begin{tabular}{l l rrrr}
    \toprule
    Probe & Rule & Llama-3.2-3B & Qwen3-1.7B & Qwen3-4B & Gemma-2-2B \\
    \midrule
    Retrieval16 & perplexity & $-0.002$ [$-0.016$, 0.012] & $-0.006$ [$-0.021$, 0.010] & $-0.002$ [$-0.016$, 0.013] & 0.015 [0.005, 0.025] \\
                & $\alpha = 1$ & 0.018 [0.003, 0.033] & 0.082 [0.064, 0.099] & 0.042 [0.025, 0.060] & 0.025 [0.011, 0.040] \\
                & $\alpha = 0.9$ & $-0.006$ [$-0.015$, 0.002] & 0.040 [0.023, 0.056] & 0.030 [0.013, 0.047] & 0.005 [$-0.008$, 0.018] \\
    \midrule
    Retrieval64 & perplexity & 0.009 [0.000, 0.017] & $-0.010$ [$-0.020$, 0.000] & 0.001 [$-0.008$, 0.010] & 0.005 [$-0.002$, 0.012] \\
                & $\alpha = 1$ & 0.013 [0.002, 0.024] & 0.032 [0.022, 0.043] & 0.049 [0.036, 0.062] & 0.021 [0.012, 0.030] \\
                & $\alpha = 0.9$ & 0.001 [$-0.008$, 0.009] & 0.026 [0.015, 0.036] & 0.050 [0.037, 0.063] & 0.012 [0.003, 0.020] \\
    \bottomrule
  \end{tabular}}
\end{table}

\paragraph{What this says about the Part~I repair.}
Section~\ref{sec:restore} could not tell whether rotation helps by removing the error's structure or by
making it smaller. Part~II says it does both, one after the other. Rotation first turns a structured,
deadzone-driven error into something much closer to rounding noise. After that, most of the remaining
damage follows the error's power.

\section{Partial Rotation and a Full-Width Fix}
\label{sec:partial}

\begin{figure}[t]
  \centering
  \includegraphics[width=\linewidth]{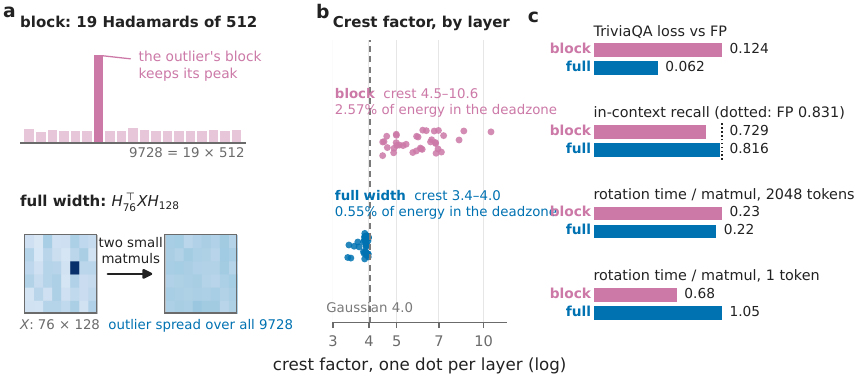}
  \caption{The full-width fix on Qwen3-4B's down-projection input (width 9728). (a) QuaRot's
  construction rotates this width as 19 independent 512-wide Hadamard blocks, so an outlier's energy stays
  inside its block. Viewing the input as a $76 \times 128$ matrix $X$, the full-width rotation
  $H_{76}^\top X H_{128}$ mixes every value with every other (schematic). (b) Crest factor of the
  down-projection input, one dot per layer (mean over 2048 non-sink WikiText-2 tokens, the same capture as
  Figure~\ref{fig:crest}b and Section~\ref{sec:partial-q25}), under the block and full-width rotations,
  with the share of the input's energy in the A4 deadzone ($\alpha = 1$). The dashed line is the mean
  crest factor of i.i.d.\ Gaussian vectors of the same width. (c) What the fix buys (rotation seed 0;
  three-seed means are in the text): TriviaQA loss against full precision and in-context recall (block:
  tuned block-rotated model, $\alpha = 0.60$; full: full-width rotation at its perplexity-picked $\alpha =
  0.70$), and the rotation's time as a fraction of the down-projection matmul on one RTX 3090
  (Table~\ref{tab:kronlat}). At one token both rotations are dominated by kernel-launch overhead.}
  \label{fig:fix}
\end{figure}

One input in our models is only rotated in blocks. In this section we show that it carries most of
Qwen3-4B's quantisation error and creates a misleading result in the noise comparison, that inputs like
it are common in open models, and that a full-width Kronecker Hadamard removes the problem. On natural
tasks it halves the closed-book TriviaQA loss, at about the block rotation's cost (Figure~\ref{fig:fix}).

Qwen3-4B's down-projection input is the only input in our models whose width QuaRot cannot build a
Hadamard matrix for. Under the block rotation it stays heavy-tailed (crest factor 6.98 and excess kurtosis
5.86 on retrieval16 inputs), and its relative 4-bit error power is 8.0\%, against 2.1--2.6\% for every
other input (Table~\ref{tab:crest}). This one input is why Qwen3-4B's retrieval row in Table~\ref{tab:k1}
looks unlike the others. The 128-wide blocks of llm-compressor's examples leave it even further from
Gaussian. On a 32-token WikiText-2 capture at layers 9, 18 and 27, the mean
crest factor is 5.4, 6.8 and 9.1 with 128-wide blocks, 4.6, 6.0 and 7.1 with our 512-wide blocks, and 3.9,
4.0 and 3.8 at full width. We are extending this capture to more tokens. We have not run 128-wide
blocks downstream.

\paragraph{A misleading result.}
With every input quantised, matched noise does 1.6--2.7 times as much retrieval16 damage as the real
error on Qwen3-4B, with every confidence interval above 1 on each of three seeds (Table~\ref{tab:partial}).
Taken at face value, this says the real error avoids the directions retrieval depends on. It does not.
If we leave the down-projection input unquantised, the ratio falls to 0.89--1.21, and if we rotate it
densely, to 0.61--1.09. The gap is a side effect of the incomplete rotation.

\begin{table}[t]
  \centering
  \small
  \caption{Part~II, Qwen3-4B, A4 + rotation, $\alpha=1$. Retrieval16 noise/real (95\% CI) per seed,
  by how the down-projection input is treated. Block: block-diagonal Hadamard (standard); FP: left
  unquantised; dense: dense random orthogonal rotation; full Had.: $H_{76} \otimes H_{128}$.}
  \label{tab:partial}
  \begin{tabular}{l l rrr}
    \toprule
    down\_proj & Control & Seed 0 & Seed 1 & Seed 2 \\
    \midrule
    block & Gaussian & 2.57 [1.94, 3.82] & 1.82 [1.47, 2.35] & 1.61 [1.32, 2.01] \\
          & sign     & 2.66 [2.01, 3.94] & 1.61 [1.29, 2.09] & 1.62 [1.33, 2.01] \\
    FP    & Gaussian & 0.89 [0.64, 1.23] & 1.21 [0.85, 1.79] & 0.91 [0.66, 1.25] \\
          & sign     & 1.03 [0.76, 1.39] & 0.99 [0.66, 1.49] & 1.06 [0.78, 1.46] \\
    dense & Gaussian & 0.85 [0.67, 1.06] & 0.61 [0.44, 0.77] & 1.03 [0.78, 1.36] \\
          & sign     & 0.90 [0.71, 1.12] & 0.66 [0.51, 0.84] & 1.09 [0.83, 1.46] \\
    full Had. & Gaussian & 0.92 [0.72, 1.18] & 0.96 [0.76, 1.21] & 0.85 [0.66, 1.06] \\
          & sign     & 0.95 [0.74, 1.22] & 0.87 [0.68, 1.10] & 1.05 [0.86, 1.28] \\
    \bottomrule
  \end{tabular}
\end{table}

\paragraph{This input's error has a perplexity part and a retrieval part.}
Quantising only the block-rotated down-projection input (Table~\ref{tab:decomp}) raises perplexity from
7.35 to about 10.3. The deadzone part alone accounts for nearly all of that (about 10.1), yet costs no
short-range retrieval ($+0.009$ [0.000, 0.018]). The grid part costs little perplexity (8.0) but lowers
retrieval16 by 0.077. Matched noise is worse still for retrieval ($-0.100$), at a perplexity of only 8.5.
This explains why noise looked worse than the real error here: most of the real error's perplexity cost
comes from a part that retrieval does not feel. Pure per-token shrinking (\emph{gain only}) is harmless
on both measures. Longer-range retrieval does not split this way. The deadzone and grid parts cost
retrieval64 about equally (0.020 and 0.018).

\begin{table}[t]
  \centering
  \small
  \caption{Part~II, Qwen3-4B. Only the block-rotated down-projection input is quantised (A4,
  $\alpha=1$). Accuracy change against full precision, three seeds pooled (6000 items, 95\% paired
  CI). Perplexity is the mean over seeds; FP perplexity 7.35. Perplexity is on the 40 WikiText-2 windows of the probe runs (Section~\ref{sec:setup}).}
  \label{tab:decomp}
  \begin{tabular}{l rr r}
    \toprule
    Arm & $\Delta$ Retrieval16 & $\Delta$ Retrieval64 & Perplexity \\
    \midrule
    Real error & $-0.043$ [$-0.054$, $-0.033$] & $-0.029$ [$-0.038$, $-0.022$] & 10.28 \\
    Deadzone only & $+0.009$ [0.000, 0.018] & $-0.020$ [$-0.027$, $-0.013$] & 10.14 \\
    Grid only & $-0.077$ [$-0.086$, $-0.068$] & $-0.018$ [$-0.025$, $-0.012$] & 8.04 \\
    Matched Gaussian & $-0.100$ [$-0.110$, $-0.089$] & $-0.056$ [$-0.065$, $-0.049$] & 8.58 \\
    Sign-randomised & $-0.115$ [$-0.125$, $-0.104$] & $-0.046$ [$-0.053$, $-0.038$] & 8.49 \\
    Gain only (per-token attenuation) & $-0.004$ [$-0.006$, $-0.001$] & $-0.002$ [$-0.004$, 0.000] & 7.36 \\
    Gain removed & $-0.043$ [$-0.054$, $-0.033$] & $-0.031$ [$-0.038$, $-0.023$] & 10.17 \\
    \bottomrule
  \end{tabular}
\end{table}

\paragraph{A dense rotation helps perplexity more than retrieval.}
Rotating this input with a dense random matrix improves perplexity (11.0 to about 9.1) but slightly
lowers retrieval16 ($-0.018$ [$-0.030$, $-0.006$], pooled over three seeds, one of which goes the other
way). It makes no measurable difference on retrieval64 ($+0.004$) and gives a small gain on induction
($+0.004$). Adding the block-rotated input to the other quantised inputs costs 0.025 [0.014, 0.037] of
retrieval16 and 0.028 of retrieval64 (pooled). So it is not free for retrieval, but its perplexity cost
(8.0 to 11.0) is far larger.

\paragraph{A full-width Hadamard rotation.}
The block rotation is a limit of QuaRot's construction, not of the width itself. The construction we use
is not new. QuaRot's rotation, which follows QuIP\#'s randomised Hadamard transform
\citep{tseng2024quipsharp}, is already a Kronecker product of a stored small Hadamard matrix and a
Walsh--Hadamard matrix. FlatQuant learns Kronecker-structured transforms \citep{sun2024flatquant}. And
compressed-tensors, the transform library under llm-compressor, has stored Hadamard orders up to 256,
including 76 and 148, since June 2025, so its default full-width path already rotates widths 9728 and
18944 in one piece.\footnote{\texttt{transform/utils/hadamards.safetensors} in compressed-tensors
\citep{compressedtensors}, extended in commit \texttt{852b1fa}.} What we add is evidence that the
block-diagonal option, the natural extension of QuaRot and one llm-compressor exposes, changes the
error as shown above; that this matters for some inputs and not others; and that the full-width
rotation closes the gap.

The matrix is $H_{9728} = H_{76} \otimes H_{128}$, with $H_{76}$ from Paley's second construction ($q =
37$) and $H_{128}$ from Sylvester's. We check that $HH^\top = 9728\,I$ exactly. As in QuaRot, we apply it
as $\mathrm{diag}(s)H/\sqrt{9728}$ with random signs $s$. If we view each token's input as a $76 \times
128$ matrix $X$, the rotation is $H_{76}^\top X H_{128}$: two small matmuls instead of one $9728 \times
9728$ matmul.

Table~\ref{tab:kronlat} gives the timings. From 256 tokens up, the full-width rotation costs about the
same as the block rotation, and from 2048 tokens up it costs about a quarter of the down-projection
matmul. At one token, every option is dominated by kernel launch overhead. The dense rotation costs about
three times the down-projection matmul at every size. Batch-1 decoding is where the full-width rotation
costs more: at one token it takes 0.091\,ms against 0.059\,ms for the block rotation (1.05 and 0.68 times
the matmul), using plain PyTorch with two separate matmuls. A cheaper version is possible. The $H_{128}$
factor can be applied with a fast Walsh--Hadamard transform, leaving the $76 \times 76$ factor as the only
extra small matmul. This is how QuaRot applies its own stored orders ($H_K \otimes H_{2^m}$), with a fused
CUDA kernel. Our unfused PyTorch version of that transform is slower, not faster (0.54\,ms at one token),
because it launches one kernel per butterfly stage. We expect a fused kernel to bring the one-token cost
close to the block rotation's, but we have not measured one.

\paragraph{The width gap is common.}
Qwen3-4B is not an isolated case. In the public configurations of 42 popular open models, 10 have at least
one linear-input width that QuaRot's stored orders cannot rotate in one piece (Table~\ref{tab:widths}).
It is almost always the MLP down-projection input. Qwen2.5-7B and 72B, Qwen3-14B and 32B, Gemma-3-27B,
Falcon3-7B, Granite-3.1-8B and DeepSeek-V2-Lite are among them. Every one of these widths has a
full-width Hadamard of the same kind ($H_m \otimes H_{2^k}$, with $H_m$ from Paley's constructions or a
Kronecker product of them). Whether the gap matters is a separate question. The last two columns of the
table measure how far the block rotation leaves each input from the full-rotation regime, and
Section~\ref{sec:partial-q25} tests the fix on two more models from the table, Qwen2.5-0.5B and
Qwen2.5-7B.

\begin{table}[t]
\centering
\small
\caption{Linear-input widths of popular open models that QuaRot's Hadamard construction ($K\cdot 2^k$, $K \in \{12, 20, 28, 36, 40, 52, 60, 108, 140, 156, 172\}$) cannot rotate in one piece, from the public configurations of 42 models. Each such width falls back to a block-diagonal rotation unless the library builds another order; the last column gives a full-width Hadamard from Paley's constructions and Kronecker products. The last two columns give, for the MLP down-projection input of the models we ran, the mean crest factor and the percentage of energy A4 ($\alpha = 1$) rounds to zero, under the block-diagonal rotation / the full-width one (2048 WikiText-2 tokens, mean over layers; -- not measured). The 32 other models are fully covered: Yi-1.5-9B, SmolLM2-1.7B, Qwen2.5-1.5B, Qwen2.5-14B, Qwen2.5-32B, Qwen2.5-3B, Qwen3-0.6B-Base, Qwen3-1.7B-Base, Qwen3-30B-A3B-Base, Qwen3-8B-Base, OLMo-2-1124-13B, OLMo-2-1124-7B, Phi-3-medium-4k-instruct, Phi-3-mini-4k-instruct, Phi-3.5-mini-instruct, Phi-4-mini-instruct, phi-4, Llama-3.2-1B, Llama-3.2-3B, Meta-Llama-3.1-70B, Meta-Llama-3.1-8B, Mistral-Nemo-Base-2407, Mistral-Small-24B-Base-2501, gemma-2-27b, gemma-2-2b, gemma-2-9b, gemma-3-12b-pt, gemma-3-1b-pt, gemma-3-4b-pt, llama-2-13b, llama-2-7b, mistral-7b-v0.3.}
\label{tab:widths}
\resizebox{\linewidth}{!}{%
\begin{tabular}{l l r r l r r}
\toprule
Model & Input & Width & $2^k\cdot$odd & Full-width Hadamard & Crest & Deadzone \% \\
\midrule
Qwen2.5-0.5B & mlp down & 4864 & 256$\cdot$19 & $H_{76}\otimes H_{64}$ & 4.13 / 3.75 & 0.74 / 0.51 \\
Qwen2.5-72B & mlp down & 29568 & 128$\cdot$231 & $H_{924}\otimes H_{32}$ & -- & -- \\
Qwen2.5-7B & mlp down & 18944 & 512$\cdot$37 & $H_{148}\otimes H_{128}$ & 4.76 / 4.01 & 1.22 / 0.62 \\
Qwen3-14B-Base & mlp down & 17408 & 1024$\cdot$17 & $H_{68}\otimes H_{256}$ & 4.37 / 3.96 & 0.90 / 0.60 \\
Qwen3-32B & mlp down & 25600 & 1024$\cdot$25 & $H_{200}\otimes H_{128}$ & 4.48 / 4.08 & 0.94 / 0.65 \\
Qwen3-4B-Base & mlp down & 9728 & 512$\cdot$19 & $H_{76}\otimes H_{128}$ & 6.06 / 3.85 & 2.57 / 0.55 \\
DeepSeek-V2-Lite & mlp down & 10944 & 64$\cdot$171 & $H_{684}\otimes H_{16}$ & -- & -- \\
DeepSeek-V2-Lite & moe down & 1408 & 128$\cdot$11 & $H_{44}\otimes H_{32}$ & -- & -- \\
granite-3.1-8b-base & mlp down & 12800 & 512$\cdot$25 & $H_{200}\otimes H_{64}$ & 4.07 / 3.99 & 0.66 / 0.61 \\
Falcon3-7B-Base & mlp down & 23040 & 512$\cdot$45 & $H_{180}\otimes H_{128}$ & 4.21 / 4.08 & 0.73 / 0.65 \\
gemma-3-27b-pt & hidden & 5376 & 256$\cdot$21 & $H_{84}\otimes H_{64}$ & -- & -- \\
gemma-3-27b-pt & mlp down & 21504 & 1024$\cdot$21 & $H_{84}\otimes H_{256}$ & -- & -- \\
\bottomrule
\end{tabular}}
\end{table}

\paragraph{The full-width rotation behaves like any fully rotated input.}
Its crest factor is 3.85 and its excess kurtosis $-0.10$, against 6.98 and 5.86 under the block
rotation. Its relative 4-bit error power is 2.5\%, in line with every other fully rotated input. The
misleading noise result goes away (Table~\ref{tab:partial}). Matched noise now does 0.85--0.96 of the real
error's retrieval16 damage and the sign-randomised substitute 0.87--1.05, with every interval containing
1, on all three seeds. Unlike the dense rotation, it also keeps induction in line (noise/real 0.90--1.06,
every interval containing 1, against 1.46 [1.27, 1.69] for the dense rotation on seed 0). On retrieval64,
one seed has noise doing less damage than the real error (0.59 [0.38, 0.82]), as on Llama and Gemma.

Perplexity falls from 11.0 to 8.5 at $\alpha = 1$, below the dense rotation's 9.1. On the probes,
retrieval16 is level with the block rotation ($-0.012$ [$-0.024$, 0.000], three seeds pooled) and with
the dense one ($+0.006$ [$-0.006$, 0.017]), and retrieval64 and induction are level or slightly better.
Quantising only this input, its perplexity cost falls from 10.2 (block) and 8.2 (dense) to 7.8, against
7.35 at full precision (seed 0). On full WikiText-2 the perplexity-picked clipping ratio moves from 0.60
to 0.70, where perplexity is 8.67, against 10.29 for the tuned block-rotated model.

\paragraph{What the fix buys on natural tasks.}
At its perplexity-picked clipping ratio, the full-width model loses 0.062 TriviaQA and 0.051 NQ-Open
accuracy against full precision. The tuned block-rotated model loses 0.124 and 0.083 (Table~\ref{tab:qa}).
These are for rotation seed 0, which is the most favourable of three. Over seeds 0--2, the TriviaQA loss
is 0.112 against 0.060, a difference of 0.052 [0.036, 0.068] (pooled over seeds, with questions
resampled). Per seed the difference is 0.062, 0.044 and 0.049, each interval above zero. The NQ-Open loss
is 0.072 against 0.056 (0.016 [0.003, 0.028]); per seed the difference is 0.032, 0.007 and 0.009, and
only seed 0 is resolved.

The remaining numbers in this paragraph use seed 0 only. The CoQA loss falls from 0.050 to 0.008 and the
DROP loss from 0.096 to 0.054, while LongBench is level (0.050 against 0.057). In-context recall rises by
0.087 [0.049, 0.124], to within 0.016 of full precision, and the commonsense average rises by 0.017
(Section~\ref{sec:bench}). With 4-bit weights added, the TriviaQA loss is 0.088 against 0.133
(Section~\ref{sec:w4a4}). The random-token retrieval probes do not show this gain;
Section~\ref{sec:bench} traces the difference to token type and format.

\begin{table}[t]
  \centering
  \small
  \caption{Part~II, Qwen3-4B down-projection input (width 9728), fp16 on one RTX 3090: median time
  in ms to rotate a batch of tokens, against the down-projection matmul itself ($9728 \to 2560$).
  Block: 19 independent 512-wide Hadamard blocks. Full Hadamard: $H_{76} \otimes H_{128}$ as two
  small matmuls. Plain PyTorch, no fused kernels.}
  \label{tab:kronlat}
  \begin{tabular}{r rrrr}
    \toprule
    Tokens & down\_proj matmul & Block & Full Hadamard & Dense \\
    \midrule
    1    & 0.087 & 0.059 & 0.091 & 0.249 \\
    256  & 0.216 & 0.081 & 0.095 & 0.762 \\
    2048 & 1.717 & 0.402 & 0.384 & 5.378 \\
    8192 & 5.958 & 1.625 & 1.479 & 20.87 \\
    \bottomrule
  \end{tabular}
\end{table}

\subsection{A second model: the gain depends on how far the block rotation falls short}
\label{sec:partial-q25}

Qwen2.5-0.5B's down-projection input (width $4864 = 19 \cdot 256$) is also rotated in blocks under
QuaRot's construction, and $H_{76} \otimes H_{64}$ rotates it at full width. We pick the clipping ratio
for this model by full WikiText-2 perplexity over $\alpha \in \{0.5, 0.6, \ldots, 1\}$, as for the other
models. Both rotations pick $\alpha = 0.80$, where perplexity is 15.15 for the full-width rotation against
15.38 for the block rotation (13.07 at full precision; 16.24 against 16.60 at $\alpha = 1$). We use one
rotation seed. Here the fix makes no difference we can resolve. The closed-book TriviaQA loss is 0.051
against 0.042 for the block rotation, a difference of $-0.009$ [$-0.028$, 0.010]. The NQ-Open loss is
0.011 against 0.018 (0.007 [$-0.006$, 0.021]), and the in-context recall loss is 0.067 against 0.062
($-0.004$ [$-0.047$, 0.040]). At $\alpha = 0.70$, the ratio picked for Qwen3-4B, the picture is the same:
the TriviaQA difference is 0.017 [$-0.001$, 0.035], NQ-Open and recall are level, and retrieval16 is
0.375 against 0.373.

The crest factor explains why. On 512-token WikiText-2 windows, captured the same way for both models,
Qwen2.5-0.5B's block-rotated input has a mean crest factor of 4.1 (4.0--4.5 across layers) and puts 0.74\%
of its energy in the A4 deadzone. That is close to its full-width values (3.8 and 0.51\%). Qwen3-4B's
block-rotated input has 6.1 (4.5--10.6) and 2.57\%, against 3.9 and 0.55\% at full width
(Figure~\ref{fig:fix}b). These WikiText-2 values are lower than the retrieval16 values of
Table~\ref{tab:crest} (6.98 for Qwen3-4B's block-rotated input) because the inputs differ. On Qwen3-4B,
where we have both, the order is the same. Narrower blocks of a narrower input leave less outlier mass
behind, so the block rotation already puts Qwen2.5-0.5B close to the full-rotation regime.

So the full-width rotation fixes block-rotated inputs that stay far from Gaussian, as Qwen3-4B's does. It
is not a general accuracy boost. Whether a model needs it can be read from its block-rotated input's crest
factor and deadzone energy before running any benchmark. Table~\ref{tab:widths} applies this check to five
more affected models, using the same capture. None is as far from the full-rotation regime as Qwen3-4B.
Qwen2.5-7B comes closest (crest factor 4.76 and 1.22\% deadzone energy under the block rotation, against
4.01 and 0.62\% at full width), followed by Qwen3-32B and Qwen3-14B (4.48 and 4.37; 0.94\% and 0.90\%).
Falcon3-7B and Granite-3.1-8B are already within 0.13 of their full-width crest factor, and their deadzone
energy (0.73\% and 0.66\%) is no higher than Qwen2.5-0.5B's, where the fix changed little. We tested the
full-width rotation downstream only on Qwen2.5-7B, the closest of them. For Qwen3-14B, Qwen3-32B,
Falcon3-7B and Granite-3.1-8B we have the measurement only, and make no claim about their downstream gain.

On Qwen2.5-7B, perplexity picks $\alpha = 0.70$ for both rotations over $\alpha \in \{0.6, \ldots, 1\}$
(7.25 for the full-width rotation against 7.34 for the block rotation; 6.85 at full precision). Over three
rotation seeds, the full-width rotation lowers the closed-book TriviaQA loss from 0.038 [0.022, 0.054] to
0.018 [0.004, 0.032], a difference of 0.020 [0.008, 0.032] (pooled over seeds, with questions resampled).
Per seed the difference is 0.033, 0.018 and 0.009, and only the first interval excludes zero. The NQ-Open
loss does not move (0.023 against 0.026; 0.003 [$-0.008$, 0.015]). Neither does in-context recall: with
500 items per task (rotation seed 0) the loss is 0.015 against 0.019, a difference of 0.003 [$-0.014$,
0.021]. An opposite-sign result on the first 150 items ($-0.022$ [$-0.056$, 0.011]) was noise. So the
benefit reaches a second model, but only on closed-book TriviaQA, at about 40\% of its size on Qwen3-4B,
and it is not robust on any single seed. Qwen3-4B's block-rotated input is the most outlier-heavy of the
seven we measured. The width gap is common, but how much damage it does depends on the model.

\section{External Validity: Deployed Quantisers, Benchmarks, and Scale}
\label{sec:external}

\subsection{Deployed quantisers}
\label{sec:realquant}

We test the split between perplexity and probes on GPTQ \citep{frantar2023gptq}, AWQ \citep{lin2024awq}
and SmoothQuant \citep{xiao2023smoothquant}, on pythia-1.4b and Qwen3-8B-Base, in their weight-only
configurations and with activations quantised to 4 bits (Table~\ref{tab:realquantfull}). Even with only
the weights quantised, retrieval drops before perplexity moves much. Under AWQ W4A16, Qwen3-8B-Base's
perplexity goes from 19.9 to 21.0 while retrieval falls from 0.480 to 0.383. AWQ at W4A4 shows the same
split. pythia-1.4b keeps \PythiaAwqIndPct{} percent of its induction but only \PythiaAwqRetPct{} percent
of its retrieval, and Qwen3-8B-Base keeps \QwenAwqIndPct{} and \QwenAwqRetPct{} percent. GPTQ at W4A4
damages both probes together, and we did not run SmoothQuant below W8A8. So the effect depends on the
method, not on activation quantisation as such. On a small set of standard benchmarks, GPTQ W4A4
collapses LAMBADA, while the four multiple-choice commonsense tasks move little under any method, partly
because they are near chance even at full precision (Figure~\ref{fig:benchmark-battery}).

\subsection{Standard benchmarks in the deployed setting}
\label{sec:bench}

We run the deployed setting through lm-evaluation-harness (v0.4.9.1) on four models. The suites are:
the zero-shot tasks QuaRot reports (ARC-easy, ARC-challenge, HellaSwag, PIQA, WinoGrande and LAMBADA,
averaged); MMLU five-shot; in-context recall (SQuAD completion, SWDE and FDA, the first 150 documents of
each); and RULER at 4k tokens (seven needle-in-a-haystack tasks and SQuAD question answering, 50 items
each). Table~\ref{tab:bench} gives the scores. Every difference we quote is paired over items, with a 95\%
bootstrap interval. The intervals are about $\pm 0.008$ for the commonsense average and MMLU, $\pm
0.03$--$0.04$ for recall, $\pm 0.01$--$0.03$ for needle-in-a-haystack and $\pm 0.12$--$0.14$ for RULER QA.
RULER QA differences smaller than about 0.13 are therefore not significant.

\begin{table}[t]
  \centering
  \small
  \setlength{\tabcolsep}{4pt}
  \caption{Part~II, standard benchmarks. Perplexity is full WikiText-2 at context 2048. CS = mean of
  ARC-e, ARC-c, HellaSwag, PIQA, WinoGrande and LAMBADA (zero-shot); recall = mean of SQuAD
  completion, SWDE and FDA; NIAH = mean of seven RULER needle tasks at 4k; RULER QA = SQuAD at 4k.
  ``Tuned'' is A4 + Hadamard with the perplexity-picked clipping ratio; ``no rot.'' is A6 without
  rotation. Indented rows are rotated A4 settings whose perplexity brackets the unrotated model's.
  Paired 95\% intervals for every comparison are in the benchmark report of the code release
  (Section~\ref{sec:repro}). Perplexity is on the full WikiText-2 test set, so it differs from the 40-window values of Tables~\ref{tab:k1}--\ref{tab:decomp}.}
  \label{tab:bench}
  \begin{tabular}{l l r rrrrr}
    \toprule
    Model & Setting & Perplexity & CS & MMLU & Recall & NIAH & RULER QA \\
    \midrule
    Llama-3.2-3B & FP & 7.81 & 0.681 & 0.565 & 0.753 & 0.996 & 0.608 \\
                 & tuned ($\alpha=0.70$) & 8.36 & 0.661 & 0.536 & 0.749 & 0.986 & 0.648 \\
                 & no rot. & 8.76 & 0.644 & 0.523 & 0.729 & 0.970 & 0.622 \\
                 & \quad $\alpha=1$ & 8.84 & 0.643 & -- & 0.729 & 0.966 & 0.508 \\
    \midrule
    Qwen3-1.7B   & FP & 9.39 & 0.640 & 0.627 & 0.809 & 0.990 & 0.662 \\
                 & tuned ($\alpha=0.70$) & 10.86 & 0.604 & 0.575 & 0.738 & 0.991 & 0.668 \\
                 & no rot. & 14.98 & 0.553 & 0.465 & 0.562 & 0.837 & 0.282 \\
                 & \quad $\alpha=0.5$ & 14.19 & 0.569 & -- & 0.640 & 0.886 & 0.528 \\
                 & \quad stochastic & 16.80 & 0.514 & -- & 0.589 & 0.863 & 0.355 \\
    \midrule
    Qwen3-4B     & FP & 7.90 & 0.699 & 0.731 & 0.831 & 0.993 & 0.837 \\
                 & tuned ($\alpha=0.60$) & 10.29 & 0.657 & 0.667 & 0.729 & 0.960 & 0.667 \\
                 & no rot. & 10.82 & 0.626 & 0.628 & 0.738 & 0.962 & 0.647 \\
                 & \quad $\alpha=0.5$ & 10.71 & 0.649 & -- & 0.738 & 0.923 & 0.747 \\
                 & \quad $\alpha=1$ & 11.82 & 0.632 & -- & 0.687 & 0.979 & 0.620 \\
                 & $\alpha=1$, dense down & 9.76 & 0.644 & -- & 0.789 & 0.973 & 0.655 \\
                 & $\alpha=1$, full Had.\ down & 9.14 & 0.660 & -- & 0.773 & 0.983 & 0.682 \\
                 & $\alpha=0.70$, full Had.\ down & 8.67 & 0.674 & -- & 0.816 & 0.987 & 0.770 \\
    \midrule
    Gemma-2-2B   & FP & 8.73 & 0.701 & 0.531 & 0.813 & 0.959 & 0.502 \\
                 & tuned ($\alpha=0.70$) & 9.27 & 0.680 & 0.498 & 0.793 & 0.931 & 0.528 \\
                 & no rot. & 11.42 & 0.624 & 0.419 & 0.644 & 0.714 & 0.295 \\
                 & \quad stochastic & 11.01 & 0.615 & -- & 0.684 & 0.773 & 0.348 \\
                 & \quad $\alpha=0.4$ & 13.33 & 0.593 & -- & 0.680 & 0.703 & 0.510 \\
    \bottomrule
  \end{tabular}
\end{table}

\paragraph{A tuned, rotated quantiser is close to lossless on two of the four models.}
With the perplexity-picked clipping ratio, Llama-3.2-3B and Gemma-2-2B keep their recall within 0.02 of
full precision (neither difference is significant) and lose 0.021 on the commonsense average. The two Qwen
models lose more recall (0.071 on Qwen3-1.7B, 0.102 on Qwen3-4B). Leaving the ratio at $\alpha = 1$ costs
more than tuning it, on both models where we ran it. That holds for the commonsense average (Llama 0.643
against 0.661, Qwen3-4B 0.632 against 0.657) and for recall (0.729 against 0.749, and 0.687 against 0.729).

\paragraph{The block-rotated input costs Qwen3-4B its recall.}
Rotating Qwen3-4B's down-projection input densely, at $\alpha = 1$, raises recall by 0.102 [0.058, 0.147]
and the commonsense average by 0.012 [0.004, 0.021] over the standard block rotation, and lowers
perplexity from 11.82 to 9.76. The densely rotated model at $\alpha = 1$ even recalls more than the
block-rotated model at its tuned ratio (0.789 against 0.729). On real documents, the one partly rotated
input matters more than the choice of clipping ratio.

The full-width Hadamard $H_{76} \otimes H_{128}$ gets the same recall gain without the dense rotation's
cost. At $\alpha = 1$ it raises recall by 0.087 [0.044, 0.129] over the block rotation, level with the dense
rotation ($-0.016$ [$-0.051$, 0.020]), and its commonsense average is higher than the dense rotation's by
0.016 [0.008, 0.025]. At its perplexity-picked ratio ($\alpha = 0.70$), it beats the tuned block-rotated
model on commonsense ($+0.017$ [0.009, 0.025]), recall ($+0.087$ [0.049, 0.124]) and NIAH ($+0.027$
[0.012, 0.044]). RULER QA is higher by 0.103, but its interval [$-0.003$, 0.213] includes zero. Its recall
is within 0.016 [$-0.044$, 0.011] of full precision, so Qwen3-4B joins Llama and Gemma as close to lossless
on recall. Its commonsense loss (0.025) is the smallest of any A4 setting on Qwen3-4B.

This disagrees with the probes. On random-token retrieval16, the dense rotation slightly lowers accuracy
($-0.018$, Section~\ref{sec:partial}) and the full-width Hadamard leaves it level ($-0.012$), while on the
recall benchmarks they raise accuracy by 0.102 and 0.087. Both can be true. The probes use short
random-token sequences with 16 key--value pairs. The recall tasks use natural documents, longer contexts,
answers that are spans rather than single tokens, and a language prior that the rotation's perplexity gain
helps.

\paragraph{Bridging the probes and the benchmarks.}
To find out which of these differences matters, we built retrieval variants that each change one thing
(Table~\ref{tab:bridge}): 4 or 8 pairs instead of 16; common English words in a natural `` key: value,''
format instead of random tokens; and 512 or 1536 tokens of WikiText-2 text between the pairs and the
query. On Qwen3-4B at $\alpha = 1$, the full-width Hadamard is level with the block rotation on every
random-token variant, including the long-context ones (differences of $-0.027$ to $+0.009$, none
significant). On the natural-format variant it raises accuracy by 0.096 [0.06, 0.13] (0.715 against
0.619). What the random-token probe misses about the rotation is token type and format, not distance or
the number of pairs. The recall gain on benchmarks is a gain on natural text. At the perplexity-picked
ratios, the full-width model is ahead on every variant (by 0.03--0.06, significant on eight of nine).

\begin{table}[t]
\centering
\small
\setlength{\tabcolsep}{4pt}
\caption{Bridge probes (1000 items each): retrieval with fewer pairs, with frequent English words in a natural `` key: value,'' format, and with 512 or 1536 tokens of WikiText-2 text between the pairs and the query. Rows give full-precision accuracy and the share of the real A4 error's ($\alpha = 1$) accuracy damage that matched noise reproduces; $^*$ marks ratios whose 95\% bootstrap interval excludes 1. The easier 4- and 8-pair probes reproduce the 16-pair pattern on every model.}
\label{tab:bridge}
\begin{tabular}{l l rrrrrr}
\toprule
Model & & 4 pairs & 8 pairs & 16 pairs & words, ``k: v,'' & +512 text & +1536 text \\
\midrule
Llama-3.2-3B & FP accuracy & 0.574 & 0.617 & 0.528 & 0.349 & 0.423 & 0.332 \\
 & Gaussian / real & 1.17 & 1.33 & 0.91 & 0.67 & 0.93 & 1.12 \\
 & sign / real & 1.12 & 0.80 & 0.94 & 0.98 & 0.77 & 1.08 \\
\midrule
Qwen3-1.7B & FP accuracy & 0.605 & 0.628 & 0.438 & 0.669 & 0.471 & 0.418 \\
 & Gaussian / real & 0.74$^*$ & 0.81 & 0.80$^*$ & 0.80$^*$ & 0.98 & 0.88 \\
 & sign / real & 0.95 & 0.94 & 0.64$^*$ & 0.95 & 0.87 & 1.06 \\
\midrule
Qwen3-4B & FP accuracy & 0.665 & 0.644 & 0.489 & 0.799 & 0.609 & 0.607 \\
 & Gaussian / real, block & 1.30 & 1.80$^*$ & 3.51$^*$ & 1.08 & 1.44$^*$ & 1.25 \\
 & Gaussian / real, full Had. & 0.79 & 0.87 & 0.85 & 1.08 & 0.97 & 0.93 \\
\midrule
Gemma-2-2B & FP accuracy & 0.611 & 0.547 & 0.381 & 0.697 & 0.408 & 0.336 \\
 & Gaussian / real & 0.57$^*$ & 0.79 & 0.88 & 0.79 & 1.00 & 0.88 \\
 & sign / real & 0.58$^*$ & 0.70$^*$ & 0.87 & 0.87 & 0.94 & 0.78$^*$ \\
\bottomrule
\end{tabular}
\end{table}

\paragraph{An easier probe gives the same picture.}
Retrieval16 accuracy at full precision is only 0.38--0.53, so many of our noise/real ratios rest on small
gaps. The 4- and 8-pair variants are easier (0.55--0.67 at full precision) and give the same ordering on
every model (Table~\ref{tab:bridge}). Matched noise reproduces about all of the real error's damage on
Llama-3.2-3B (0.80--1.33, intervals including 1), less of it on Qwen3-1.7B and Gemma-2-2B (0.57--0.95),
more of it under Qwen3-4B's block rotation (1.30--1.80), and about all of it under Qwen3-4B's full-width
rotation (0.79--0.87, intervals including 1). The conclusions of Section~\ref{sec:rot} do not depend on
the hard probe.

\begin{figure}[t]
  \centering
  \includegraphics[width=\linewidth]{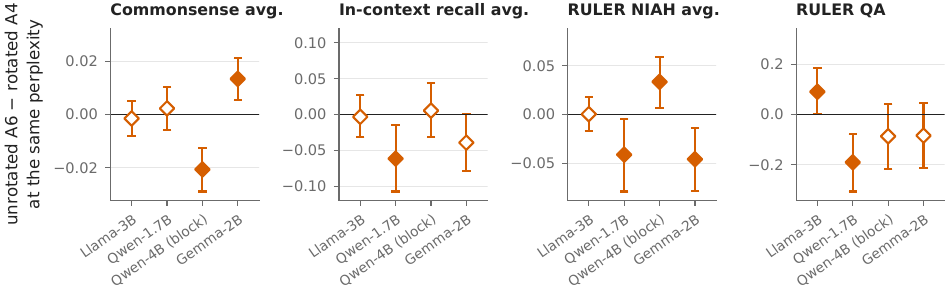}
  \caption{Dropping rotation at equal perplexity. Score of the unrotated A6 model minus that of a rotated
  A4 model at the same perplexity. The rotated score is interpolated linearly in cross-entropy between the
  two rotated settings whose perplexity brackets the unrotated model's (Llama-3.2-3B: $\alpha = 0.70$ and
  1; Qwen3-1.7B: $\alpha = 0.5$ and stochastic rounding; Qwen3-4B: block-rotated $\alpha = 0.5$ and 1;
  Gemma-2-2B: stochastic rounding and $\alpha = 0.4$). Bars are paired 95\% bootstrap intervals over
  items, and filled markers exclude zero. Dropping rotation costs in-context recall on Qwen3-1.7B and,
  borderline, on Gemma-2-2B, and NIAH on both, while their commonsense averages do not drop. Raw scores
  against perplexity are in Figure~\ref{fig:bench-raw} (Appendix~\ref{app:part2}).}
  \label{fig:bench}
\end{figure}

\paragraph{At equal perplexity, dropping rotation costs recall on two models.}
If perplexity misses what structured error does, an unrotated model should do worse on in-context tasks
than a rotated model with the same perplexity. Figure~\ref{fig:bench} interpolates the rotated settings to
the unrotated model's perplexity. Recall falls by 0.061 [0.015, 0.107] on Qwen3-1.7B and by 0.039 [0.000,
0.079] on Gemma-2-2B, and NIAH by 0.041 and 0.046 (both intervals excluding zero), while the commonsense
average is level or higher. On Llama-3.2-3B and Qwen3-4B, recall is level.

The direct comparisons with each neighbour tell the same story. On Qwen3-1.7B, against the rotated model
with lower perplexity ($\alpha = 0.5$), the unrotated model loses 0.078 [0.027, 0.129] of recall, 0.049 of
NIAH and 0.247 of RULER QA, but only 0.015 of the commonsense average. Against stochastic rounding, which
has higher perplexity, recall and NIAH are level and the commonsense average is 0.040 higher. Gemma-2-2B
shows a similar but weaker pattern. The unrotated model has a higher commonsense average than both
neighbours (by 0.009 and 0.031), yet it loses SQuAD completion to both (by 0.093 and 0.127), and RULER QA
to the higher-perplexity one by 0.215 [0.087, 0.348]. Its recall average is lower than both by about
0.04, which is not significant. On Llama-3.2-3B, the unrotated model and the rotated $\alpha = 1$ model,
0.08 apart in perplexity, are level on the commonsense average, recall and NIAH, and the unrotated model
is better on RULER QA. On Qwen3-4B, against the rotated $\alpha = 0.5$ model, recall is level, and the
unrotated model is better on NIAH (by 0.039) and worse on the commonsense average (by 0.022).

So perplexity's blind spot shows up on standard benchmarks for two of the four models, and in both the
commonsense average barely notices it. On Qwen3-1.7B the unrotated model trails the $\alpha = 0.5$ model
by 0.015 on the commonsense average, but by 0.078 on recall and 0.247 on RULER QA. On Gemma-2-2B it leads
both neighbours on the commonsense average while losing to them on SQuAD. We cannot say why these two
models show the gap and the others do not. The raw crest factor does not order them
(Table~\ref{tab:crest}).

\subsection{Adding 4-bit weights}
\label{sec:w4a4}

So far we have quantised activations only. A deployed W4A4 model also quantises its weights, so we add
GPTQ \citep{frantar2023gptq} INT4 weights in the same rotated basis as the activations, as QuaRot does.
We use per-output-channel scales, MSE-chosen clipping and act-order, with Hessians from the
full-precision model's inputs on 128 WikiText-2 training sequences of 2048 tokens. We then rerun the
probes and benchmarks (Table~\ref{tab:w4a4}).

\begin{table}[t]
\centering
\small
\setlength{\tabcolsep}{4pt}
\caption{Adding 4-bit weights (W4A4). Weights are GPTQ INT4 per output channel in the same rotated basis as the activations; activations are A4 as in the rest of Part~II. Probe accuracy over 2000 items (standard error about 0.011 for retrieval16); matched-noise rows replace the activation error of the $\alpha=1$ setting with noise of the same power, keeping the 4-bit weights. Perplexity is on the full WikiText-2 test set.}
\label{tab:w4a4}
\begin{tabular}{l l r rrr}
\toprule
Model & Setting & Perplexity & Induction & Retrieval16 & Retrieval64 \\
\midrule
Llama-3.2-3B & FP & 7.81 & 0.971 & 0.495 & 0.193 \\
 & W4 only & 8.11 & 0.967 & 0.472 & 0.148 \\
 & tuned ($\alpha=0.70$) & 8.36 & 0.961 & 0.443 & 0.137 \\
 & W4 + tuned ($\alpha=0.70$) & 8.74 & 0.955 & 0.382 & 0.092 \\
 & $\alpha=1$ & 8.84 & 0.954 & 0.419 & 0.127 \\
 & W4 + $\alpha=1$ & 9.27 & 0.944 & 0.377 & 0.079 \\
 & W4 + matched Gaussian ($\alpha=1$) & 9.37 & 0.944 & 0.391 & 0.087 \\
 & W4 + matched sign noise ($\alpha=1$) & 9.30 & 0.944 & 0.386 & 0.101 \\
\midrule
Qwen3-1.7B & FP & 9.39 & 0.978 & 0.442 & 0.116 \\
 & W4 only & 9.68 & 0.977 & 0.493 & 0.142 \\
 & tuned ($\alpha=0.70$) & 10.86 & 0.971 & 0.378 & 0.106 \\
 & W4 + tuned ($\alpha=0.70$) & 11.25 & 0.969 & 0.414 & 0.138 \\
 & $\alpha=1$ & 11.99 & 0.964 & 0.283 & 0.070 \\
 & W4 + $\alpha=1$ & 12.39 & 0.959 & 0.332 & 0.084 \\
 & W4 + matched Gaussian ($\alpha=1$) & 12.31 & 0.957 & 0.354 & 0.092 \\
 & W4 + matched sign noise ($\alpha=1$) & 12.44 & 0.957 & 0.346 & 0.094 \\
\midrule
Qwen3-4B & FP & 7.90 & 0.978 & 0.510 & 0.200 \\
 & W4 only & 8.09 & 0.977 & 0.464 & 0.166 \\
 & tuned ($\alpha=0.60$) & 10.29 & 0.973 & 0.488 & 0.203 \\
 & W4 + tuned ($\alpha=0.60$) & 10.58 & 0.970 & 0.423 & 0.165 \\
 & $\alpha=1$ & 11.82 & 0.968 & 0.455 & 0.148 \\
 & W4 + $\alpha=1$ & 12.17 & 0.966 & 0.395 & 0.137 \\
 & W4 + matched Gaussian ($\alpha=1$) & 10.54 & 0.962 & 0.327 & 0.090 \\
 & W4 + matched sign noise ($\alpha=1$) & 10.43 & 0.964 & 0.320 & 0.095 \\
 & W4 only (full Had.) & 8.08 & 0.978 & 0.481 & 0.181 \\
 & full Had., $\alpha=0.70$ & 8.68 & 0.975 & 0.505 & 0.193 \\
 & W4 + full Had., $\alpha=0.70$ & 8.90 & 0.973 & 0.465 & 0.175 \\
 & full Had., $\alpha=1$ & 9.15 & 0.973 & 0.426 & 0.146 \\
 & W4 + full Had., $\alpha=1$ & 9.41 & 0.971 & 0.401 & 0.132 \\
 & W4 + matched Gaussian, full Had. & 9.41 & 0.971 & 0.393 & 0.132 \\
 & W4 + matched sign noise, full Had. & 9.38 & 0.970 & 0.395 & 0.124 \\
\midrule
Gemma-2-2B & FP & 8.73 & 0.983 & 0.380 & 0.119 \\
 & W4 only & 8.94 & 0.980 & 0.359 & 0.100 \\
 & tuned ($\alpha=0.70$) & 9.27 & 0.978 & 0.353 & 0.102 \\
 & W4 + tuned ($\alpha=0.70$) & 9.53 & 0.975 & 0.297 & 0.081 \\
 & $\alpha=1$ & 9.67 & 0.974 & 0.303 & 0.072 \\
 & W4 + $\alpha=1$ & 9.98 & 0.968 & 0.282 & 0.056 \\
 & W4 + matched Gaussian ($\alpha=1$) & 10.00 & 0.970 & 0.304 & 0.074 \\
 & W4 + matched sign noise ($\alpha=1$) & 9.95 & 0.971 & 0.293 & 0.071 \\
\bottomrule
\end{tabular}
\end{table}

\paragraph{The weights add retrieval damage that perplexity barely shows.}
On Llama-3.2-3B, Qwen3-4B and Gemma-2-2B, adding 4-bit weights to the tuned A4 setting lowers retrieval16
by a further 0.055--0.066 (each interval excludes zero), while perplexity rises by only 0.26--0.38. On
Qwen3-4B with the full-width Hadamard, the weights cost 0.041 [0.023, 0.058] of retrieval16 and 0.22 of
perplexity, and the QA losses grow to match (NQ-Open from $-0.051$ to $-0.070$, Table~\ref{tab:qa}). The
closed-book losses grow on every model. Adding 4-bit weights to the tuned setting takes the TriviaQA loss
from $-0.059$ to $-0.133$ on Llama-3.2-3B, from $-0.124$ to $-0.133$ on Qwen3-4B with the block rotation,
and from $-0.075$ to $-0.093$ on Gemma-2-2B, each time with perplexity rising by under 0.4. Qwen3-1.7B is
the exception. Its 4-bit weights raise retrieval16, with or without quantised activations (by
0.036--0.051), while lowering every QA score (TriviaQA by 0.094 at W4A4 against 0.059 at A4). We do not
have an explanation for the probe gain. The natural tasks move in the expected direction.

\paragraph{Matched noise still reproduces most of the damage.}
If we keep the 4-bit weights and replace the activation error with matched noise, the noise reproduces
0.78--0.93 of the real error's retrieval16 damage on Llama-3.2-3B, Qwen3-1.7B and Gemma-2-2B, and
1.05--1.07 on Qwen3-4B with the full-width Hadamard. That is at least as much as without weight
quantisation (0.55--0.78 on Llama-3.2-3B and Gemma-2-2B, Section~\ref{sec:discussion}). The
activation-error findings therefore hold in a W4A4 model.

\subsection{A learned rotation}
\label{sec:spin}

The most widely deployed rotations are now learned \citep{liu2024spinquant,sun2024flatquant}. We train a
SpinQuant-style rotation in our setting. One rotation is shared by every input of width $d_{\text{model}}$
(the query, key, value, gate and up inputs of every layer, like SpinQuant's $R_1$). A second is shared by
every output-projection input (coarser than SpinQuant's per-layer $R_2$). Both start at the QuaRot
Hadamard, are parameterised as $R_0 \exp(A - A^\top)$, and are trained with Adam for 400 steps on the A4
model's next-token loss on WikiText-2 training windows, passing gradients straight through the rounding.
The down-projection keeps its fixed Hadamard, as SpinQuant keeps $R_4$ online. We keep the rotation with
the lowest held-out perplexity. This is our own reimplementation of the idea, not the SpinQuant release.

On Qwen3-1.7B at the tuned clipping ratio, the learned rotation lowers perplexity from 10.86 to 10.62, but
also lowers retrieval16 by 0.036 [0.018, 0.054] and retrieval64 by 0.029 [0.016, 0.042]. A rotation trained
on the language-modelling loss thus moves perplexity and retrieval in opposite directions. This is the
mismatch this paper is about, now produced by the optimiser itself. At $\alpha = 1$, training barely moves
perplexity (11.99 to 12.12 on the full test set) and raises retrieval16 by 0.029 [0.010, 0.048]. The regime
picture roughly holds. In the learned basis, matched Gaussian and sign-randomised noise reproduce 0.96
[0.82, 1.13] and 1.06 [0.90, 1.24] of the real error's retrieval16 damage, 0.68--0.70 of its retrieval64
damage, and 1.3--1.4 times its induction damage.

Qwen3-4B shows the same mismatch more strongly. At the tuned ratio, the learned rotation lowers perplexity
from 10.29 to 9.73 but lowers retrieval16 by 0.062 [0.043, 0.081] and retrieval64 by 0.066 [0.051, 0.081].
At $\alpha = 1$ it lowers perplexity slightly (11.82 to 11.67) and retrieval16 by 0.025 [0.005, 0.045].
Closed-book QA follows perplexity rather than the probes: TriviaQA rises by 0.026 [0.003, 0.049] and
NQ-Open is level, and on Qwen3-1.7B both are level. Because the learned rotation keeps the block-rotated
down-projection input, it also keeps the misleading noise result of Section~\ref{sec:partial}: matched
noise does 1.8--1.9 times the real error's retrieval16 damage, as under the block Hadamard. Learning a
rotation for the other inputs does not remove the need to rotate that input in full.

\subsection{Scale}
\label{sec:scale}
\label{sec:scale-old}

In Part~I, Qwen2.5-32B keeps nearly all of its induction accuracy (0.983) at three bits, where the
four-model set had already collapsed, and breaks down at two (0.086 under uniform quantisation against
0.868 under matched noise). Instruction-tuned Qwen3-8B collapses between four and three bits (0.940 to
0.007), and rotation recovers induction there too (Appendix~\ref{app:scale}). In Part~II, Qwen3-8B-Base,
all of whose widths QuaRot can rotate in one piece, is in the full-rotation regime as expected. Every
rotated input has a crest factor of 3.6--3.9 (Table~\ref{tab:crest}), the deadzone accounts for 0.08 of the
induction damage, and matched noise reproduces the real error's damage on both probes (1.16 and 1.11;
Table~\ref{tab:k1}). On one seed, the clipping ratio perplexity picks for Qwen3-8B ($\alpha = 0.80$) is
within the confidence interval of the retrieval-optimal ratio on both retrieval16 (gap 0.011 [$-0.020$,
0.042]) and retrieval64 (0.018 [$-0.001$, 0.037]). With a single seed, we leave it out of
Table~\ref{tab:clip}.

\section{Discussion}
\label{sec:discussion}

\paragraph{Order versus size.}
The two parts of the paper fit together once we separate the order of quantisers from the size of their
damage. Within a model and a regime, perplexity orders quantisers correctly, both by their damage to
probes (Part~I) and to natural tasks (Section~\ref{sec:natural}). What it does not tell you is how much
is lost, and it falls furthest short on what running text weighs least: rarely stored facts and retrieval
from the context. Across regimes, the order itself can fail. Without rotation, the error is structured,
mostly signal deleted by the deadzone, and it hurts in-context computation more than its size or its
perplexity cost suggests. A full rotation removes most of that structure, and the error then behaves much
more like rounding noise of the same power. The risky cases are in between: incomplete rotations, or
unrotated inputs left in an otherwise rotated pipeline. The learned rotation shows that the two measures
can be traded against each other directly. Trained for perplexity, it gives up retrieval
(Section~\ref{sec:spin}).

\paragraph{Practical advice.}
\begin{itemize}
\item After rotation, check the crest factor of every quantised input. If it is well above the Gaussian
value for its width (about 4), the input is not fully rotated. This is a screening check, not a
guarantee: a Gaussian-like crest factor did not rule out a retrieval gap on Llama-3.2-3B or Gemma-2-2B.
\item If a library cannot rotate a width in one piece, look for a Kronecker-product Hadamard ($H_m \otimes
H_{2^k}$, with $H_m$ from a Paley construction) before falling back to blocks.
\item Do not use stochastic rounding for activations.
\item Tune the clipping ratio rather than leaving $\alpha = 1$. After a full rotation, perplexity is a
sound way to choose it.
\item Report a closed-book QA score next to perplexity. It is cheap, its relative change is at least as
large as perplexity's and up to twice as large, and it measures the knowledge quantisation removes first.
\item Do not compare rotations with random-token retrieval probes alone. They rank natural-task damage
less well than perplexity does, and they miss gains on natural text.
\end{itemize}

\paragraph{What does not fit.}
Llama-3.2-3B and Gemma-2-2B keep a gap on retrieval even under full rotation, with the real error worse
than noise (matched noise reproduces 0.55--0.78 of the damage). On Gemma, matched noise also falls short
on induction (0.81). So ``size decides'' holds for induction on four of the five models, and only roughly
for retrieval. Retrieval alone also does not separate the regimes (Section~\ref{sec:norot}).

\section{Limitations}
\label{sec:limitations}

\textbf{Scope.} Part~I perturbs one residual-stream block (end to end only as a variant), with four
decoder-only models running the full set of probes and retrieval resolved to
$\pm\HalfwidthRetrieval{}$. Part~II covers five models up to 8B, with natural-task evaluation on four of
them (up to 4B). We did not run the natural tasks on 8B or larger models, and on Qwen2.5-7B we ran only
closed-book QA and recall, to test the fix. The downstream benefit of the full-width rotation is resolved
on Qwen3-4B TriviaQA on each of three rotation seeds. On Qwen3-4B NQ-Open and Qwen2.5-7B TriviaQA it is
resolved only when the three seeds are pooled, and other natural-task comparisons use one seed. On
Qwen2.5-0.5B, whose block-rotated input is already near Gaussian, there is no resolved difference at
either clipping ratio we ran (Section~\ref{sec:partial-q25}). For four more affected models we measure
only the block-rotated input (Table~\ref{tab:widths}), not downstream accuracy. The decomposition of the
block-rotated input is on Qwen3-4B only. The attention-margin test is on one model, and the circuit
analysis, including path patching, is on three, with one rotation seed throughout.

\textbf{Tasks.} Closed-book QA uses 1000 questions per dataset and LongBench 100 items per task, so small
LongBench differences are not resolved. Gemma-2-2B's LongBench inputs exceed its context, and two models
score near the floor on DROP at full precision. Our measure of how firmly a fact is stored (how many other
models answer it) is coarse, and the matched-noise comparison on closed-book QA is on two models.

\textbf{Quantiser.} Activations use per-token RTN. Weights are either full precision or GPTQ INT4 in the
rotated basis (Section~\ref{sec:w4a4}). No run uses an integer kernel. Our learned rotation is a
SpinQuant-style reimplementation (shared rotations, 400 steps), not the SpinQuant or FlatQuant release,
and we ran it on two models.

\textbf{Diagnostic versus deployable evidence.} Feature restoration in Part~I is a perplexity-only
oracle. All latency figures use plain PyTorch matmuls rather than fused kernels. They compare rotations
with each other but do not predict end-to-end serving latency.

\textbf{Crest factor.} We use the crest factor as a warning sign, not as a validated predictor. It only
summarises the largest value. Two inputs with the same crest factor can put different fractions of their
values, and of their energy, in the deadzone, and it does not predict which models show the benchmark gap
(Section~\ref{sec:bench}).

\textbf{Statistics.} Confidence intervals are paired bootstraps over items. They do not include variation
over rotation seeds, models or task families, so on their own they do not show that a result generalises
across them. We report such variation separately where we have it. Most natural-task results use one
rotation seed (the full-width fix's closed-book QA results use three). The ranking analysis of
Section~\ref{sec:qa} resamples settings within models.

\paragraph{Reproducibility.}
\label{sec:repro}
The code release contains the activation hooks and quantiser (Equation~\ref{eq:quantiser}), every
rotation we use (QuaRot's, block-diagonal, dense, and $H_{76} \otimes H_{128}$ with its tests), the probe
generators with their seeds, the lm-evaluation-harness driver and task lists, the per-seed and per-item
results behind every table, and the scripts that build the tables, figures and benchmark report from
them. Every model load is checked against a recorded full-precision perplexity. The anonymised release
is available at
\url{https://anonymous.4open.science/r/Perplexity-Cost-Understates-What-Activation-Quantisation-Breaks-907A}
and as supplementary material, and will be made public on acceptance.

\section{Related Work}
\label{sec:related}

\textbf{Compression beyond perplexity.} Perplexity and zero-shot suites under-report compression damage
to knowledge-heavy and in-context tasks \citep{jaiswal2024compressing,dutta2024accuracy}. Closest to us,
\citet{duan2026quantization} show that perplexity can miss damage to sparse-autoencoder features under
quantisation. We add three things: controls that separate the error's structure from its size (matched
Gaussian and sign-randomised noise at every setting); a split into regimes by the input's crest factor,
which says when the blind spot appears and when it does not; and the deployed setting (rotated A4 and
W4A4, end to end), tested on natural question answering as well as probes.

\textbf{Activation outliers and rotations.} Outliers limit low-bit activation quantisation
\citep{dettmers2022llmint8,xiao2023smoothquant,wei2023outlier,yuan2023rptq}. Weight and activation methods
select salient channels or reconstruct locally \citep{frantar2023gptq,lin2024awq}, and others allocate
precision by sensitivity \citep{kim2024squeezellm,heo2024rethinking}. Rotations remove privileged
coordinates \citep{ashkboos2024quarot,liu2024spinquant,liu2024duquant,sun2024flatquant,shao2024omniquant}.
We show what rotation changes about the error, and where block-diagonal versions fall short.

\textbf{Quantisation noise.} The additive-noise model and the conditions under which it fails go back to
\citet{bennett1948spectra}. \citet{li2026mxfp4} split MXFP4 quantisation error in reinforcement-learning
training into a reducible bias, a recoverable deadzone and an irreducible floor. Our deadzone is that of
per-token integer activation quantisation at inference. We measure what it does to in-context computation
rather than to a training objective, show that full rotation removes it, and find that correcting its gain
does not recover the damage in our setting (Section~\ref{sec:rot}).

\textbf{Interpretability and similarity.} Sparse dictionaries recover features from residual activations
\citep{bricken2023monosemanticity,cunningham2024sae,gao2024scaling}. Our CKA result is an example of the
failure that critiques of similarity measures warn about \citep{kornblith2019cka,ding2021grounding}.
Induction and copying circuits \citep{olsson2022induction} motivate our probes.

\section{Conclusion}
\label{sec:conclusion}

Perplexity ranks activation quantisers correctly but understates what they remove, and what they remove
first is long-tail knowledge and retrieval from the context. The ranking is reliable while the error
behaves like noise, as it does after a full rotation. Where an input is unrotated, or rotated in blocks
that leave it far from Gaussian, the error is signal deleted by the rounding deadzone, and perplexity can
misjudge it. Three cheap practices help: check each input's crest factor and deadzone energy, use a
full-width rotation where the block rotation leaves an input outlier-heavy, and report a knowledge-heavy
task next to perplexity. A full-width Hadamard exists for every block-rotated width in our survey of
popular open models. On Qwen3-4B it halves the closed-book TriviaQA loss on each of three rotation seeds,
and on Qwen2.5-7B it lowers it from 0.038 to 0.018 (pooled over seeds; not resolved on two of the three
seeds alone). On Qwen2.5-0.5B, whose
block-rotated input is already near Gaussian, it changes little. The other two practices make visible the
damage that perplexity understates.

\bibliography{refs}
\bibliographystyle{tmlr}

\newpage
\appendix
\section{Part II: full outlier statistics and per-seed results}
\label{app:part2}

Table~\ref{tab:crestfull} gives the full outlier statistics behind Table~\ref{tab:crest}: per-token
crest factor (mean and 99th percentile over tokens, averaged over layers, attention-sink position
excluded), excess kurtosis, and the relative error power of per-token A4 RTN with $\alpha = 1$,
$\mathbb{E}\|\varepsilon\|^2 / \mathbb{E}\|h\|^2$, measured on retrieval16 inputs. WikiText-2 inputs
give the same picture (largest difference: Qwen3-4B's block-rotated down-projection input, crest
6.07 and kurtosis 3.75 on WikiText). ``dense'' is a dense random orthogonal rotation, computed only
for the one width QuaRot's construction does not cover.

\begingroup
\small
\setlength{\tabcolsep}{4pt}
\begin{longtable}{llrrrrr}
  \caption{Linear-input outlier statistics, retrieval16 inputs.}
  \label{tab:crestfull}\\
  \toprule
  Model & Input & Rotation & Crest (mean) & Crest (p99) & Excess kurtosis & Rel.\ A4 error power \\
  \midrule
  \endfirsthead
  \toprule
  Model & Input & Rotation & Crest (mean) & Crest (p99) & Excess kurtosis & Rel.\ A4 error power \\
  \midrule
  \endhead
Llama-3.2-3B & q/k/v & raw & 12.00 & 17.63 & 24.41 & 21.5\% \\
Llama-3.2-3B & q/k/v & hadamard & 3.71 & 4.58 & -0.02 & 2.4\% \\
Llama-3.2-3B & o & raw & 10.83 & 17.27 & 21.08 & 14.5\% \\
Llama-3.2-3B & o & hadamard & 3.70 & 4.59 & -0.01 & 2.3\% \\
Llama-3.2-3B & gate/up & raw & 7.46 & 11.37 & 4.45 & 10.0\% \\
Llama-3.2-3B & gate/up & hadamard & 3.73 & 4.65 & -0.00 & 2.4\% \\
Llama-3.2-3B & down & raw & 27.18 & 56.25 & 222.17 & 37.8\% \\
Llama-3.2-3B & down & hadamard & 3.88 & 4.72 & -0.06 & 2.6\% \\
Qwen3-1.7B-Base & q/k/v & raw & 19.71 & 26.69 & 139.99 & 24.9\% \\
Qwen3-1.7B-Base & q/k/v & hadamard & 3.46 & 4.26 & -0.14 & 2.1\% \\
Qwen3-1.7B-Base & o & raw & 9.57 & 14.62 & 15.73 & 11.7\% \\
Qwen3-1.7B-Base & o & hadamard & 3.58 & 4.46 & -0.02 & 2.2\% \\
Qwen3-1.7B-Base & gate/up & raw & 15.73 & 25.26 & 70.74 & 29.6\% \\
Qwen3-1.7B-Base & gate/up & hadamard & 3.55 & 4.43 & -0.06 & 2.2\% \\
Qwen3-1.7B-Base & down & raw & 28.28 & 51.90 & 333.23 & 24.3\% \\
Qwen3-1.7B-Base & down & hadamard & 3.73 & 4.55 & -0.11 & 2.4\% \\
Qwen3-4B-Base & q/k/v & raw & 19.73 & 27.36 & 132.47 & 26.7\% \\
Qwen3-4B-Base & q/k/v & hadamard & 3.50 & 4.34 & -0.10 & 2.1\% \\
Qwen3-4B-Base & o & raw & 10.33 & 16.12 & 16.84 & 12.8\% \\
Qwen3-4B-Base & o & hadamard & 3.77 & 4.64 & -0.01 & 2.4\% \\
Qwen3-4B-Base & gate/up & raw & 16.02 & 23.83 & 77.87 & 29.8\% \\
Qwen3-4B-Base & gate/up & hadamard & 3.61 & 4.48 & -0.05 & 2.2\% \\
Qwen3-4B-Base & down & raw & 33.07 & 66.41 & 458.67 & 33.9\% \\
Qwen3-4B-Base & down & hadamard & 6.98 & 9.53 & 5.86 & 8.0\% \\
Qwen3-4B-Base & down & dense & 4.01 & 4.86 & -0.00 & 2.8\% \\
Qwen3-8B-Base & q/k/v & raw & 25.78 & 34.42 & 228.39 & 29.4\% \\
Qwen3-8B-Base & q/k/v & hadamard & 3.61 & 4.37 & -0.13 & 2.2\% \\
Qwen3-8B-Base & o & raw & 10.31 & 16.33 & 17.40 & 12.9\% \\
Qwen3-8B-Base & o & hadamard & 3.78 & 4.66 & -0.00 & 2.5\% \\
Qwen3-8B-Base & gate/up & raw & 21.59 & 30.67 & 228.36 & 35.5\% \\
Qwen3-8B-Base & gate/up & hadamard & 3.65 & 4.52 & -0.12 & 2.3\% \\
Qwen3-8B-Base & down & raw & 38.63 & 75.91 & 764.95 & 34.0\% \\
Qwen3-8B-Base & down & hadamard & 3.87 & 4.70 & -0.13 & 2.6\% \\
  \bottomrule
\end{longtable}
\endgroup

\begin{figure}[htbp]
  \centering
  \includegraphics[width=\linewidth]{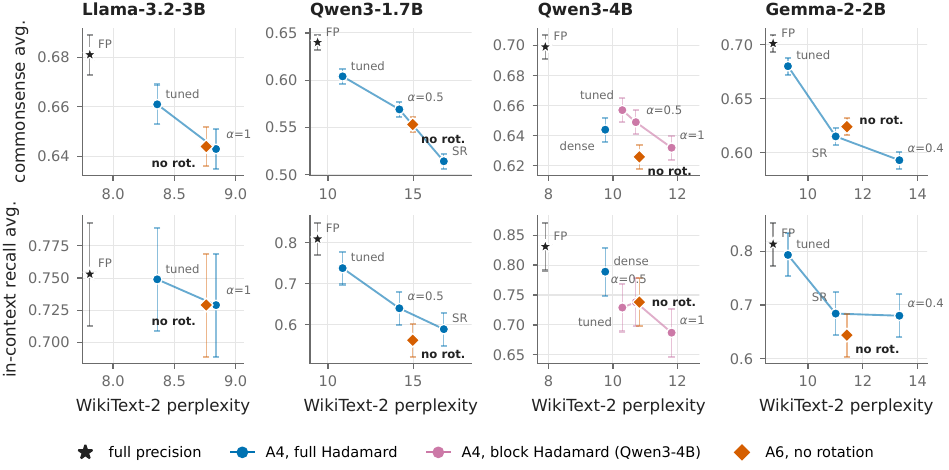}
  \caption{Standard benchmarks against perplexity (raw scores behind Figure~\ref{fig:bench}). Each point is one setting (Table~\ref{tab:bench});
  bars are 95\% intervals. Rotated settings are joined in order of perplexity. Top: on the commonsense
  average, the unrotated model (orange diamond) sits on or above the rotated line for three models
  and below it for Qwen3-4B.
  Bottom: on in-context recall, it falls below the line on Qwen3-1.7B and Gemma-2-2B, and on it for
  Llama-3.2-3B and Qwen3-4B. On Qwen3-4B, the densely rotated model recalls more than every
  block-rotated setting.}
  \label{fig:bench-raw}
\end{figure}

\section{Where the Damage Flows: Collective Retrieval, Sparse Induction}
\label{sec:circuit}

We use activation patching~\citep{meng2022locating,wang2023interpretability,heimersheim2024use} to ask where
the damage passes through the model: restore components of the quantised model to their full-precision
values and measure how much of the target log-probability gap closes, as
$(\overline{\ell}_{\text{patched}} - \overline{\ell}_{\text{A4}})/(\overline{\ell}_{\text{FP}} -
\overline{\ell}_{\text{A4}})$ with paired item-bootstrap 95\% intervals. The settings are rotated A4 at
$\alpha = 1$ (Qwen3-4B with the full-width Hadamard), matched Gaussian noise, and A6 without rotation, on
Qwen3-4B, Llama-3.2-3B and Qwen3-1.7B. Heads are ranked by attribution patching~\citep{syed2023attribution}
on 500 calibration items, and every restoration is measured exactly on 500 separate test items
(Appendix~\ref{app:circuit}). One natural account of retrieval failure, that noise in the attention logits
of the retrieval heads exceeds the margin to the strongest distractor, does not hold: a margin-to-noise model
fixed in advance explains almost none of the item-level failures (McFadden pseudo-$R^2$ 0.010;
Appendix~\ref{sec:margin}).

\begin{figure}[htbp]
  \centering
  \includegraphics[width=\linewidth]{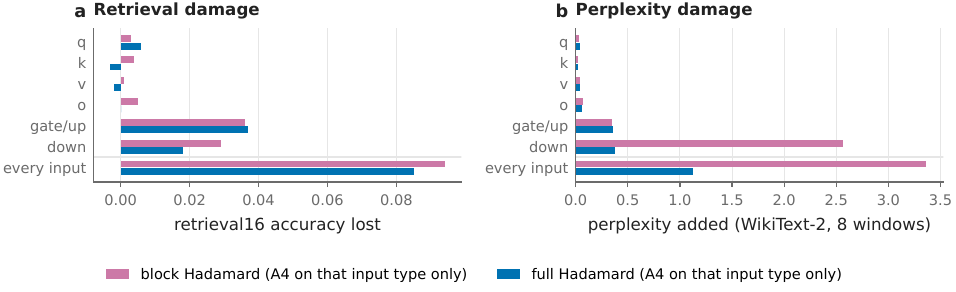}
  \caption{Perplexity damage and retrieval damage come from different inputs (Qwen3-4B, A4,
  $\alpha = 1$, only the named input type quantised in every layer). (a) Retrieval16 accuracy lost.
  (b) Perplexity added. The block-rotated down-projection input carries most of the perplexity cost
  and the full-width rotation removes it, but retrieval loses as much with either rotation and the
  loss comes mostly from the MLP inputs.}
  \label{fig:circuit-localise}
\end{figure}

\paragraph{Perplexity damage and retrieval damage enter at different inputs.}
Quantising one input type at a time on Qwen3-4B (Figure~\ref{fig:circuit-localise}, Appendix~\ref{app:circuit}),
the block-rotated down-projection input alone raises perplexity from 6.75 to 9.32, most of the full A4 cost
(10.11), and the full-width rotation removes that cost, yet retrieval16 is about as low under either rotation
(0.425 and 0.434, against 0.519). On all three models the MLP inputs carry most of the perplexity cost; the
retrieval loss is spread over input types (on Llama-3.2-3B no input type alone costs more than 0.017 of the
0.060 total).

\begin{figure}[t]
  \centering
  \includegraphics[width=\linewidth]{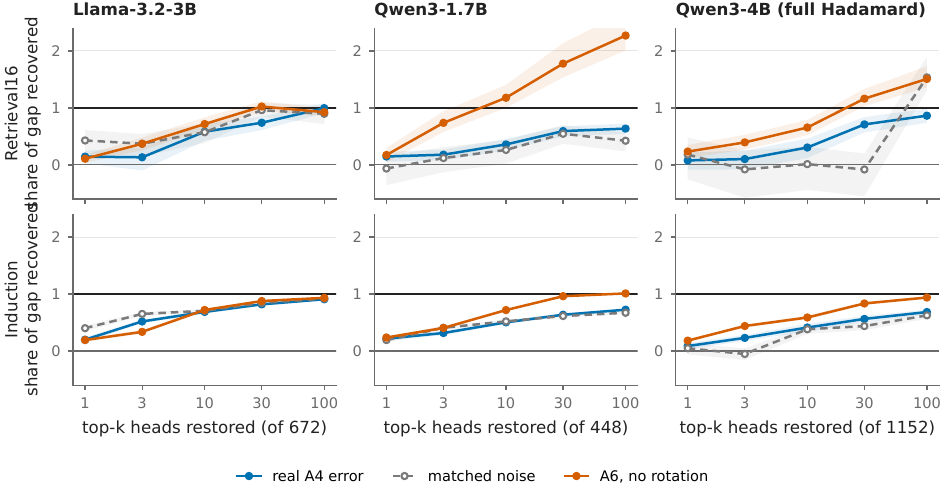}
  \caption{Restoring the top-$k$ heads on three models (rotated A4 at $\alpha = 1$, matched Gaussian
  noise, and A6 without rotation; Qwen3-4B with the full-width Hadamard). Share of the target log-prob
  gap recovered, with 95\% bootstrap bands. On every model a few dozen heads carry most of the damage. The
  separation between real error and matched noise on Qwen3-4B's retrieval does not appear on Llama-3.2-3B
  or Qwen3-1.7B.}
  \label{fig:circuit-xmodel}
\end{figure}

\paragraph{A few dozen heads carry the damage, for noise as for the real error.}
Restoring the top 30 attention heads (of 1152, 672 and 448) recovers 0.71 [0.57, 0.84], 0.74 [0.61, 0.85]
and 0.59 [0.50, 0.68] of the real error's retrieval gap on Qwen3-4B, Llama-3.2-3B and Qwen3-1.7B
(Figure~\ref{fig:circuit-xmodel}). Matched noise concentrates in the same heads on Llama-3.2-3B and Qwen3-1.7B
(0.96 and 0.55); only on Qwen3-4B does it not ($-0.08$, with a wide interval). The retrieval heads are where an
activation perturbation of this size does its damage, whether it is quantisation error or noise, consistent
with the error behaving much like noise after full rotation (Section~\ref{sec:rot}). The structured error of
an unrotated model is the exception: restoring the top 30 heads overshoots full precision (1.78 [1.53, 2.11]
on Qwen3-1.7B), as it does under Qwen3-4B's block rotation.

\begin{figure}[t]
  \centering
  \includegraphics[width=\linewidth]{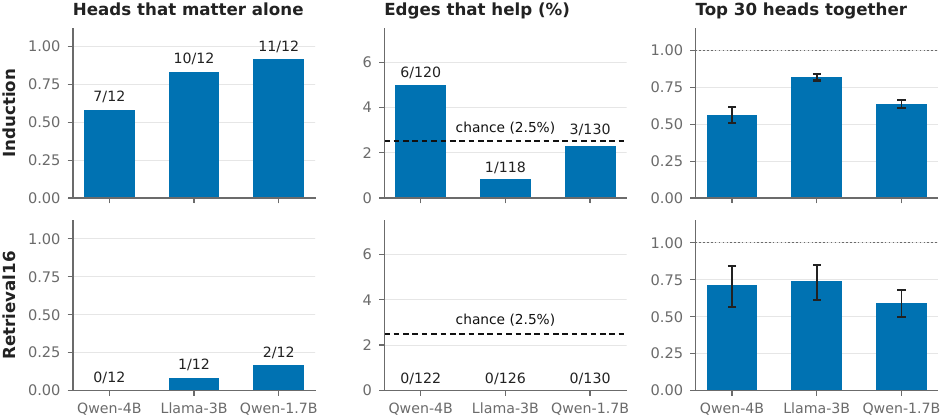}
  \caption{Induction has a circuit; retrieval does not. Path patching among the 12 heads with the largest
  calibration attribution per task, on three models (A4, $\alpha = 1$, full Hadamard; Qwen3-4B with
  $H_{76} \otimes H_{128}$). Left: share of those heads whose restoration alone recovers part of the gap
  (95\% interval above zero). Middle: share of head-to-head edges (query/key or value input of a later top
  head) whose restoration alone helps; the dashed line is the 2.5\% expected by chance at this threshold.
  Right: share of the gap recovered by restoring the top 30 heads together, with 95\% intervals; the dotted
  line is full recovery. Retrieval damage is recovered only collectively. Circuit diagrams are in
  Figures~\ref{fig:circuit-diagram} and~\ref{fig:circuit-diagram-xmodel} (Appendix~\ref{app:circuit}).}
  \label{fig:circuit-summary}
\end{figure}

\paragraph{Retrieval damage is collective; induction has a sparse circuit.}
Path patching~\citep{wang2023interpretability,goldowskydill2023localizing} between the 12 top heads for each
task delivers a sender head's full-precision output to one receiver only (the query and key, or the value, of
a later top head, or the final norm), with the rest of the model left quantised (Figure~\ref{fig:circuit-summary}).
On retrieval the answer is the same on all three models. No head-to-head edge recovers anything (0 of 122, 126
and 130 edges have a positive interval, and a few are significantly negative), almost no head matters on its own
(0, 1 and 2 of 12), and the only positive paths are direct writes to the logits (6, 4 and 5 of 12). Yet the same
heads restored together recover 0.59--0.74 of the gap. Induction is the contrast: 7, 10 and 11 of its 12 top heads
matter individually, and on Qwen3-4B and Qwen3-1.7B the significant edges run into the queries and keys of later
heads, the composition pattern of induction heads~\citep{olsson2022induction} (on Qwen3-1.7B, L5H9 into L6H7;
Figure~\ref{fig:circuit-diagram-xmodel}). The node-level contrast is the robust result. Individual edges are
not: 6, 1 and 3 induction edges have a positive interval (of 120, 118 and 130), against about 3 expected by
chance at an uncorrected 95\% threshold, and Llama-3.2-3B's one edge (L2H15 into the queries and keys of
L14H22) is the only one there. We read the edges as suggestive of query--key composition, not as a mapped circuit.
Retrieval failure under quantisation is not carried by any one path that could be patched; it appears when many
heads are perturbed at once, which is why a per-head margin model does not predict it.

\begin{figure}[htbp]
  \centering
  \includegraphics[width=\linewidth]{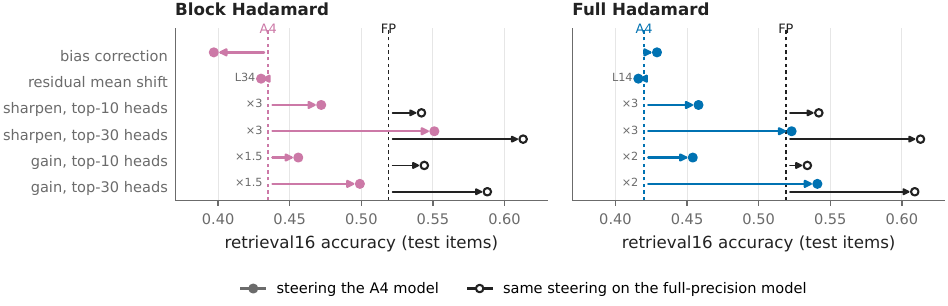}
  \caption{Cheap repairs on retrieval16 (Qwen3-4B, A4, $\alpha = 1$). Coloured arrows: from the A4
  model's accuracy to its accuracy with the repair. Black arrows: the same repair applied to the
  full-precision model. Strengths and layers are chosen on calibration items (label beside each
  point); accuracy is on test items. Steering the retrieval heads helps the A4 model, but it helps
  the full-precision model about as much, so it is a general retrieval boost rather than a repair of
  quantisation damage.}
  \label{fig:circuit-steering}
\end{figure}

\begin{figure}[htbp]
  \centering
  \includegraphics[width=\linewidth]{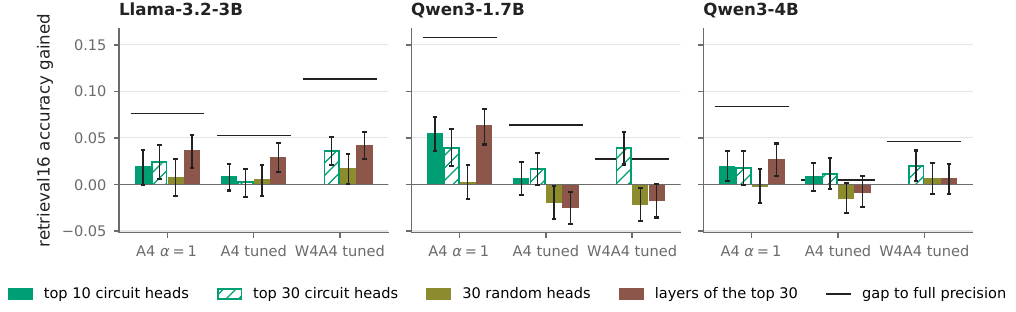}
  \caption{Retrieval16 accuracy gained by keeping attention heads at 8-bit activations, on top of A4
  ($\alpha = 1$ and the tuned ratio) and W4A4. Circuit heads are the top-$k$ by calibration attribution;
  random heads use the same budget; ``layers'' puts every attention input of the layers containing the
  top 30 at 8 bits. Bars are paired differences with 95\% intervals; the black line is the gap to full
  precision. Natural-QA effects are in Table~\ref{tab:headmix}.}
  \label{fig:headmix}
\end{figure}

\paragraph{No targeted repair works.}
Consistent with collective damage, the targeted repairs we tried fail, with every strength chosen on calibration
items. Steering the 30 heads with the most full-precision attention on the answer raises A4 retrieval16 on
every model (Qwen3-4B 0.420 to 0.541, Llama-3.2-3B 0.431 to 0.593, Qwen3-1.7B 0.277 to 0.342), but the same
steering raises the full-precision model about as much (to 0.609, 0.659 and 0.465), so the gap to full precision
remains: it is a lever on retrieval, not a repair of quantisation (Figure~\ref{fig:circuit-steering},
Appendix~\ref{app:circuit}). Subtracting the mean error and shifting the residual stream do not help. Keeping the
top heads at 8-bit activations recovers part of the probe gap at $\alpha = 1$ (top 10 heads: 34\% of it on
Qwen3-1.7B with 0.7\% of multiply--accumulates at 8 bits, 25\% on Llama-3.2-3B with 0.6\%, 23\% on Qwen3-4B with
0.3\%). The same budget on random heads recovers nothing on the Qwen models and about half as much as the circuit
heads on Llama-3.2-3B (significant only in the W4A4 model: 15\% against 32\% of the gap), but no head-mixing variant changes TriviaQA, NQ-Open or LongBench beyond its
interval (Figure~\ref{fig:headmix} and Table~\ref{tab:headmix}, Appendix~\ref{app:circuit}). The closed-book
losses of Section~\ref{sec:longtail} are losses of stored knowledge, which protecting retrieval heads does not
reach.

\paragraph{Caveats.}
One rotation seed throughout; path patching uses the 12 top heads per task, so it can miss structure among
lower-ranked heads. Restoring a head restores its output before the output projection, whose own input is
still quantised. Under the block rotation the quantised model is numerically fragile (individual items move by
up to 9 nats across GPUs), so every comparison is made within one run. Edge recoveries do not add up to group
recoveries; we read them as evidence about structure, not as a decomposition.

\subsection{Details and full results}
\label{app:circuit}

\paragraph{Probes and items.}
All circuit experiments use the synthetic retrieval16 and induction probes of Section~\ref{sec:setup}.
Anything that is chosen (head rankings, steering strengths, steering layers, mean-error and mean-shift
vectors) is fitted on 500 calibration items generated with a separate seed; all reported numbers are on
500 (patching, path patching) or 1000 (localisation, steering) test items. The metric is the mean
log-probability of the target tokens; accuracy is reported alongside.

\paragraph{Attribution and exact patching.}
Heads are ranked by attribution patching, $(a_{\text{FP}} - a_{\text{Q}})\cdot\partial\ell/\partial a$
evaluated at the quantised point, where $a$ is a head's slice of the output-projection input and the
gradient passes through the quantiser as a straight-through estimator. We use the attribution only as a
ranking: its per-head values add to more than the measured gap (they are not additive), and every claim
in this appendix is from exact patching. A unit test checks that the attribution matches exact
patching when only one layer is quantised and the perturbation is small. Exact head patching replaces
the head's output-projection input with its full-precision value; pattern-only and value-only patching
replace the head's attention probabilities or its (GQA-expanded) value vectors.

\paragraph{Path patching.}
For sender head $U$ in layer $l_U$ we compute $\delta_U = W_O^{(U)}(a^{\text{FP}}_U - a^{\text{Q}}_U)$, the
change a full-precision $U$ would write into the residual stream, and add it only to the chosen receivers:
the input of a later head's query and key projections, or of its value projection (each receiver
recomputes its RMSNorm and re-quantises the patched input, and for keys and values only the receiver's
own copy of its GQA group is replaced); the input of the final norm (direct path); every MLP input from
layer $l_U$ on; or all of these at once, which a unit test checks equals restoring $U$ outright. Downstream
computation then runs on the quantised model as usual. We use the 12 heads with the largest calibration
attribution as senders and receivers, giving 166--168 conditions per task.

\paragraph{Steering.}
Bias correction subtracts, from every quantised linear input, the mean quantisation error measured on 16
WikiText-2 calibration windows (position 0 separately); it is deployable as a bias $W\mu$. The residual
mean shift adds the calibration mean of $h_{\text{FP}} - h_{\text{Q}}$ after one layer, chosen on
calibration items. Sharpening scales the attention logits of the $k \in \{10, 30\}$ heads with the most
full-precision attention mass on the answer position; gain scales those heads' outputs. For the chosen
strength we also apply the same steering to the full-precision model and measure WikiText-2 perplexity
(8 windows of 2048 tokens).

\begin{table}[h]
\centering
\scriptsize
\setlength{\tabcolsep}{3.5pt}
\caption{Exact head patching on Qwen3-4B (500 test items per task). Share of the target log-prob gap recovered by restoring the top-$k$ heads (ranked by attribution on 500 calibration items), and, for the top 30, only their attention pattern or only their values; 95\% paired bootstrap intervals.}
\label{tab:circuit-patching}
\begin{tabular}{lcccccc}
\toprule
setting & acc FP$\to$Q & top 10 & top 30 & top 100 & pattern, top 30 & values, top 30 \\
\midrule
\multicolumn{7}{l}{\textit{Retrieval16}} \\
A4, block Hadamard & 0.524$\to$0.456 & 1.31 [0.76, 3.37] & 2.07 [1.31, 6.44] & 3.63 [2.24, 10.64] & 0.95 [0.39, 2.10] & 1.77 [1.09, 4.75] \\
noise, block Hadamard & 0.524$\to$0.364 & 0.26 [0.04, 0.44] & 0.33 [0.14, 0.49] & 0.78 [0.65, 0.90] & -0.12 [-0.39, 0.09] & 0.04 [-0.23, 0.24] \\
A4, full Hadamard & 0.524$\to$0.416 & 0.30 [0.10, 0.46] & 0.71 [0.57, 0.84] & 0.86 [0.74, 0.98] & 0.47 [0.27, 0.63] & 0.24 [-0.01, 0.42] \\
noise, full Hadamard & 0.524$\to$0.446 & 0.01 [-0.42, 0.30] & -0.08 [-0.51, 0.20] & 1.54 [1.32, 1.88] & -0.60 [-1.27, -0.20] & 0.34 [-0.02, 0.64] \\
A6, no rotation & 0.524$\to$0.372 & 0.66 [0.53, 0.78] & 1.16 [1.04, 1.34] & 1.51 [1.37, 1.73] & 0.86 [0.72, 1.00] & 0.68 [0.52, 0.85] \\
\multicolumn{7}{l}{\textit{Induction}} \\
A4, block Hadamard & 0.978$\to$0.969 & 0.49 [0.43, 0.54] & 0.67 [0.61, 0.72] & 0.79 [0.75, 0.84] & 0.29 [0.22, 0.35] & 0.51 [0.46, 0.57] \\
noise, block Hadamard & 0.978$\to$0.967 & 0.31 [0.26, 0.37] & 0.40 [0.34, 0.45] & 0.50 [0.45, 0.54] & 0.13 [0.06, 0.19] & 0.27 [0.21, 0.32] \\
A4, full Hadamard & 0.978$\to$0.972 & 0.41 [0.34, 0.48] & 0.56 [0.50, 0.61] & 0.68 [0.64, 0.73] & 0.27 [0.20, 0.34] & 0.44 [0.38, 0.51] \\
noise, full Hadamard & 0.978$\to$0.972 & 0.38 [0.30, 0.45] & 0.44 [0.36, 0.50] & 0.63 [0.57, 0.67] & 0.26 [0.17, 0.34] & 0.41 [0.34, 0.48] \\
A6, no rotation & 0.978$\to$0.960 & 0.59 [0.55, 0.62] & 0.83 [0.81, 0.86] & 0.94 [0.92, 0.96] & 0.44 [0.40, 0.48] & 0.63 [0.60, 0.67] \\
\bottomrule
\end{tabular}
\end{table}

\begin{table}[h]
\centering
\scriptsize
\setlength{\tabcolsep}{3.5pt}
\caption{Steering the 30 heads with the most full-precision attention on the answer, retrieval16 test accuracy as A4 model / full-precision model with the same steering. Strength chosen on calibration items within the main grid (sharpen $\times$0.8--3, gain $\times$0.8--4) and within the extended grid (sharpen $\times$3--8, gain $\times$4--8).}
\label{tab:circuit-steer}
\begin{tabular}{lcccccc}
\toprule
setting & FP & Q & sharpen & sharpen, ext. & gain & gain, ext. \\
\midrule
A4, block Hadamard & 0.519 & 0.435 & $\times$3: 0.551 / 0.613 & $\times$4: 0.536 / 0.616 & $\times$1.5: 0.499 / 0.588 & $\times$4: 0.012 / 0.020 \\
noise, block Hadamard & 0.519 & 0.343 & $\times$3: 0.470 / 0.613 & $\times$8: 0.471 / 0.614 & $\times$2: 0.464 / 0.609 & $\times$4: 0.000 / 0.020 \\
A4, full Hadamard & 0.519 & 0.420 & $\times$3: 0.523 / 0.613 & $\times$8: 0.537 / 0.615 & $\times$2: 0.541 / 0.609 & $\times$4: 0.017 / 0.020 \\
noise, full Hadamard & 0.519 & 0.433 & $\times$3: 0.541 / 0.613 & $\times$8: 0.537 / 0.614 & $\times$2: 0.526 / 0.608 & $\times$4: 0.006 / 0.020 \\
\bottomrule
\end{tabular}
\end{table}

\begin{figure}[htbp]
  \centering
  \includegraphics[width=\linewidth]{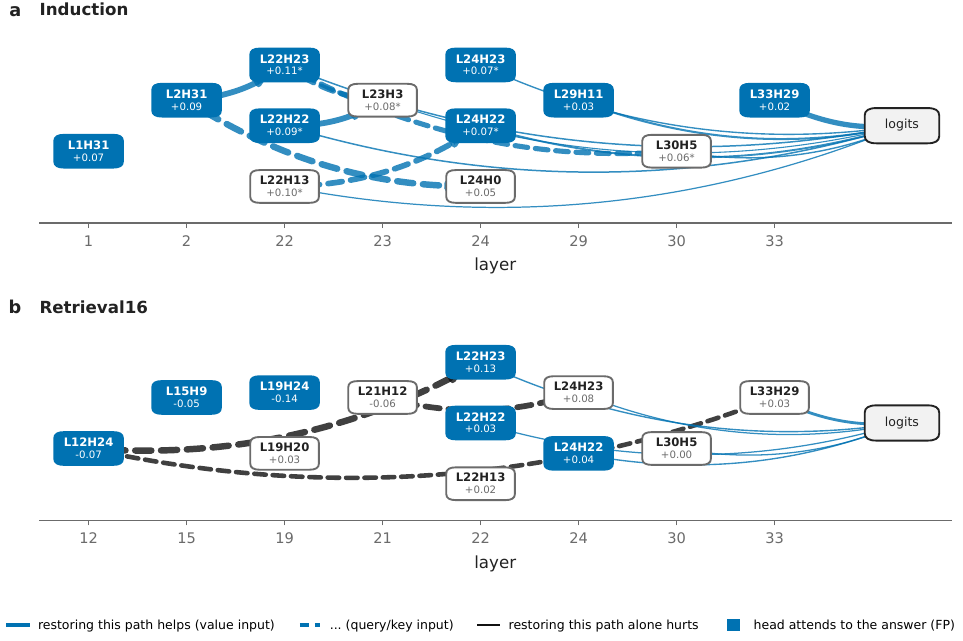}
  \caption{Circuits recovered by path patching (Qwen3-4B, A4, full-width Hadamard). Nodes are the 12
  heads with the largest calibration attribution for each task, placed by layer; filled nodes put at
  least 0.3 of their full-precision attention on the answer position. The number in each node is the
  recovery from restoring that head alone (every reader of its output; * = 95\% interval excludes zero).
  An arrow from head $U$ to head $D$ restores $U$'s full-precision output only in $D$'s value input
  (solid) or query and key inputs (dashed); an arrow to the logits restores it only at the final norm.
  Only edges whose 95\% interval excludes zero are drawn; width is proportional to the recovery.
  (a) Induction has a sparse circuit whose heads matter one at a time: an early head (L2H31) feeds the
  queries and keys of heads in layers 22--24, which write to the logits. (b) On retrieval no single head and no
  head-to-head edge helps on its own, and several edges hurt; only direct paths to the logits are
  positive. The 30 heads restored together recover 0.71 of the gap (Figure~\ref{fig:circuit-xmodel}). Llama-3.2-3B and Qwen3-1.7B show the same contrast (Figure~\ref{fig:circuit-diagram-xmodel}, Appendix~\ref{app:circuit}).}
  \label{fig:circuit-diagram}
\end{figure}

\begin{figure}[htbp]
  \centering
  \includegraphics[width=\linewidth]{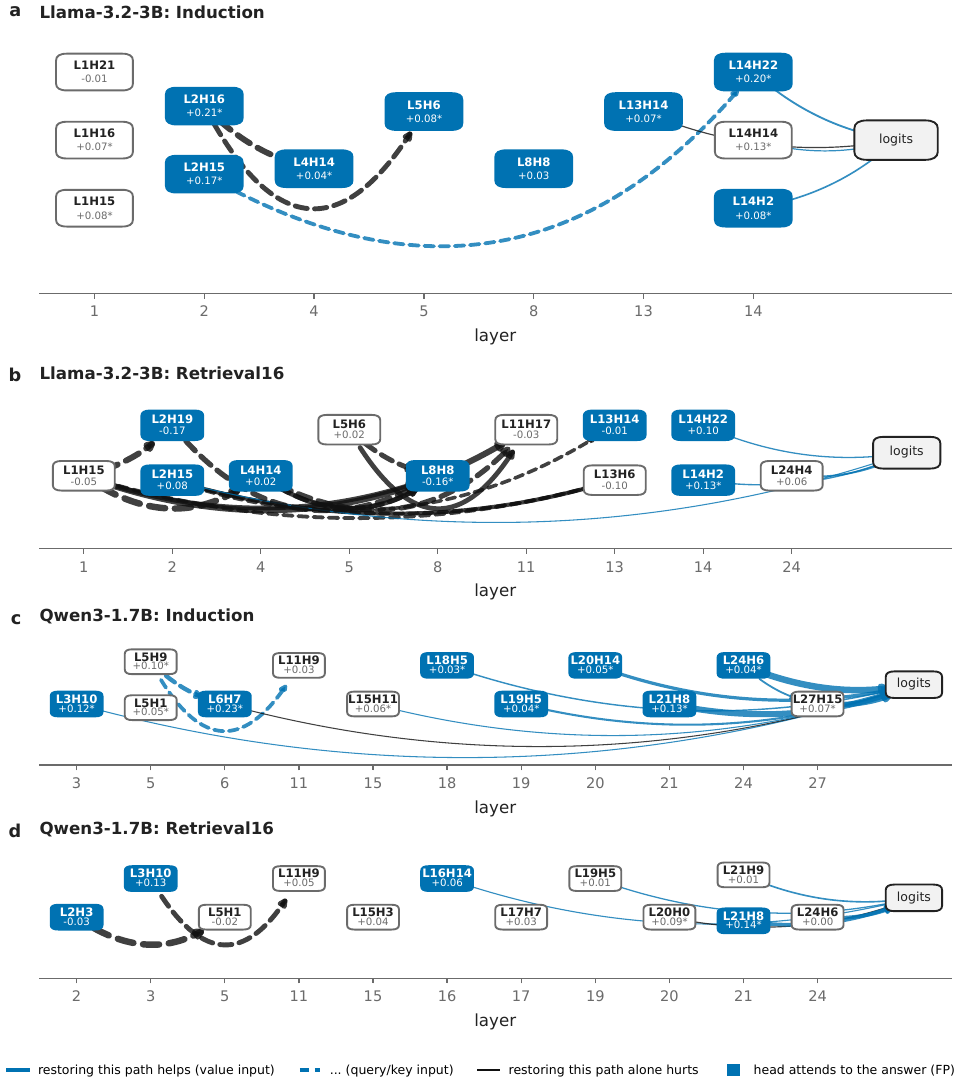}
  \caption{Path patching on Llama-3.2-3B and Qwen3-1.7B (A4, full Hadamard, $\alpha = 1$), drawn as in
  Figure~\ref{fig:circuit-diagram}: the 12 top heads per task by layer, the number in each node the
  recovery from restoring that head alone (* = 95\% interval excludes zero), and only edges whose
  interval excludes zero (solid: value input; dashed: query and key input; blue helps, black hurts).
  Edges spanning more than one layer are bent below the nodes and pass through none of them.
  (a, c) Induction: 10 and 11 of the 12 heads matter on their own; of the few positive head-to-head edges,
  Llama-3.2-3B's one (L2H15$\to$L14H22) and two of Qwen3-1.7B's three (L5H9$\to$L6H7, L5H9$\to$L11H9) feed queries
  and keys, the third the values of L27H15, but their number is close to what chance gives. (b, d) Retrieval16: no head-to-head edge helps, several hurt (14 on
  Llama-3.2-3B), only 1 and 2 heads matter alone, and the positive paths are direct writes to the logits.
  Restored together, the top 30 heads recover 0.74 and 0.59 of the gap (Figure~\ref{fig:circuit-xmodel}).}
  \label{fig:circuit-diagram-xmodel}
\end{figure}

\begin{figure}[htbp]
  \centering
  \includegraphics[width=\linewidth]{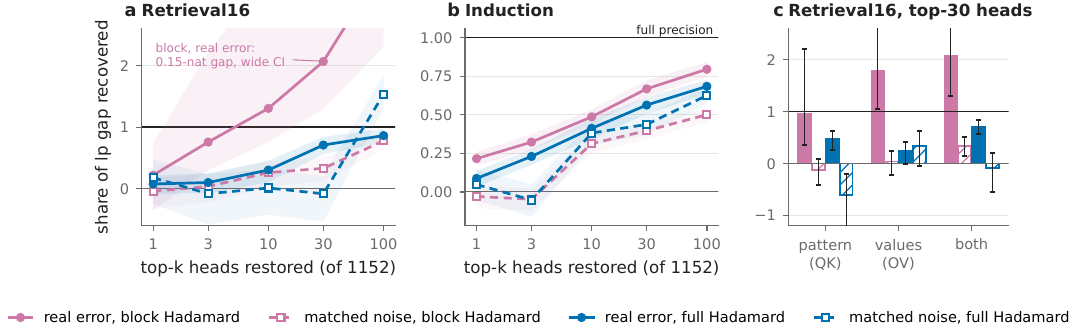}
  \caption{Restoring the top attention heads to full precision. (a, b) Share of the target log-prob
  gap recovered by restoring the top-$k$ heads (ranked on calibration items), for the real A4 error
  (solid) and matched Gaussian noise (dashed), under block (pink) and full-width (blue) Hadamard
  rotation; bands are 95\% bootstrap intervals. On retrieval the real error's damage is concentrated
  in the top heads and the noise's is not; on induction the two behave alike. The block rotation's
  retrieval gap is small (0.15 nats), so its ratio is wide. (c) Retrieval16, top-30 heads: restoring
  only the attention pattern (QK), only the values (OV), or both.}
  \label{fig:circuit-patching}
\end{figure}

\begin{table}[t]
\centering
\footnotesize
\setlength{\tabcolsep}{3.5pt}
\caption{Keeping circuit heads at 8 bits. Base rows give the score; the rows below each give the paired difference from it, bold where the 95\% interval excludes zero. ``Circuit heads'' are the heads with the largest attribution on calibration retrieval16 items (Section~\ref{sec:circuit}); their query, key and value outputs and their output-projection inputs use 8-bit activations. ``Random heads'' is the same budget on random heads; ``their layers'' puts every attention input of the layers containing the top 30 at 8 bits. A8 is the share of all linear-layer multiply--accumulates with an 8-bit activation. Probes: 2000 items; QA as in Table~\ref{tab:qa}; perplexity on the full WikiText-2 test set.}
\label{tab:headmix}
\begin{tabular}{l l r r rr rrr}
\toprule
Model & Setting & A8 (\%) & Perplexity & R16 & R64 & TriviaQA & NQ & LongBench \\
\midrule
Llama-3.2-3B & $\alpha=1$ & -- & 8.84 & 0.419 & 0.127 & 0.446 & 0.165 & 0.254 \\
 & \quad + 10 circuit heads & 0.6 & 8.83 & \textbf{+0.019} & +0.012 & -- & -- & -- \\
 & \quad + 30 circuit heads & 1.6 & 8.81 & \textbf{+0.024} & \textbf{+0.020} & -- & -- & -- \\
 & \quad + 30 random heads & 1.7 & 8.81 & +0.008 & +0.001 & -- & -- & -- \\
 & \quad + their layers & 13.4 & 8.65 & \textbf{+0.036} & \textbf{+0.027} & -- & -- & -- \\
\addlinespace
 & tuned ($\alpha=0.70$) & -- & 8.36 & 0.443 & 0.137 & 0.503 & 0.182 & 0.266 \\
 & \quad + 10 circuit heads & 0.6 & 8.35 & +0.008 & \textbf{+0.015} & -- & -- & -- \\
 & \quad + 30 circuit heads & 1.6 & 8.33 & +0.003 & \textbf{+0.029} & -0.008 & +0.003 & +0.004 \\
 & \quad + 30 random heads & 1.7 & 8.34 & +0.005 & +0.007 & -0.008 & -0.004 & +0.001 \\
 & \quad + their layers & 13.4 & 8.25 & \textbf{+0.028} & \textbf{+0.038} & +0.008 & +0.013 & -0.005 \\
\addlinespace
 & W4 + tuned ($\alpha=0.70$) & -- & 8.74 & 0.382 & 0.092 & 0.429 & 0.149 & 0.258 \\
 & \quad + 30 circuit heads & 1.6 & 8.71 & \textbf{+0.036} & \textbf{+0.019} & +0.006 & +0.007 & +0.000 \\
 & \quad + 30 random heads & 1.7 & 8.73 & \textbf{+0.017} & +0.005 & -- & -- & -- \\
 & \quad + their layers & 13.4 & 8.61 & \textbf{+0.041} & \textbf{+0.029} & -- & -- & -- \\
\midrule
Qwen3-1.7B & $\alpha=1$ & -- & 11.99 & 0.283 & 0.070 & 0.248 & 0.088 & -- \\
 & \quad + 10 circuit heads & 0.7 & 11.92 & \textbf{+0.054} & \textbf{+0.014} & -- & -- & -- \\
 & \quad + 30 circuit heads & 2.2 & 11.89 & \textbf{+0.040} & \textbf{+0.014} & -- & -- & -- \\
 & \quad + 30 random heads & 2.2 & 11.90 & +0.002 & +0.000 & -- & -- & -- \\
 & \quad + their layers & 17.0 & 11.57 & \textbf{+0.062} & \textbf{+0.021} & -- & -- & -- \\
\addlinespace
 & tuned ($\alpha=0.70$) & -- & 10.86 & 0.378 & 0.106 & 0.296 & 0.113 & 0.264 \\
 & \quad + 10 circuit heads & 0.7 & 10.82 & +0.007 & \textbf{+0.020} & -- & -- & -- \\
 & \quad + 30 circuit heads & 2.2 & 10.81 & +0.017 & +0.011 & -0.003 & +0.003 & +0.007 \\
 & \quad + 30 random heads & 2.2 & 10.82 & \textbf{-0.020} & -0.003 & +0.005 & +0.006 & -0.007 \\
 & \quad + their layers & 17.0 & 10.57 & \textbf{-0.025} & \textbf{-0.020} & +0.010 & +0.003 & +0.010 \\
\addlinespace
 & W4 + tuned ($\alpha=0.70$) & -- & 11.25 & 0.414 & 0.138 & 0.261 & 0.088 & 0.253 \\
 & \quad + 30 circuit heads & 2.2 & 11.20 & \textbf{+0.039} & +0.008 & +0.015 & -0.002 & -0.004 \\
 & \quad + 30 random heads & 2.2 & 11.20 & \textbf{-0.021} & \textbf{-0.016} & -- & -- & -- \\
 & \quad + their layers & 17.0 & 10.94 & -0.017 & \textbf{-0.020} & -- & -- & -- \\
\midrule
Qwen3-4B & full Had., $\alpha=1$ & -- & 9.15 & 0.426 & 0.146 & -- & -- & -- \\
 & \quad + 10 circuit heads & 0.3 & 9.13 & \textbf{+0.020} & \textbf{+0.017} & -- & -- & -- \\
 & \quad + 30 circuit heads & 1.0 & 9.13 & +0.018 & \textbf{+0.019} & -- & -- & -- \\
 & \quad + 30 random heads & 1.1 & 9.11 & -0.002 & +0.005 & -- & -- & -- \\
 & \quad + their layers & 10.8 & 8.99 & \textbf{+0.027} & \textbf{+0.032} & -- & -- & -- \\
\addlinespace
 & full Had., $\alpha=0.70$ & -- & 8.68 & 0.505 & 0.193 & 0.411 & 0.173 & 0.293 \\
 & \quad + 10 circuit heads & 0.3 & 8.67 & +0.009 & +0.006 & -- & -- & -- \\
 & \quad + 30 circuit heads & 1.0 & 8.66 & +0.012 & +0.002 & +0.005 & -0.009 & +0.011 \\
 & \quad + 30 random heads & 1.1 & 8.67 & -0.014 & -0.005 & +0.001 & -0.007 & +0.004 \\
 & \quad + their layers & 10.8 & 8.58 & -0.008 & -0.005 & +0.011 & -0.001 & +0.021 \\
\addlinespace
 & W4 + full Had., $\alpha=0.70$ & -- & 8.90 & 0.465 & 0.175 & 0.385 & 0.154 & 0.276 \\
 & \quad + 30 circuit heads & 1.0 & 8.88 & \textbf{+0.020} & +0.007 & -0.005 & -0.006 & -0.002 \\
 & \quad + 30 random heads & 1.1 & 8.90 & +0.006 & -0.002 & -- & -- & -- \\
 & \quad + their layers & 10.8 & 8.83 & +0.006 & +0.004 & -- & -- & -- \\
\bottomrule
\end{tabular}
\end{table}

\paragraph{Extended steering grid.}
Table~\ref{tab:circuit-steer} adds a grid of stronger interventions. Sharpening is flat between
$\times 3$ and $\times 8$ (calibration log-probability within 0.1 nats), so the main grid's optimum was not
an edge artefact. Output gains of $\times 4$ and above break the model: retrieval falls to about 0.01 and
perplexity rises to between 48 and several thousand. In every setting the same steering raises the
full-precision model to 0.61, so the A4 model stays 0.06--0.14 below it after steering; this holds
for matched noise as for the real error.

\section{The damage does not run through attention margins}
\label{sec:margin}

A natural account of retrieval failure is that quantisation noise in the attention logits of the
retrieval heads exceeds the margin between the correct key and the strongest distractor. We captured
per-head attention logits for the 64 heads with full-precision attention mass at least 0.1 on the
target, on Qwen3-4B, under six quantisation settings (rotated RTN at two clipping ratios, stochastic
rounding, matched Gaussian, sign-randomised, and A6 without rotation), restricted to items the
full-precision model gets right. We set a kill criterion in advance: held-out (leave-one-setting-out)
McFadden pseudo-$R^2$ below 0.2.

The model fails it. The pre-registered margin-to-noise predictor reaches 0.010 on retrieval16 and 0.033
on induction; variants using the top head or the minimum over four heads do no better, and the
parameter-free prediction $\Phi(-m/\sigma)$ predicts a 1\% failure rate where 37\% is observed. The
margin at the output logits reaches 0.147 and 0.403, so failure is predictable, just not from attention
margins. At the level of settings, matched noise perturbs the selected heads' margins less than real
RTN does (median $\sigma$ 0.74 against 0.82) yet breaks more items (37\% against 25\%), consistent with
Section~\ref{sec:partial}: the extra damage sits at an MLP input, not at the query--key product. On
induction, per-head argmax flips match the Gaussian prediction closely (0.051 against 0.052), so the
noise model of the logits is fine; it is the link from head-level flips to item failure that does not
hold. This is one model and one seed, and per-head margins ignore how heads combine.

\section{Part I in detail: sensitivity, features and repair}
\label{app:part1}

This appendix holds the parts of the single-block study (Part~I) that Section~\ref{sec:mask} summarises.

\subsection{Calibration of perplexity against probe accuracy}
\label{sec:calibration}

Perplexity ranks methods the way the probes do. Over \Pairs{} single-block within-model comparisons
between the three \emph{deployable arms} (Uniform, Protected-Fisher, and Rotated, as opposed to
controls that only exist to isolate a mechanism), the arm perplexity prefers keeps less induction in
\ReversalInd{} percent of cases and less retrieval in \ReversalRet{} percent. It does not say how much
is lost. Table~\ref{tab:dose} reads the same comparison as a dose response: where perplexity has
risen by a factor of 1.2 to 1.5 (\MildBandCount{} settings), induction keeps \MildBandInduction{} of
its intact accuracy, while retrieval has fallen to \MildBandRetrieval{}, from \TightBandRetrieval{}
one band below. Resampling models rather than settings gives induction in
[\BootIndLow{}, \BootIndHigh{}] and retrieval in [\BootRetLow{}, \BootRetHigh{}]; the intervals do not
overlap, and all families agree in direction (Table~\ref{tab:dosemodel}).

\begin{table}[t]
  \centering
  \small
  \caption{Part~I. Probe accuracy by perplexity band, relative to the unquantised model. Pooled over
  the three deployable arms, all models, depths, and bitwidths, single-block and end-to-end
  settings together.}
  \label{tab:dose}
  \begin{tabular}{lrrr}
    \toprule
    Perplexity, relative to intact & Settings & Induction & Retrieval \\
    \midrule
    \TableDoseBody
    \bottomrule
  \end{tabular}
\end{table}

\subsection{Granularity decides whether the error is noise-like}
\label{sec:premium}

If the sign pattern comes from which coordinates share a scale, how widely the quantiser shares it
should decide how organised the error is. Table~\ref{tab:premium} (Appendix~\ref{app:regime}) reports
the \emph{structure premium}, quantised perplexity divided by the perplexity of Gaussian noise of
identical per-channel mean squared error. At fixed bitwidth it differs by orders of magnitude with
granularity: per-token scaling runs from \PremTokenEight{} at eight bits to \PremTokenTwo{} at two,
while per-tensor scaling is already at \PremTensorEight{} at eight bits (over \NumPairs{} matched
pairs the two arms agree in measured error to a median of \ControlMedian{} percent). The per-tensor
result is the extreme case of the effect \citet{dettmers2022llmint8} identified, a single outlier
channel setting the scale. Part of it reflects the massive first-position activation, which the
tensor-wide maximum does not exclude.

\subsection{Sensitivity is not magnitude}
\label{sec:sensitivity}

The rank correlation between per-coordinate sensitivity $F_j$ (Equation~\ref{eq:fisher-loss}) and
mean absolute activation has median \SpearmanMedian{} (range \SpearmanMin{} to \SpearmanMax{}), and the
64 most sensitive and 64 largest coordinates share a median of only \OverlapMedian{} of their members
(Table~\ref{tab:sensitivity}), agreeing with \citet{heo2024rethinking}. Protecting channels chosen by
magnitude is not the same as protecting the channels the loss depends on.

\subsection{Looking Directly at the Activation}
\label{sec:artifact-analysis}

Since quantisation and matched noise act on the same residual-stream activation, if they differ in
effect they should differ in how they change it. We measure this through sparse dictionaries: for
Pythia we train sparse autoencoders at the intervention depth
\citep{bricken2023monosemanticity,cunningham2024sae,gao2024scaling}; for the other families we use
released dictionaries (GemmaScope \citep{lieberum2024gemmascope}, Qwen-Scope
\citep{qwen2026qwenscope}, and the GPT-2 dictionaries of \citet{gao2024scaling}). Their widths,
sparsities and corpora differ (Appendix~\ref{app:features}), so we argue feature-level claims within
a family only.

\subsubsection{Global similarity does not track the failure}
\label{sec:cka}
\label{sec:concentration}

\begin{figure}[htbp]
  \centering
  \includegraphics[width=0.48\linewidth]{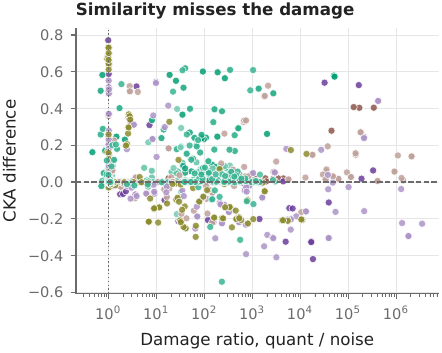}
  \hfill
  \includegraphics[width=0.48\linewidth]{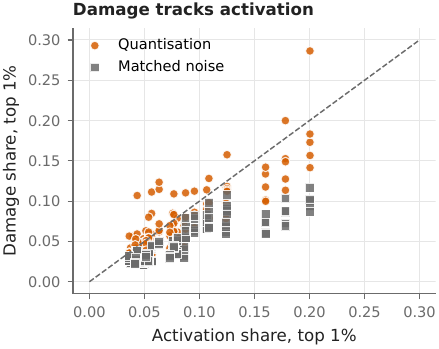}
  \caption{Left: difference in CKA score between the two arms against how much more damage
  quantisation does. Points above the dashed line are settings where the more damaging arm
  scores as the more similar one (Section~\ref{sec:cka}). Right: share of damage falling on
  the most active one percent of features, against the share of activation those features
  carry. Quantisation sits on the diagonal, matched noise below it.}
  \label{fig:mechanism}
\end{figure}

Across the \CkaTotal{} quantised/matched-noise pairs, \CkaInversionPct{} percent are settings where the
arm doing \emph{more} damage gets the \emph{higher} centred-kernel-alignment (CKA) score, no better
than a coin flip. Standard diagnostics misorder settings by distance from baseline \MisorderMse{}
percent of the time for mean squared error, \MisorderSds{} percent for subspace distortion, and
\MisorderCka{} percent for CKA (Figure~\ref{fig:mechanism}). At the most damaged setting, quantisation
raises perplexity to \WorstBlowup{} times baseline while still scoring \WorstCka{} on CKA
\citep[cf.][]{ding2021grounding}. Damage tracks activation mass: quantisation puts about 5.6 percent
of its damage in the top one percent of features, close to their \ActivationShare{} percent share of
activation, while matched noise spreads onto low-activation features (Table~\ref{tab:concentration}).

\subsubsection{The error is directional: the deadzone}
\label{sec:shunt}
\label{sec:causal}

Matching the size of the error does not match its sign. The summed shift in feature activation over
its full-precision scale is \ShuntQuantFour{} at four bits, $-0.0397$ at three and \ShuntQuantTwo{}
at two under quantisation, against \ShuntNoiseFour{}, $+0.0242$ and \ShuntNoiseTwo{} under matched
noise (Table~\ref{tab:shunt}). The reason is the deadzone: values below half a step round to zero,
biasing features downward, as \citet{li2026mxfp4} find for MXFP4; symmetric noise has no deadzone and
is rectified upward by a dictionary whose features are non-negative. A second-moment match leaves
this sign structure free to differ. Zeroing the most damaged features is worse than leaving them,
and restoring them to full precision recovers most of the perplexity loss, more than restoring
random or least-damaged features (Table~\ref{tab:causal}). This restoration test is perplexity-only
and an oracle, and the most damaged features are largely the most active ones, so it localises the
damage rather than proving these features carry the probes' computation.
Part~II returns to the deadzone with an exact decomposition (Section~\ref{sec:norot}).

\subsection{Representation-Aligned Repair}
\label{sec:restore}

We test two interventions at matched average bitwidth (Table~\ref{tab:pareto}). The first protects
the most loss-sensitive directions of Section~\ref{sec:sensitivity} at high precision and drops the
rest to a matched average (\emph{Protected, Fisher}). It improves perplexity over uniform
quantisation only at three bits and below (324.3 against 1{,}310.6) and is worse at four bits and up
(89.7 against 55.1); induction stays at 0.007 at three average bits, barely above uniform's 0.001.
The directions it preserves are the ones perplexity depends on, not the ones the probes need
(Appendix~\ref{sec:allocation-full}). The second selects no subspace: it quantises in a randomly
rotated frame (\emph{Rotated}), spreading the activation's energy across coordinates. At three
average bits it reaches perplexity 33.1 with induction 0.980. The same rotation also lowers the
per-coordinate maximum that sets the step, so Part~I cannot say whether rotation helps by removing
structure or by shrinking the error; Section~\ref{sec:rot} answers this.

\begin{table}[t]
  \centering
  \small
  \caption{Part~I. Perplexity and probe accuracy at matched average bitwidth, median over model
  families at three relative depths. Full-precision perplexity is 21.6.}
  \label{tab:pareto}
  \begin{tabular}{lrrrr}
    \toprule
    Precision & Average bits & Perplexity & Induction & Retrieval \\
    \midrule
    \TableParetoBody
    \bottomrule
  \end{tabular}
\end{table}

\paragraph{End to end.}
\label{sec:alllayer}
Quantising every block from the intervened one onward (Table~\ref{tab:alllayerfull}) shifts but does
not change the direction: at 4 average bits Uniform collapses both probes (induction 0.003, retrieval
0.000) while Rotated reaches induction 0.968 but retrieval only 0.534. In this variant matched noise
of the uniform quantiser's size also collapses induction. Both arms are at the floor there (the
uniform arm's perplexity is in the thousands), so the comparison cannot separate structure from
size; Part~II revisits the question at error levels where the model still works.

\paragraph{Hadamard rotation and latency.}
\label{sec:quarot}
Replacing the dense random-orthogonal matrix by QuaRot's Hadamard construction matches it everywhere
and exceeds it in the harder regimes (induction \DenseSixBitPerTensor{} to \QuarotSixBitPerTensor{} at
per-tensor 6 bits; \DenseTwoBitPerToken{} to \QuarotTwoBitPerToken{} at per-token 2 bits;
Appendix~\ref{app:quarot}). Applied as a dense matmul, the rotation adds \RotationLatencyLow{} to
\RotationLatencyHigh{} percent forward-pass latency per block, an upper bound on a fused kernel's cost.

\section{Deployed quantisers: full results}
\label{app:realquant}

This expands the comparison in Section~\ref{sec:realquant} to every arm, bitwidth, and
configuration we measured: three deployed quantisers (GPTQ, AWQ, SmoothQuant), each at its
standard weight-only setting (W3A16, W4A16, and W8A8 for SmoothQuant, which supports 8-bit
activations natively), and, where the implementation supports it, with activations also
quantised to 4 bits (W4A4). We run these on pythia-1.4b and Qwen3-8B-Base at the same three
relative depths as every other table in this paper, using each quantiser's reference
implementation (GPTQ via AutoGPTQ, AWQ via the original llm-awq repository, SmoothQuant via
its official release) rather than our own reimplementation.

\begin{table}[h]
  \centering
\small
  \caption{Full deployed-quantiser sweep. Target separates arms that perturb weights from our
own arms that perturb the residual stream; comparisons across that boundary are not
matched-budget (\S\ref{sec:limitations}). GPTQ's W4A4 perplexity, two orders of magnitude
worse than AWQ's, follows from its weight-reconstruction calibration objective and is not a
claim about its intended weight-only use. pythia-1.4b's SmoothQuant and AWQ W4A16 rows are
independently measured and coincide almost exactly (verified against the per-layer raw output);
at this mild, weight-only setting the two methods barely perturb this model differently. The
Uniform (activation) and Rotated (activation) rows use the end-to-end variant of
Section~\ref{sec:alllayer} (every block from the intervened one onward), not the single-block
setting the rest of the paper reports by default, since GPTQ/AWQ/SmoothQuant also quantise the
whole model; this is why they read far worse than the single-block figures elsewhere.}
  \label{tab:realquantfull}
  \begin{tabular}{llrlrrr}
    \toprule
    Arm & Target & Bits & Config & Perplexity & Induction & Retrieval \\
    \midrule
    \TableRealQuantAppendixBody
    \bottomrule
  \end{tabular}
\end{table}

\begin{figure}[h]
  \centering
  \includegraphics[width=0.85\linewidth]{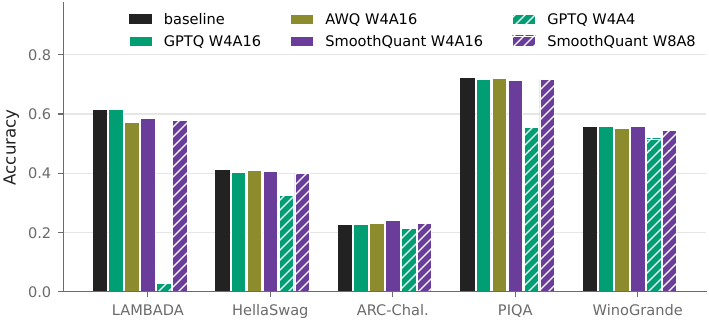}
  \caption{Standard benchmarks under each deployed-quantiser arm, pythia-1.4b (referenced from
Section~\ref{sec:realquant}). GPTQ's W4A4 collapse on LAMBADA (a next-token-prediction task
that needs the passage's context, like this paper's retrieval probe) matches the mechanistic
damage; the four multiple-choice commonsense benchmarks move little under any arm, GPTQ W4A4
included.}
  \label{fig:benchmark-battery}
\end{figure}

\section{Hadamard rotation: full comparison and latency}
\label{app:quarot}

\paragraph{Construction.} We copy QuaRot's Hadamard construction verbatim (Apache-2.0): a
random $\pm 1$ sign diagonal pushed through a fast Walsh-Hadamard transform, with a
precomputed special-order Hadamard factor for hidden sizes that are not a power of two
\citep{ashkboos2024quarot}. It replaces the dense random-orthogonal matrix used elsewhere in
this paper, as a \texttt{hadamard\_per\_token} or \texttt{hadamard\_per\_tensor} scaling
mode. We run it on the same four models, three depths, and five bitwidths as
Table~\ref{tab:induction}.

\begin{table}[h]
  \centering
\small
  \caption{The dense-versus-Hadamard comparison of Section~\ref{sec:quarot}, both
granularities in full.}
  \label{tab:quarotfull}
  \begin{tabular}{lrrrr}
    \toprule
    & \multicolumn{2}{c}{Per token} & \multicolumn{2}{c}{Per tensor} \\
    \cmidrule(lr){2-3}\cmidrule(lr){4-5}
    Bits & Dense & Hadamard & Dense & Hadamard \\
    \midrule
    \TableQuarotMainBody
    \bottomrule
  \end{tabular}
\end{table}

\paragraph{Latency.} Forward-pass-only, batch 16, sequence length 128 (matching the perplexity
evaluation), CUDA-event timing on one mid-network block per model, using the same hook every
other measurement in this paper runs through.

\begin{table}[h]
  \centering
\small
  \caption{Measured latency: baseline, the rotation alone, and the rotation with 4-bit
quantisation (Section~\ref{sec:quarot}). The quantisation step adds little next to the
rotation matmul on every model. We time the rotation as a dense $d \times d$ matmul, the way
every rotation arm in this paper applies it, not as QuaRot's fused fast-Hadamard-transform
kernel, which a production deployment would use and which would cost less.}
  \label{tab:quarotlatency}
  \begin{tabular}{lrrr}
    \toprule
    Model & Baseline (ms) & +rotation (ms) & +rotation+quantise, 4-bit (ms) \\
    \midrule
    \TableQuarotLatencyBody
    \bottomrule
  \end{tabular}
\end{table}

\begin{figure}[h]
  \centering
  \includegraphics[width=0.48\linewidth]{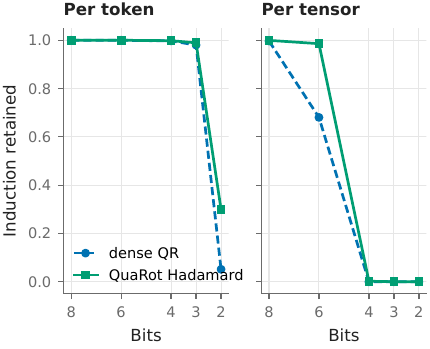}
  \hfill
  \includegraphics[width=0.48\linewidth]{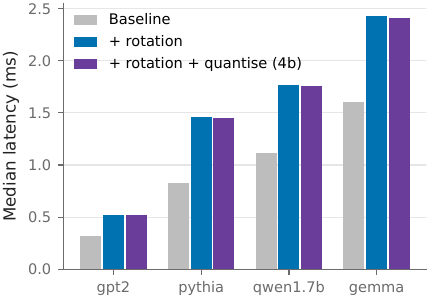}
  \caption{Left: Table~\ref{tab:quarotfull}, QuaRot's Hadamard construction against the diagnostic dense rotation, median induction over four models at three depths, with one scale per token and per tensor. Right:
  Table~\ref{tab:quarotlatency}, measured forward-pass latency added by the rotation, with and
  without the 4-bit quantisation step, on the same four models.}
  \label{fig:quarot-latency}
\end{figure}

\section{Scale and instruction tuning: full results}
\label{app:scale}

We use the same single-block pipeline as the main four-model sweep (three relative depths,
bitwidths 8/6/4/3/2, uniform, matched Gaussian noise, and rotated arms under per-token
scaling), run on Qwen2.5-32B and Qwen3-8B (the instruction-tuned release; Qwen3's naming
convention is the reverse of what the pattern elsewhere in this paper would suggest: the bare
name is the chat model, and \texttt{-Base} is the pretrain-only checkpoint). Baselines:
Qwen2.5-32B scores 0.974 induction, 0.539 retrieval, perplexity 15.2 intact; Qwen3-8B scores
0.970 induction, 0.379 retrieval, perplexity 25.6 intact. Qwen3-8B's intact perplexity is
higher than that of its base counterpart, as expected for a chat-tuned model scored on raw
next-token prediction over web text rather than on its own training distribution.

\begin{table}[h]
  \centering
\small
  \caption{Qwen3-8B (instruct), full bitwidth sweep, matching the schema of
Table~\ref{tab:induction}. Retained fraction is accuracy on $\hat h$ divided by accuracy on
$h$; values fractionally above 1 (e.g.\ the 1.005 below) reflect sampling noise at 256
positions rather than the perturbation improving on the intact model, and recur at several
other points in this paper's tables.}
  \label{tab:qwen3instfull}
  \begin{tabular}{lrrrrr}
    \toprule
    Perturbation & 8 bits & 6 bits & 4 bits & 3 bits & 2 bits \\
    \midrule
    Uniform & 0.999 & 0.997 & 0.940 & 0.007 & 0.000 \\
    Gaussian noise, matched & 1.000 & 0.999 & 0.981 & 0.878 & 0.671 \\
    Rotated & 1.000 & 1.000 & 1.005 & 0.993 & 0.251 \\
    \bottomrule
  \end{tabular}
\end{table}

\begin{table}[h]
  \centering
\small
  \caption{Qwen2.5-32B, full bitwidth sweep, matching the schema of Table~\ref{tab:induction}.
The collapse the four-model set shows at three to four bits does not appear here until two
bits, where quantisation (0.086) again trails both controls, but the ordering is not identical
to the core sweep: rotation (0.219) recovers only partially and, unlike at every other setting
in this paper, trails matched Gaussian noise (0.868) rather than approaching full recovery.}
  \label{tab:qwen25full}
  \begin{tabular}{lrrrrr}
    \toprule
    Perturbation & 8 bits & 6 bits & 4 bits & 3 bits & 2 bits \\
    \midrule
    Uniform & 1.000 & 0.999 & 1.000 & 0.983 & 0.086 \\
    Gaussian noise, matched & 1.000 & 1.000 & 0.998 & 0.993 & 0.868 \\
    Rotated & 1.000 & 1.000 & 1.001 & 0.996 & 0.219 \\
    \bottomrule
  \end{tabular}
\end{table}

\begin{figure}[h]
  \centering
  \includegraphics[width=0.95\linewidth]{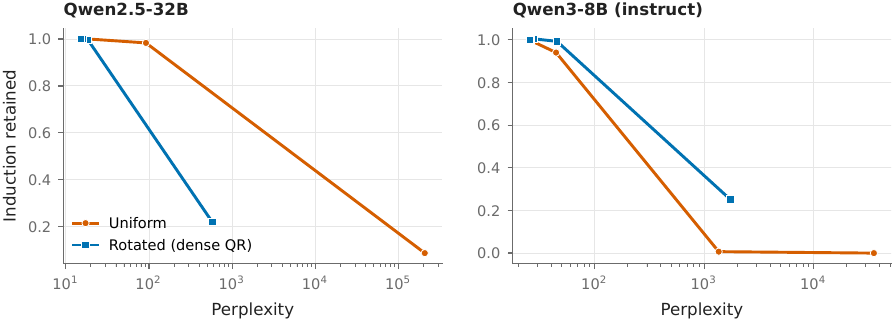}
  \caption{Tables~\ref{tab:qwen3instfull} and \ref{tab:qwen25full} plotted on the perplexity
axis instead of the bitwidth axis, all five bitwidths for both arms and both models.
At every bitwidth the dense rotation has lower perplexity and at least as much induction as uniform quantisation, and on Qwen3-8B (instruct) it retains more induction at every perplexity. On Qwen2.5-32B the curves cross: uniform 3-bit keeps 0.98 of induction at perplexity 92, while rotated 2-bit keeps 0.22 at perplexity 577.}
  \label{fig:new-models}
\end{figure}

\section{Bootstrap confidence intervals}
\label{app:bootstrap}

Section~\ref{sec:calibration} reports bootstrap intervals for the dose-response claim
(Table~\ref{tab:dose}). We resample models rather than settings, so the interval reflects
variation between models rather than the correlation between settings within one model. We
extend the same procedure (4{,}000 resamples, cluster bootstrap over models) to six further
tables that state point estimates without uncertainty in the main text:
Tables~\ref{tab:probes}, \ref{tab:ladder}, \ref{tab:induction}, \ref{tab:quarotfull} at 3
bits, and the allocation and restoration tables of Appendix~\ref{app:regime}.

\begin{figure}[h]
  \centering
  \includegraphics[width=0.98\linewidth]{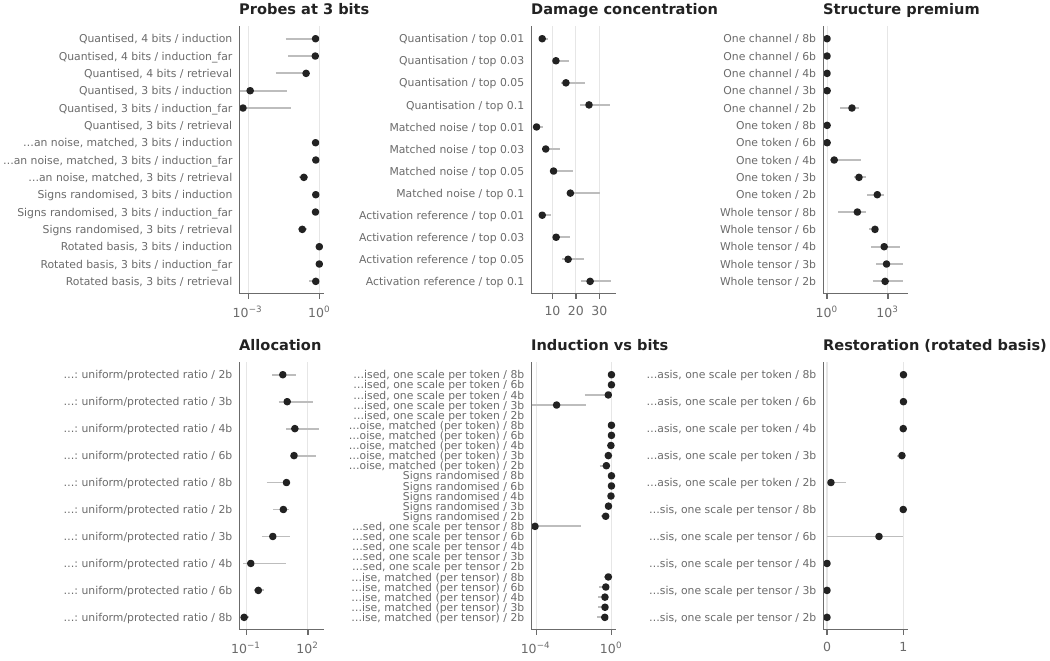}
  \caption{Point estimate and 95\% CI, cluster bootstrap over models, for every row of the six
tables above. Most intervals are tight relative to the effect sizes the main text reports;
the widest interval in the set is at the transition this paper's central claim depends on:
quantised induction at 3 bits (top left panel), the regime in which a point estimate is
least trustworthy and the one every number in Section~\ref{sec:induction} comes from. The
ordering between arms holds in every resample even where the absolute level does not.}
  \label{fig:bootstrap-cis}
\end{figure}

\section{Sign-pattern substitute ladder: full results}
\label{app:ladder-table}

Table~\ref{tab:ladder} gives the full substitute ladder summarised in
Section~\ref{sec:ladder}: the share of quantisation's excess perplexity reproduced by each
substitute for the quantisation error, at every bitwidth we measure. The second row holds
every error magnitude exactly and randomises only the signs; the last row is not a substitute
for the error but the same quantiser applied in a rotated basis (Section~\ref{sec:restore}).

\begin{table}[h]
  \centering
  \small
  \caption{Share of quantisation's excess perplexity reproduced by each substitute, under
  per-token scaling. Median over four families at three relative depths.}
  \label{tab:ladder}
  \begin{tabular}{lrrrrr}
    \toprule
    Substitute for the error & 8 bits & 6 bits & 4 bits & 3 bits & 2 bits \\
    \midrule
    \TableLadderBody
    \bottomrule
  \end{tabular}
\end{table}

\section{Extended derivation: the diagonal noise account}
\label{app:theory-full}

Modelling quantisation error as additive noise is an old idea, and so are its failure
conditions. \citet{bennett1948spectra} set out why the model breaks: the error is a
deterministic function of the signal, not an independent draw, and the approximation holds
only when the step is small relative to the signal's variation. Low-bit activation
quantisation sits exactly in that regime. What nobody has measured is how much that matters
in a transformer's residual stream.

Write $h\in\mathbb{R}^{d}$ for the residual stream at the intervened layer, and
$\varepsilon = Q(h) - h$ for the error the quantiser adds to it. Expanding the loss to second
order and keeping only the diagonal gives Equation~\ref{eq:fisher-loss} in
Section~\ref{sec:theory}, where the sum runs over residual coordinates, the ones the
quantiser and our control both act on. Under this account, the induced loss depends on the
error only through its per-coordinate squares. So two perturbations matched in per-channel
second moment should cost the same in expectation, whatever their signs or their dependence
on $h$.

Equation~\ref{eq:fisher-loss} is a hypothesis, and it is the hypothesis our control tests. If
an arm carrying quantisation error and an arm carrying a matched perturbation diverge, that
shows this diagonal, magnitude-only account is inadequate. It does not show that magnitude
cannot matter. That is a weaker claim, for two reasons. First, the diagonal form discards
cross-coordinate terms, and correlated error is exactly where those omitted terms are
largest. Second, matching a second moment leaves higher moments, cross-token structure, and
the dependence of $\varepsilon$ on $h$ all free to differ. That is why Section~\ref{sec:noise}
does not rest on one control: Table~\ref{tab:ladder} replaces the Gaussian arm with
substitutes that hold progressively more of the error fixed, including one that preserves
$\lvert\varepsilon_j\rvert$ for every coordinate of every token and alters only the signs, so
a gap against it cannot be a magnitude artefact in the residual basis.

One consequence is worth stating in advance. Setting a coordinate to zero replaces its error
term with $\varepsilon_j = -h_j$, which is larger than the error it displaces. So deletion
should never help under Equation~\ref{eq:fisher-loss}; Section~\ref{sec:causal}'s causal
interventions test this directly.

\begin{figure}[htbp]
  \centering
  \includegraphics[width=0.55\linewidth]{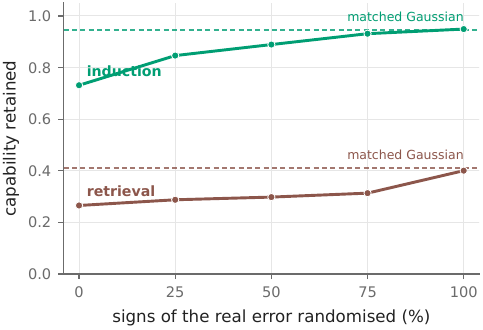}
  \caption{Capability retained as the fraction of resampled signs rises from 0 to 100 percent
  (solid lines, markers), against the matched-Gaussian control at the same bitwidths and models (dashed); medians over four families and five bitwidths. Both probes recover smoothly and monotonically, meeting the Gaussian
  reference almost exactly at 100 percent.}
  \label{fig:signdial}
\end{figure}

Section~\ref{sec:ladder} reads the two endpoints of this ladder (no signs resampled, and all
of them) off Table~\ref{tab:ladder}. Figure~\ref{fig:signdial} and Table~\ref{tab:signdial}
give the full dial between them, resampling the fraction of signs from 0 to 100 percent in
steps of 25. Both probes recover smoothly and monotonically as the fraction rises, converging
onto the matched-Gaussian control rather than jumping to it: this is the stronger form of the
claim that Equation~\ref{eq:fisher-loss}'s magnitude-only account is inadequate, a dial with a
shape rather than two mismatched arms.

\begin{table}[htbp]
  \centering
  \small
  \caption{Capability retained as the fraction of resampled signs rises from 0 (the original
  quantisation error) to 100 percent (full sign randomisation), against the matched-Gaussian
  reference at the same fixed model, layer, and bitwidth (constant across the dial by
  construction, since only the quantisation arm's sign fraction varies); this single-setting
  reference is not the pooled Gaussian-noise median of Table~\ref{tab:induction}, which
  aggregates over every family and bitwidth.}
  \label{tab:signdial}
  \begin{tabular}{lrrrr}
    \toprule
    Signs randomised & Induction & Induction (Gaussian) & Retrieval & Retrieval (Gaussian) \\
    \midrule
    \TableSignDialBody
    \bottomrule
  \end{tabular}
\end{table}

\section{Feature-dictionary apparatus}
\label{app:features}

Section~\ref{sec:artifact-analysis} analyses the activation artifact through a sparse
dictionary rather than in the raw residual basis, since the residual stream is not axis
aligned and represents more features than it has dimensions \citep{elhage2022toy}. This
appendix gives the apparatus that section relies on.

For Pythia we train top-$k$ sparse autoencoders at the intervention depth
\citep{bricken2023monosemanticity,cunningham2024sae,gao2024scaling}, with dictionary width
eight times the residual width and sparsity fixed at 64 active features, so the comparison
across scale in Table~\ref{tab:dictionaries} is not also a comparison across dictionary
capacity. Reconstruction error grows with model size at fixed expansion. For the other three
families we use released dictionaries: GemmaScope \citep{lieberum2024gemmascope}, Qwen-Scope
\citep{qwen2026qwenscope}, and the GPT-2 dictionaries of \citet{gao2024scaling} as
redistributed in SAELens. Using dictionaries we did not fit ourselves removes the concern
that we chose the basis to suit the result. Their widths, sparsities, and corpora differ, so
Section~\ref{sec:artifact-analysis} argues feature-level claims within a family, not across
families.

\begin{table}[htbp]
  \centering
  \caption{The dictionaries trained for this study, one row per Pythia model, median over
  the depths measured. Width is held at eight times the residual width and sparsity at 64
  active features, so the comparison across scale is not also a comparison across
  dictionary capacity. Reconstruction error grows with model size at fixed expansion.}
  \label{tab:dictionaries}
  \begin{tabular}{lrrrrr}
    \toprule
    Model & $d_{\mathrm{model}}$ & Width & Expansion & Active & FVU \\
    \midrule
    \TableDictionaryBody
    \bottomrule
  \end{tabular}
\end{table}

Given a set of features, Section~\ref{sec:artifact-analysis} compares three states of the
same quantised activation: \emph{keep} leaves it alone, \emph{zero} sets the selected feature
activations to zero, and \emph{restore} replaces them with their full-precision values while
carrying the part of the activation the dictionary does not model through unchanged, so that
restoring the empty set reproduces the quantised activation exactly and no reconstruction
error enters the comparison. Restoration is an oracle intervention: it writes back values
taken from the full-precision model, so it identifies which directions carry the damage
without being a repair anyone could deploy (Section~\ref{sec:limitations}).

\begin{table}[htbp]
  \centering
  \caption{Share of total damage carried by the most affected features, against the share of
  activation those features carry (Section~\ref{sec:cka}). Quantisation
  concentrates about 5.6 percent of its damage in the top one percent, close
  to the activation share of \ActivationShare{} percent; noise of identical size concentrates
  far less. Median over four families.}
  \label{tab:concentration}
  \begin{tabular}{lrrrr}
    \toprule
    Share of damage on the top & 1\% & 3\% & 5\% & 10\% \\
    \midrule
    \TableConcentrationBody
    \bottomrule
  \end{tabular}
\end{table}

\begin{table}[htbp]
  \centering
  \caption{Signed shift of feature activations, as a fraction of their full-precision scale
  (Section~\ref{sec:shunt}). Quantisation moves features down and noise of identical
  per-channel size moves them up. The split appears at the same bitwidth as the perplexity
  gap. Median over four families.}
  \label{tab:shunt}
  \begin{tabular}{lrr}
    \toprule
    Bits & Quantisation & Matched noise \\
    \midrule
    \TableShuntBody
    \bottomrule
  \end{tabular}
\end{table}

\begin{table}[htbp]
  \centering
  \caption{Causal interventions (Section~\ref{sec:causal}), perplexity relative to the
  quantised model. Values below one are a recovery, above one a further loss. Restoring the
  most damaged features dominates restoring as many random or least damaged ones at every
  size, and zeroing them is harmful throughout. Median over four families.}
  \label{tab:causal}
  \begin{tabular}{lrrrr}
    \toprule
    & \multicolumn{3}{c}{Restore} & Zero \\
    \cmidrule(lr){2-4}\cmidrule(lr){5-5}
    Features & Most damaged & Random & Least damaged & Most damaged \\
    \midrule
    \TableCausalBody
    \bottomrule
  \end{tabular}
\end{table}

\begin{figure}[htbp]
  \centering
  \includegraphics[width=0.6\linewidth]{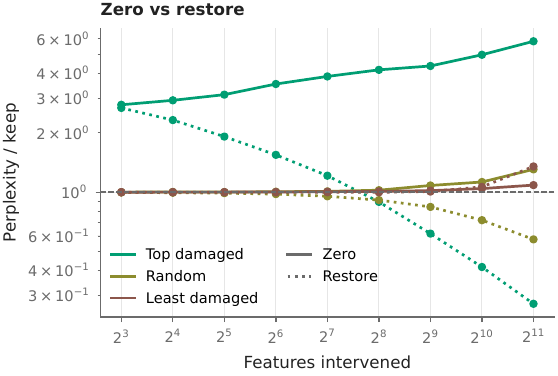}
  \caption{Table~\ref{tab:causal} plotted across every intervention size measured: perplexity
  relative to the quantised model as the number of intervened features grows, for restoration
  of the most damaged, random, and least damaged sets, and for zeroing the most damaged set.}
  \label{fig:causal}
\end{figure}

\section{Granularity and sensitivity: full results}
\label{app:regime}

These measurements set the scene for Section~\ref{sec:noise}; they do not carry its argument
on their own. This appendix gives the per-model resolution of the structure premium
(Section~\ref{sec:premium}) and the full results of the subspace-aware allocation experiment
(Section~\ref{sec:restore}).

\subsection{Structure premium, pooled and by model}

Section~\ref{sec:premium} summarises the structure premium in prose; Table~\ref{tab:premium}
gives the full pooled table it is drawn from, and Table~\ref{tab:scale} resolves the same
quantity per model. Every model is
indistinguishable from noise at eight and six bits; where it departs below that varies with
both size and architecture, so the pooled median in the main text is a summary of a pattern
that holds broadly, not a single shared threshold.

\begin{table}[htbp]
  \centering
  \caption{Structure premium: quantised perplexity divided by the perplexity of Gaussian noise
  of identical per-channel mean squared error. A value of one means the error behaves exactly
  like noise of its size. Median over \NumModels{} models and three relative depths.}
  \label{tab:premium}
  \begin{tabular}{lrrrrr}
    \toprule
    Scale shared over & 8 bits & 6 bits & 4 bits & 3 bits & 2 bits \\
    \midrule
    One channel & \PremFeatureEight{} & \PremFeatureSix{} & \PremFeatureFour{} & \PremFeatureThree{} & \PremFeatureTwo{} \\
    One token   & \PremTokenEight{}   & \PremTokenSix{}   & \PremTokenFour{}   & \PremTokenThree{}   & \PremTokenTwo{} \\
    Whole tensor & \PremTensorEight{} & \PremTensorSix{}  & \PremTensorFour{}  & \PremTensorThree{}  & \PremTensorTwo{} \\
    \bottomrule
  \end{tabular}
\end{table}

\begin{table}[htbp]
  \centering
  \caption{Structure premium per model under per-token scaling, the granularity deployed
  quantisers use. Median over three relative depths. Every model is indistinguishable from
  noise at eight and six bits. Where it departs below that varies with both size and
  architecture.}
  \label{tab:scale}
  \begin{tabular}{lrrrrr}
    \toprule
    Model & 8 bits & 6 bits & 4 bits & 3 bits & 2 bits \\
    \midrule
    \TableScaleBody
    \bottomrule
  \end{tabular}
\end{table}

\begin{figure}[htbp]
  \centering
  \includegraphics[width=0.55\linewidth]{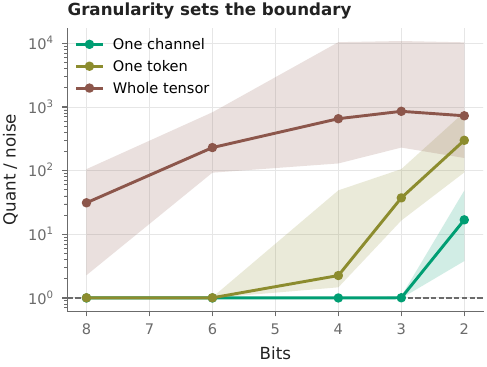}
  \caption{Structure premium against bitwidth (Table~\ref{tab:premium}, main text). A value of
  one, marked by the dashed line, means quantisation and matched noise do equal harm. Curves
  separate by how widely the quantiser shares a scale, not by bitwidth.}
  \label{fig:structure}
\end{figure}

\subsection{Sensitivity versus magnitude, by family and by depth}

\begin{figure}[htbp]
  \centering
  \includegraphics[width=0.48\linewidth]{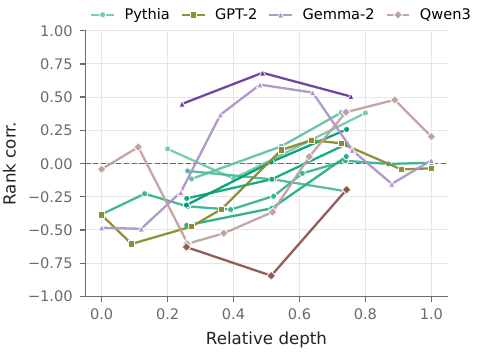}
  \hfill
  \includegraphics[width=0.48\linewidth]{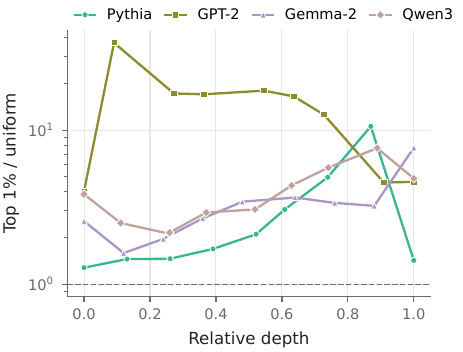}
  \caption{Left: rank correlation between sensitivity and activation magnitude against
  relative depth. Right: sensitivity mass in the most sensitive one percent of coordinates, as
  a multiple of a flat spectrum. Only the models swept at every depth are drawn.}
  \label{fig:sensitivity}
\end{figure}

Section~\ref{sec:sensitivity} summarises the mismatch between loss sensitivity and activation
magnitude in prose; Table~\ref{tab:sensitivity} gives the per-family numbers behind it, and
Figure~\ref{fig:sensitivity} resolves both quantities by relative depth for the models swept
at every depth.

\begin{table}[htbp]
  \centering
  \caption{Left: loss sensitivity against activation magnitude, by family. Rank correlation is
  Spearman's over residual coordinates, and overlap is the share of members common to the 64
  most sensitive coordinates and the 64 largest by mean absolute activation. Right:
  sensitivity mass in the most sensitive one percent of coordinates, as a multiple of what a
  flat spectrum would place there, at relative depth below 0.2, between 0.4 and 0.6, and
  above 0.8.}
  \label{tab:sensitivity}
  \begin{tabular}{lrlrrrr}
    \toprule
    & \multicolumn{3}{c}{Sensitivity vs magnitude} & \multicolumn{3}{c}{Top 1\% Fisher, over uniform} \\
    \cmidrule(lr){2-4}\cmidrule(lr){5-7}
    Family & Median corr. & Range & Overlap & Shallow & Middle & Deep \\
    \midrule
    \TableSensitivityBody
    \bottomrule
  \end{tabular}
\end{table}

\subsection{Subspace-aware allocation: full results}
\label{sec:allocation-full}

Section~\ref{sec:restore} reports protecting the most Fisher-sensitive directions
(Section~\ref{sec:sensitivity}) under the per-token scaling used in the main text, where it helps
only at three average bits and below. Here we give the full sweep, which adds per-tensor scaling:
there, protection is worth an order of magnitude in perplexity, while under per-token scaling it
is a liability from four bits up. Table~\ref{tab:allocation} holds a small set of the most
sensitive directions at high precision and drops the rest to a matched average bitwidth, on
all \NumModels{} models.

\begin{table}[htbp]
  \centering
  \caption{Precision that protects the most sensitive directions, against uniform precision
  at the same average bitwidth. Perplexities are for per-tensor scaling; the final column
  gives the same ratio under per-token scaling. Median over \NumModels{} models and three
  relative depths. Protection is worth an order of magnitude where the error is organised
  and is a liability where it is already close to noise.}
  \label{tab:allocation}
  \begin{tabular}{lrrrr}
    \toprule
    & \multicolumn{3}{c}{Per tensor} & Per token \\
    \cmidrule(lr){2-4}\cmidrule(lr){5-5}
    Average bits & Uniform & Protected & Ratio & Ratio \\
    \midrule
    \TableAllocationBody
    \bottomrule
  \end{tabular}
\end{table}

The two columns of ratios track the structure premium of Table~\ref{tab:premium}, not the
bitwidth. Protection is funded by taking bits from every unprotected direction, so it only
pays off when the error it removes is organised enough to be worth more than the bits it
costs. Per-tensor error is organised at every bitwidth; per-token error is not. That is why
one column never inverts and the other inverts as soon as quantisation stops being
destructive: under per-tensor scaling, protecting a small subspace is worth between 5.9 and
22.9 times in perplexity at every bitwidth we measure, including eight bits; under per-token
scaling it is worth 6.3 times at two bits, but then it turns into a liability, costing between
two and fourteen times from three bits upward.

\begin{figure}[htbp]
  \centering
  \includegraphics[width=0.55\linewidth]{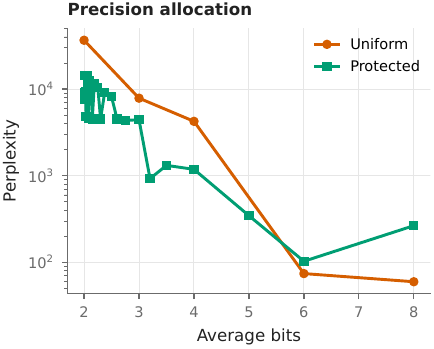}
  \caption{Perplexity against average bitwidth under uniform precision and under precision
  protecting the most sensitive directions.}
  \label{fig:allocation}
\end{figure}

Table~\ref{tab:protect} isolates the same effect at a fixed split (a small subspace held at
eight bits, the remainder dropped to two) rather than a continuum of average bitwidths.
Fisher-selected protection tracks uniform precision closely at three and four bits, where
protection has little to work with, and pulls far ahead by six bits, where the subspace it
protects is doing most of the surviving work; but, as Section~\ref{sec:restore} reports,
this perplexity advantage does not translate into the induction/retrieval accuracy the probes
of Section~\ref{sec:mask} measure, because the protected subspace is selected by loss
sensitivity, not by the coordinate-aligned pattern those probes are sensitive to.

\begin{table}[htbp]
  \centering
  \caption{Induction retained when a subspace is held at eight bits and the remainder at
  two, against uniform precision costing no less on average. Fisher selects the most
  sensitive directions; random selects as many at random. The uniform reference is the cheapest
  uniform bitwidth we ran at or above the row's average cost (``at $b$'' gives it); there is no
  uniform 5-bit run, so the 5.00-bit row is compared with 6 bits, which favours uniform precision.
  Median over four families.}
  \label{tab:protect}
  \begin{tabular}{lrrr}
    \toprule
    Average bits & Fisher-selected & Random & Uniform reference \\
    \midrule
    \TableProtectBody
    \bottomrule
  \end{tabular}
\end{table}

\section{Tables resolved by bitwidth and by model}
\label{app:tables}

The main text states these results and reads them off the summary tables there. Here we
resolve the same measurements by bitwidth, by model, and by family, for readers who want the
full breakdown rather than the median. We order them to follow the main text's own sections,
rather than dump them as one undifferentiated block.

Below is Table~\ref{tab:probes}'s substitute ladder (Section~\ref{sec:induction}), resolved
by bitwidth. The quantised row collapses between four and three bits, while every substitute
below it degrades gradually across the same range. This is the bitwidth-resolved version of
the gap the main text reads off the median.

\begin{table}[htbp]
  \centering
  \caption{Fraction of induction accuracy retained under each perturbation, median over
  four families at three relative depths. The intact models score between 0.954 and 0.987,
  and the chance floor is zero. Rows within a block carry the same per-channel error
  magnitude as the quantised row above them.}
  \label{tab:induction}
  \begin{tabular}{lrrrrr}
    \toprule
    Perturbation & 8 bits & 6 bits & 4 bits & 3 bits & 2 bits \\
    \midrule
    \TableInductionBody
    \bottomrule
  \end{tabular}
\end{table}

\begin{figure}[htbp]
  \centering
  \includegraphics[width=0.55\linewidth]{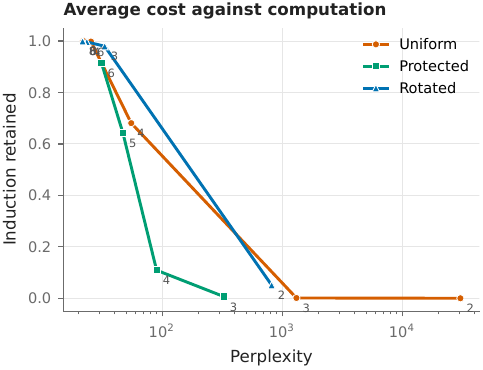}
  \caption{Table~\ref{tab:pareto} (main text) on a perplexity axis. Subspace-selection
  (Protected) tracks uniform quantisation and then collapses with it; rotation (Rotated)
  reaches a materially better tradeoff at every average bitwidth from four down to two.}
  \label{fig:pareto}
\end{figure}

Rotation's repair tracks the same rows. Compared against the unrotated per-token block of
Table~\ref{tab:induction} above, it stays near intact one bitwidth further down (0.980 at
three bits, against 0.001 unrotated), before it too collapses at two bits. Per-tensor
rotation only helps at eight and six bits, and has already collapsed by four. This fits with
per-tensor error being organised enough that even a rotated frame cannot spread it out in
time.

\begin{table}[htbp]
  \centering
  \caption{Induction retained when the same quantiser is applied in a randomly rotated
  frame. Compare the unrotated rows of Table~\ref{tab:induction}. Median over four families
  at three relative depths.}
  \label{tab:restore}
  \begin{tabular}{lrrrrr}
    \toprule
    Perturbation & 8 bits & 6 bits & 4 bits & 3 bits & 2 bits \\
    \midrule
    \TableRestoreBody
    \bottomrule
  \end{tabular}
\end{table}

Below is Section~\ref{sec:calibration}'s dose-response band (Table~\ref{tab:dose}), resolved
by family rather than pooled. Every family here has its retrieval median below its induction
median in the same 1.2-to-1.5 band. So the split is not an artefact of pooling four families
into one row.

\begin{table}[htbp]
  \centering
  \caption{The headline band of Table~\ref{tab:dose}, perplexity between 1.2 and 1.5 times
  intact, resolved by family. Ranges are over the settings in the band, which differ in
  depth, bitwidth, arm and whether one block or every block is quantised, so a range is the
  spread of a heterogeneous set rather than an interval on a repeated measurement. Every
  family has retrieval below induction. Qwen3-1.7B-Base's retrieval range reaches above 1.0
  (retention exceeding the intact model); at \PositionsRetrieval{} positions this reflects
  probe measurement noise, not a real gain from quantisation.}
  \label{tab:dosemodel}
  \begin{tabular}{lrrlrl}
    \toprule
    & & \multicolumn{2}{c}{Induction} & \multicolumn{2}{c}{Retrieval} \\
    \cmidrule(lr){3-4}\cmidrule(lr){5-6}
    Family & Settings & Median & Range & Median & Range \\
    \midrule
    \TableDoseModelBody
    \bottomrule
  \end{tabular}
\end{table}

Below is Section~\ref{sec:restore}'s end-to-end counterpart: every block from the intervened
one onward is quantised, rather than just one. The same ordering between uniform, protected,
and rotated arms holds, but the usable range is narrower. Rotation still reaches 0.968
induction at four bits, close to the single-block setting's 0.997, but at three bits it has
fallen to 0.380, where the single-block setting was already at 0.980. So the same repair
needs one more bit here to do the same job.

\begin{table}[htbp]
  \centering
  \small
  \caption{Every block from the intervened one onward quantised, at every average bitwidth
  measured, median over four families. Full precision is 21.6. This resolves
  Section~\ref{sec:alllayer}, which reads the four and six bit rows.}
  \label{tab:alllayerfull}
  \begin{tabular}{lrrrr}
    \toprule
    Precision & Average bits & Perplexity & Induction & Retrieval \\
    \midrule
    \TableAllLayerFullBody
    \bottomrule
  \end{tabular}
\end{table}

The comparisons above pool over four families by taking the median at each cell. The
differences between those families are real and worth seeing directly, per model rather than
pooled:

\begin{table}[h!]
  \centering
  \caption{Per model rather than pooled, so that heterogeneity between families is visible.
  Intact accuracy is the unquantised model. The last three columns are the fraction of
  induction retained. Pythia-1.4B is the outlier: it has already lost induction at four bits
  where the others retain most of it, and it is the model whose median pulls the pooled four
  bit figure down.}
  \label{tab:permodel}
  \begin{tabular}{lrrrrr}
    \toprule
    & \multicolumn{2}{c}{Intact accuracy} & \multicolumn{3}{c}{Induction retained} \\
    \cmidrule(lr){2-3}\cmidrule(lr){4-6}
    Model & Induction & Retrieval & Quant 4b & Quant 3b & Rotated 3b \\
    \midrule
    \TablePerModelBody
    \bottomrule
  \end{tabular}
\end{table}

At four bits per token the retained fraction
runs from 0.014 in Pythia-1.4B to 0.926 in GPT-2, so the pooled median at that bitwidth sits
between two different behaviours, rather than describing one common behaviour. The
comparisons the paper rests on are not like this: at three bits every model retains
essentially nothing under quantisation, and between 0.864 and 0.992 under rotation. The
direction is the same in all four models, so the median is a fair summary there.

\end{document}